\documentclass[twoside,11pt]{article}
\usepackage{jair, theapa, rawfonts}

\usepackage[implicit=false,colorlinks=false,allbordercolors={1 1 1}]{hyperref}
\usepackage{graphicx}
\usepackage{subcaption}
\usepackage{booktabs}
\usepackage{amsmath,amsfonts}
\usepackage{bbm}
\usepackage[inline]{enumitem}
\usepackage[shortcuts,acronym]{glossaries}
\usepackage{placeins}

\DeclareMathOperator*{\argmin}{arg\,min \;}

\newcommand{\ul}[1]{\underline{#1}}
\newcommand*{\given}[1][]{\;#1\vert\;}
\newcommand*{\name}[1]{\textsc{#1}}
\newcommand*\diff{\mathop{}\!\mathrm{d}}
\newcommand{\ie}{i.e.}
\newcommand{\eg}{e.g.}

\newacronym{AutoML}{AutoML}{Automated machine learning}
\newacronym{CASH}{CASH}{combined algorithm selection and hyperparameter optimization}
\newacronym{CRISP-DM}{CRISP-DM}{cross-industry standard process for data mining}
\newacronym{EPM}{EPM}{empirical performance model}
\newacronym{DAG}{DAG}{directed acyclic graph}
\newacronym{HPO}{HPO}{hyperparameter optimization}
\newacronym{HTN}{HTN}{Hierarchical task network}
\newacronym{KDE}{KDE}{kernel density estimation}
\newacronym{ML}{ML}{machine learning}
\newacronym{PCA}{PCA}{principal component analysis}
\newacronym{SMBO}{SMBO}{sequential model-based optimization}
\newacronym{SVM}{SVM}{support-vector machine}
\newacronym{TDSP}{TDSP}{Team Data Science Process}
\newacronym{TPE}{TPE}{tree-structured Parzen estimator}

\newtheorem{definition}{Definition}

\usepackage[normalem]{ulem}
\usepackage{color}

\jairheading{70}{2021}{409-474}{12/2019}{01/2021}
\ShortHeadings{Benchmark and Survey of Automated Machine Learning Frameworks}{Z\"oller \& Huber}
\firstpageno{409}

\begin{document}

\title{
	Benchmark and Survey of Automated \\ Machine Learning Frameworks
}

\author{\name Marc-Andr\'e Z\"oller \email marc.zoeller@usu.com \\
		\addr USU Software AG\\
		R\"uppurrer Str. 1, Karlsruhe, Germany
		\AND
		\name Marco F. Huber \email marco.huber@ieee.org \\
		\addr Institute of Industrial Manufacturing and Management IFF, \\
		University of Stuttgart, Allmandring 25, Stuttgart, Germany \& \\
		Fraunhofer Institute for Manufacturing Engineering and Automation IPA \\
		Nobelstr. 12, Stuttgart, Germany}

\maketitle


\begin{abstract}%
\Ac{ML} has become a vital part in many aspects of our daily life. However, building well performing machine learning applications requires highly specialized data scientists and domain experts. \ac{AutoML} aims to reduce the demand for data scientists by enabling domain experts to build machine learning applications automatically without extensive knowledge of statistics and machine learning. This paper is a combination of a survey on current \ac{AutoML} methods and a benchmark of popular \ac{AutoML} frameworks on real data sets. Driven by the selected frameworks for evaluation, we summarize and review important \ac{AutoML} techniques and methods concerning every step in building an \ac{ML} pipeline. The selected \ac{AutoML} frameworks are evaluated on \(137\) data sets from established \ac{AutoML} benchmark suits.
\end{abstract}

\section{Introduction}

In recent years \ac{ML} is becoming ever more important: automatic speech recognition, self-driving cars or predictive maintenance in Industry 4.0 are build upon \ac{ML}. \ac{ML} is nowadays able to beat human beings in tasks often described as too complex for computers, \eg, \name{AlphaGO} \shortcite{Silver2017} was able to beat the human champion in \name{GO}. Such examples are powered by extremely specialized and complex \ac{ML} pipelines.

In order to build such an \ac{ML} pipeline, a highly trained team of human experts is necessary: data scientists have profound knowledge of \ac{ML} algorithms and statistics; domain experts often have a longstanding experience within a specific domain. Together, those human experts can build a sensible \ac{ML} pipeline containing specialized data preprocessing, domain-driven meaningful feature engineering and fine-tuned models leading to astonishing predictive power. Usually, this process is a very complex task, performed in an iterative manner with trial and error. As a consequence, building good \ac{ML} pipelines is a long and expensive endeavor and practitioners often use a suboptimal default \ac{ML} pipeline.

\ac{AutoML} aims to improve the current way of building \ac{ML} applications by automation. \ac{ML} experts can profit from \ac{AutoML} by automating tedious tasks like \ac{HPO} leading to a higher efficiency. Domain experts can be enabled to build \ac{ML} pipelines on their own without having to rely on a data scientist.

It is important to note that \ac{AutoML} is not a new trend. Starting from the 1990s, commercial solutions offered automatic \ac{HPO} for selected classification algorithms via grid search \shortcite{Dinsmore2016}. Adaptations of grid search to test possible configurations in a greedy best-first approach are available since 1995 \shortcite{Kohavi1995}. In the early 2000s, the first efficient strategies for \ac{HPO} have been proposed. For limited settings, \eg, tuning \(C\) and \(\gamma\) of a \ac{SVM} \shortcite{Momma2002,Chapelle2002,Chen2004}, it was proven that guided search strategies yield better results than grid search in less time. Also in 2004, the first approaches for automatic feature selection have been published \shortcite{Samanta2004}. \emph{Full model selection} \shortcite{Escalante2009} was the first attempt to build a complete \ac{ML} pipeline automatically by selecting a preprocessing, feature selection and classification algorithm simultaneously while tuning the hyperparameters of each method. Testing this approach on various data sets, the potential of this domain-agnostic method was proven \shortcite{Guyon2008}. Starting from 2011, many different methods applying Bayesian optimization for hyperparameter tuning \shortcite{Bergstra2011,Snoek2012} and model selection \shortcite{Thornton2013} have been proposed. In 2015, the first method for automatic feature engineering without domain knowledge was proposed \shortcite{Kanter2015}. Building variable shaped pipelines is possible since 2016 \shortcite{Olson2016}. In 2017 and 2018, the topic \ac{AutoML} received a lot of attention in the media with the release of commercial \ac{AutoML} solutions from various global players \shortcite{Golovin2017,Clouder2018,Baidu2018,Das2020}. Simultaneously, research in the area of \ac{AutoML} gained significant traction leading to many performance improvements. Recent methods are able to reduce the runtime of \ac{AutoML} procedures from several hours to mere minutes \shortcite{Hutter2018a}.

This paper is a combination of a short survey on \ac{AutoML} and an evaluation of frameworks for \ac{AutoML} and \ac{HPO} on real data. We select \(14\) different \ac{AutoML} and \ac{HPO} frameworks in total for evaluation. The techniques used by those frameworks are summarized to provide an overview for the reader. This way, research concerning the automation of any aspect of an \ac{ML} pipeline is reviewed: determining the pipeline structure, selecting an \ac{ML} algorithm for each stage in a pipeline and tuning each algorithm. The paper focuses on classic machine learning and does \textbf{not} consider neural network architecture search while still many of the ideas can be transferred. Most topics discussed in this survey are large enough to be handled in dedicated surveys. Consequently, this paper does not aim to handle each topic in exhaustive depth but aims to provide a profound overview. The contributions are:
\begin{itemize}
	\item We introduce a mathematical formulation covering the complete procedure of automatic \ac{ML} pipeline synthesis and compare it with existing problem formulations.
	\item We review open-source frameworks for building \ac{ML} pipelines automatically.
	\item An evaluation of eight \ac{HPO} algorithms on \(137\) real data sets is conducted. To the best of our knowledge, this is the first independent benchmark of \ac{HPO} algorithms.
	\item An empirical evaluation of six \ac{AutoML} frameworks on \(73\) real data sets is performed. To the best of our knowledge, this is the most extensive evaluation---in terms of tested frameworks as well as used data sets---of \ac{AutoML} frameworks.
\end{itemize}
In doing so, readers will get a comprehensive overview of state-of-the-art \ac{AutoML} algorithms. All important stages of building an \ac{ML} pipeline automatically are introduced and existing approaches are evaluated. This allows revealing the limitations of current approaches and raising open research questions.

Lately, several surveys regarding \ac{AutoML} have been published. \shortciteA{Elshawi2019} and \shortciteA{He2019} focus on automatic neural network architecture search---which is not covered in this survey---and only briefly introduce methods for classic machine learning. \shortciteA{Quanming2018} and \shortciteA{Hutter2018} cover less steps of the pipeline creation process and do not provide an empirical evaluation of the presented methods. Finally, \shortciteA{Tuggener2019} provides only a high-level overview.

Two benchmarks of \ac{AutoML} methods have been published so far. \shortciteA{Balaji2018} and \shortciteA{Gijsbers2019} evaluate various \ac{AutoML} frameworks on real data sets. Our evaluations exceed those benchmarks in terms of evaluated data sets as well as evaluated frameworks. Both benchmarks focus only on a performance comparison while we also take a look at the obtained \ac{ML} models and pipelines. Furthermore, both benchmarks do not consider \ac{HPO} methods.

In Section~\ref{sec:problem_formulation} a mathematical sound formulation of the automatic construction of \ac{ML} pipelines is given. Section~\ref{sec:pipeline_structure} presents different strategies for determining a pipeline structure. Various approaches for \ac{ML} model selection and \ac{HPO} are theoretically explained in Section~\ref{sec:cash}. Next, methods for automatic data cleaning (Section~\ref{sec:data_cleaning}) and feature engineering (Section~\ref{sec:feature_engineering}) are introduced. Measures for improving the performance of the generated pipelines as well as decreasing the optimization runtime are explained in Section~\ref{sec:performance_improvements}. Section~\ref{sec:selected_frameworks} introduces the evaluated \ac{AutoML} frameworks. The evaluation is presented in Section~\ref{sec:experiments}. Opportunities for further research are presented in Section~\ref{sec:future_research} followed by a short conclusion in Section~\ref{sec:conclusion}.

\section{Problem Formulation}
\label{sec:problem_formulation}

An \ac{ML} pipeline \(h: \mathbb{X} \rightarrow \mathbb{Y}\) is a sequential combination of various algorithms that transforms a feature vector \(\vec{x} \in \mathbb{X}\) into a target value \(y \in \mathbb{Y}\), \eg, a class label for a classification problem. Let a fixed set of basic algorithms, \eg, various classification, imputation and feature selection algorithms, be given as \(\mathcal{A} = \left\{ A^{(1)}, A^{(2)}, \dots, A^{(n)} \right\} \). Each algorithm \(A^{(i)}\) is configured by a vector of hyperparameters \(\vec{\lambda}^{(i)}\) from the domain \(\Lambda_{A^{(i)}}\).

Without loss of generality, let a pipeline structure be modeled as a \ac{DAG}. Each node represents a basic algorithm. The edges represent the flow of an input data set through the different algorithms. Often the \ac{DAG} structure is restricted by implicit constraints, \ie, a pipeline for a classification problem has to have a classification algorithm as the last step. Let \(G\) denote the set of valid pipeline structures and \(\left| g \right|\) denote the length of a pipeline, \ie, the number of nodes in \(g \in G\).

\begin{definition}[Machine Learning Pipeline]
	Let a triplet \((g, \vec{A}, \vec{\lambda})\) define an \ac{ML} pipeline with \(g \in G\) a valid pipeline structure, \(\vec{A} \in \mathcal{A}^{|g|} \) a vector consisting of the selected algorithm for each node and \(\vec{\lambda}\) a vector comprising the hyperparameters of all selected algorithms. The pipeline is denoted as \( \mathcal{P}_{g, \vec{A}, \vec{\lambda}} \).
\end{definition}

Following the notation from empirical risk minimization, let \(P(\mathbb{X}, \mathbb{Y})\) be a joint probability distribution of the feature space \(\mathbb{X}\) and target space \(\mathbb{Y}\) known as a \textit{generative model}. We denote a pipeline trained on the generative model \(P\) as \( \mathcal{P}_{g, \vec{A}, \vec{\lambda}, P} \).

\begin{definition}[True Pipeline Performance]
Let a pipeline \(\mathcal{P}_{g, \vec{A}, \vec{\lambda}}\) be given. Given a loss function \(\mathcal{L}(\cdot, \cdot)\) and a generative model \(P(\mathbb{X}, \mathbb{Y})\), the performance of \(\mathcal{P}_{g, \vec{A}, \vec{\lambda}, P}\) is calculated as
\begin{equation}
\label{eq:true_pipeline_performance}
	R \left(\mathcal{P}_{g, \vec{A}, \vec{\lambda}, P}, P \right) = \mathbb{E} \big( \mathcal{L}(h(\mathbb{X}), \mathbb{Y}) \big) = \int \mathcal{L}\big(h(\mathbb{X}), \mathbb{Y}\big) \diff P(\mathbb{X}, \mathbb{Y}),
\end{equation}
with \(h(\mathbb{X})\) being the predicted output of \(\mathcal{P}_{g, \vec{A}, \vec{\lambda}, P}\).
\end{definition}
Let an \emph{\ac{ML} task} be defined by a generative model, loss function and an \ac{ML} problem type, \eg, classification or regression. Generating an \ac{ML} pipeline for a given \ac{ML} task can be split into three tasks: first, the structure of the pipeline has to be determined, \eg, selecting how many preprocessing and feature engineering steps are necessary, how the data flows through the pipeline and how many models have to be trained. Next, for each step an algorithm has to be selected. Finally, for each selected algorithm its corresponding hyperparameters have to be selected. All steps have to be completed to actually evaluate the pipeline performance.

\begin{definition}[Pipeline Creation Problem]
\label{def:pcp}
	Let a set of algorithms \(\mathcal{A}\) with an according domain of hyperparameters \(\Lambda_{(\cdot)}\), a set of valid pipeline structures \(G\) and a generative model \(P(\mathbb{X}, \mathbb{Y})\) be given. The pipeline creation problem consists of finding a pipeline structure in combination with a joint algorithm and hyperparameter selection that minimizes the loss
\begin{equation}
\label{eq:pcp}
	(g, \vec{A}, \vec{\lambda})^\star \in \argmin_{g \in G, \vec{A} \in \mathcal{A}^{|g|}, \vec{\lambda} \in \Lambda} R \left(\mathcal{P}_{g, \vec{A}, \vec{\lambda}, P}, P \right) .
\end{equation}
\end{definition}
In general, Equation~\eqref{eq:pcp} cannot be computed directly as the distribution \(P(\mathbb{X}, \mathbb{Y})\) is unknown. Instead, let a finite set of observations \(D = \{ \left( \vec{x}_{1}, y_{1} \right), \dots, \left( \vec{x}_{m}, y_{m} \right) \}\) of \(m\) i.i.d samples drawn from \(P(\mathbb{X}, \mathbb{Y})\)
be given. Equation~\eqref{eq:true_pipeline_performance} can be adapted to \(D\) to calculate an \textit{empirical pipeline performance} as
\begin{equation}
\label{eq:empirical_pcp}
	\hat{R} \left(\mathcal{P}_{g, \vec{A}, \vec{\lambda}, D}, D \right) = \dfrac{1}{m} \sum_{i = 1}^m \mathcal{L} \left( h(x_{i}), y_{i} \right) .
\end{equation}
To limit the effects of overfitting, Equation~\eqref{eq:empirical_pcp} is often augmented by cross-validation. Let the data set \(D\) be split into \(k\) folds \( \{ D_{\text{valid}}^{(1)}, \dots, D_{\text{valid}}^{(k)} \} \) and \( \{ D_{\text{train}}^{(1)}, \dots, D_{\text{train}}^{(k)} \} \) such that \( D_{\text{train}}^{(i)} = D \setminus D_{\text{valid}}^{(i)} \). The final objective function is defined as
\begin{equation*}
	(g, \vec{A}, \vec{\lambda})^\star \in \argmin_{g \in G, \vec{A} \in \mathcal{A}^{|g|}, \vec{\lambda} \in \Lambda} \dfrac{1}{k} \sum_{i = 1}^k \hat{R} \left(\mathcal{P}_{g, \vec{A}, \vec{\lambda}, D^{(i)}_{\text{train}}}, D^{(i)}_{\text{valid}} \right) .
\end{equation*}

This problem formulation is a generalization of existing problem formulations. Current problem formulations only consider selecting and tuning a single algorithm (\eg, \shortciteR{Escalante2009,Bergstra2011}) or a linear sequence of algorithms with (arbitrary but) fixed length (\eg, \shortciteR{Thornton2013,Zhang2016,Alaa2018,Hutter2018}). \shortciteA{Salvador2017a} model an \ac{ML} pipeline with Petri-nets \shortcite{Petri1962} instead of a \ac{DAG}. Using additional constraints, the Petri-net is enforced to represent a \ac{DAG}. Even though this approach is more expressive than \acp{DAG}, the additional model capabilities are currently not utilized in the context of \ac{AutoML}.

Using Equation~\eqref{eq:pcp}, the pipeline creation problem is formulated as a black box optimization problem. Finding the global optimum in such equations has been the subject of decades of study \shortcite{Snyman2005}. Many different algorithms have been proposed to solve specific problem instances efficiently, for example convex optimization. To use these methods, the features and shape of the underlying objective function---in this case the loss \(\mathcal{L}\)---have to be known to select applicable solvers. In general, it is not possible to predict any properties of the loss function or even formulate it as closed-form expression as it depends on the generative model. Consequently, efficient solvers, like convex or gradient-based optimization, cannot be used for Equation~\eqref{eq:pcp} \shortcite{Luo2016}.

\begin{figure}[ht]
	\centering
	\includegraphics[width=0.5\linewidth]{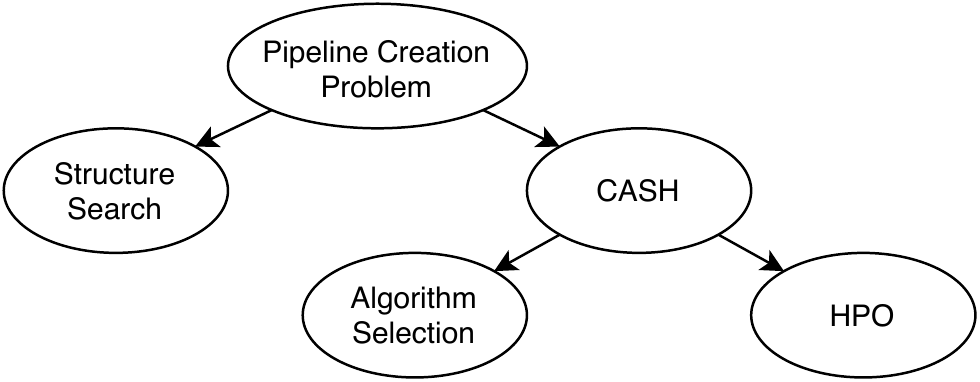}
	\caption{Subproblems of the pipeline creation problem.}
	\label{fig:problem_overview}
\end{figure}

Human \ac{ML} experts usually solve the pipeline creation problem in an iterative manner: At first a simple pipeline structure with standard algorithms and default hyperparameters is selected. Next, the pipeline structure is adapted, potentially new algorithms are selected and hyperparameters are refined. This procedure is repeated until the overall performance is sufficient. In contrast, most current state-of-the-art algorithms solve the pipeline creation problem in a single step. Figure~\ref{fig:problem_overview} shows a schematic representation of the different optimization problems for the automatic composition of \ac{ML} pipelines. Solutions for each subproblem are presented in the following sections.

\section{Pipeline Structure Creation}
\label{sec:pipeline_structure}
\begin{figure}[hb]
	\centering
	\includegraphics[width=0.7\linewidth]{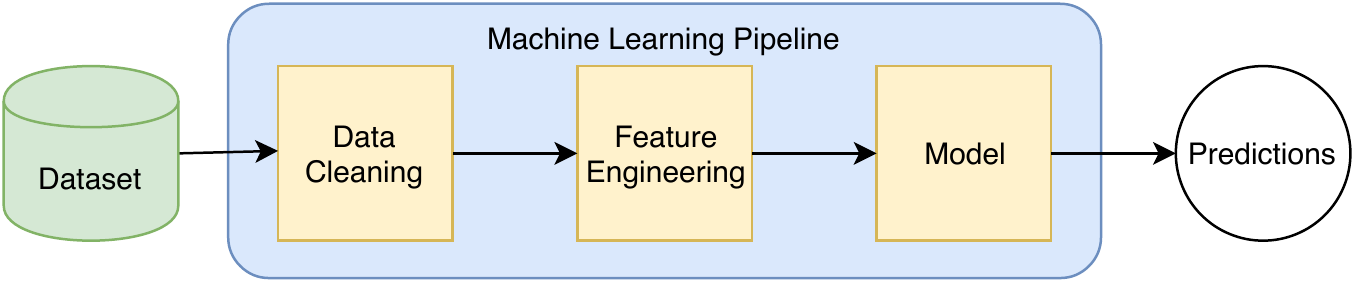}
	\caption{Prototypical \ac{ML} pipeline. First, the input data is cleaned and features are extracted. The transformed input is passed through an \ac{ML} model to create predictions.}
	\label{fig:ml_pipeline}
\end{figure}

The first task for building an \ac{ML} pipeline is creating the pipeline structure. Common best practices suggest a basic \ac{ML} pipeline layout as displayed in Figure~\ref{fig:ml_pipeline} \shortcite{Kegl2017,Ayria2018,Zhou2018}. At first, the input data is cleaned in multiple distinct steps, like imputation of missing data and one-hot encoding of categorical input. Next, relevant features are selected and new features created. This stage highly depends on the underlying domain. Finally, a single model is trained on the previously selected features. In practice this simple pipeline is usually adapted and extended by experienced data scientists.

\subsection{Fixed Structure}
\begin{figure}[ht]
	\centering
	\includegraphics[width=0.7\linewidth]{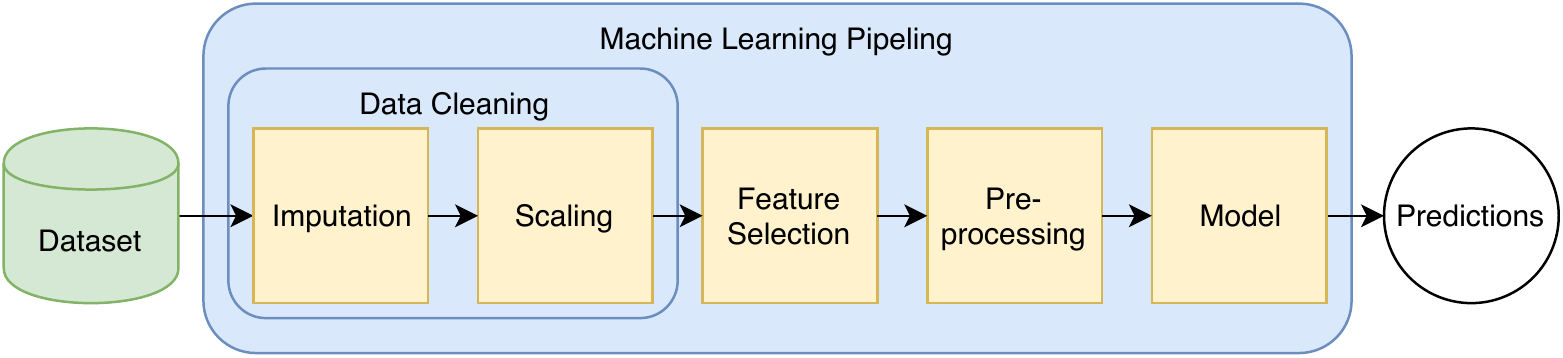}
	\caption{Fixed \ac{ML} pipeline used by most \ac{AutoML} frameworks. Minor differences exist regarding the implemented data cleaning steps.}
	\label{fig:automl_fixed_pipeline}
\end{figure}

Many \ac{AutoML} frameworks do not solve the structure selection because they are preset to the fixed pipeline structure displayed in Figure~\ref{fig:automl_fixed_pipeline} (\eg, \shortciteR{Komer2014,Feurer2015,Swearingen2017,Parry2019,McGushion2019}). Resembling the best practice pipeline closely, the pipeline is a linear sequence of multiple data cleaning steps, a feature selection step, one variable preprocessing step and exactly one modeling step. The preprocessing step chooses one algorithm from a set of well known algorithms, \eg, various matrix decomposition algorithms. Regarding data cleaning, the pipeline structure differs. Yet, often the two steps imputation and scaling are implemented. Often single steps in this pipeline could be omitted as the data set is not affected by this specific step, \eg, an imputation without missing values.

By using a pipeline with a fixed structure, the complexity of determining a graph structure \(g\) is eliminated completely and the pipeline creation problem is reduced to selecting a preprocessing and modeling algorithm. Even though this approach greatly reduces the complexity of the pipeline creation problem, it may lead to inferior pipeline performances for complex data sets requiring, for example, multiple preprocessing steps. Yet, for many problems with high quality training data a simple pipeline structure may still be sufficient.

\subsection{Variable Structure}
\begin{figure}
	\centering
	\includegraphics[width=\linewidth]{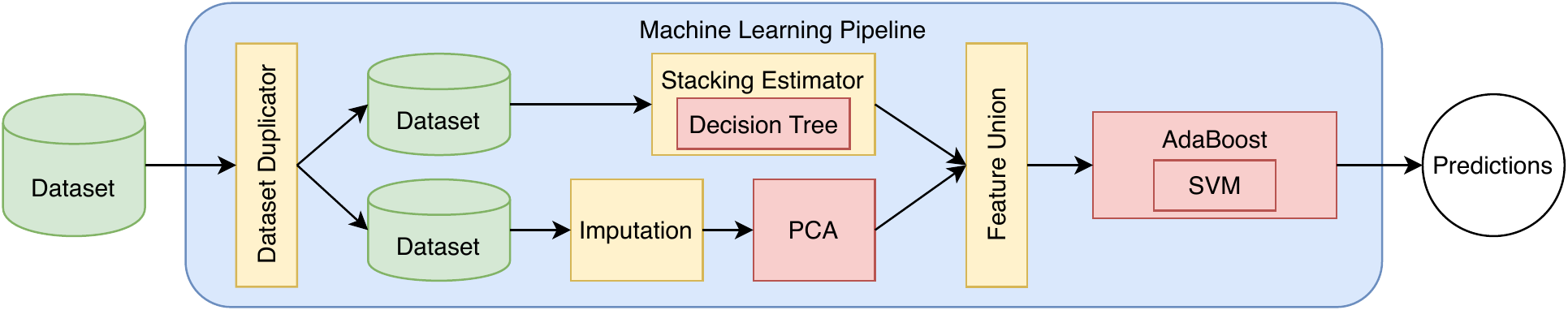}
	\caption{Specialized \ac{ML} pipeline for a specific \ac{ML} task.}
	\label{fig:flexible_ml_pipeline}
\end{figure}

Data science experts usually build highly specialized pipelines for a given \ac{ML} task to obtain the best results. Fixed shaped \ac{ML} pipelines lack this flexibility to adapt to a specific task. Several approaches for building flexible pipelines automatically exist that are all based on the same principal ideas: a pipeline consists of a set of \ac{ML} primitives---namely the basic algorithms \(\mathcal{A}\)---, an \emph{data set duplicator} to clone a data set and a \emph{feature union} operator to combine multiple data sets. The data set duplicator is used to create parallel paths in the pipeline; parallel paths can be joined via a feature union. A pipeline using all these operators is displayed in Figure~\ref{fig:flexible_ml_pipeline}.

The first method to build flexible \ac{ML} pipelines automatically was introduced by \shortciteA{Olson2016} and is based on genetic programming \cite{Koza1992,Banzhaf1997}. Genetic programming has been used for automatic program code generation for a long time \shortcite{Poli2008}. Yet, the application to pipeline structure synthesis is quite recent. Pipelines are interpreted as tree structures that are generated via genetic programming. Two individuals are combined by selecting sub-graphs of the pipeline structures and combining these sub-graphs to a new graph. Mutation is implemented by random addition or deletion of a node. This way, flexible pipelines can be generated.

\acp{HTN} \shortcite{Ghallab2004} are a method from automated planning that recursively partition a complex problem into easier subproblems. These subproblems are again decomposed until only atomic terminal operations are left. This procedure can be visualized as a graph structure. Each node represents a (potentially incomplete) pipeline; each edge the decomposition of a complex step into sub-steps. When all complex problems are replaced by \ac{ML} primitives, an \ac{ML} pipeline is obtained. Using this abstraction, the problem of finding an \ac{ML} pipeline structure is reduced to finding the best leaf node in the graph \shortcite{Mohr2018}.

Monte-Carlo tree search \shortcite{Kocsis2006,Browne2012} is a heuristic best-first tree search algorithm. Similar to hierarchical planning, \ac{ML} pipeline structure generation is reduced to finding the best node in the search tree. However, instead of decomposing complex tasks, pipelines with increasing complexity are created iteratively \shortcite{Rakotoarison2019}.

Self-play \shortcite{Lake2017} is a reinforcement learning strategy that has received a lot of attention lately due to the recent successes of \name{AlphaZero} \shortcite{Silver2017}. Instead of learning from a fixed data set, the algorithm creates new training examples by playing against itself. Pipeline structure search can also be considered as a game \shortcite{Drori2018}: an \ac{ML} pipeline and the training data set represent the current board state \(s\); for each step the player can choose between the three actions adding, removing or replacing a single node in the pipeline; the loss of the pipeline is used as a score \(\nu(s)\). In an iterative procedure, a neural network in combination with Monte-Carlo tree search is used to select a pipeline structure \(g\) by predicting its performance and probabilities which action to chose in this state \shortcite{Drori2018}.

Methods for variable-shaped pipeline construction often do not consider dependencies between different pipeline stages and constraints on the complete pipeline. For example, genetic programming could create a pipeline for a classification task without any classification algorithm \shortcite{Olson2016a}. To prevent such defective pipelines, the pipeline creation can be restricted by a grammar \shortcite{DeSa2017,Drori2019}. In doing so, reasonable but still flexible pipelines can be created.

\section{Algorithm Selection and Hyperparameter Optimization}
\label{sec:cash}

Let a structure \(g \in G\), a loss function \(\mathcal{L}\) and a training set \(D\) be given. For each node in \(g\) an algorithm has to be selected and configured via hyperparameters. This section introduces various methods for algorithm selection and configuration.

A notion first introduced by \shortciteA{Thornton2013} and since then adopted by many others is the \ac{CASH} problem. Instead of selecting an algorithm first and optimizing its hyperparameters later, both steps are executed simultaneously. This problem is formulated as a black box optimization problem leading to a minimization problem quite similar to the pipeline creation problem in Equation~\eqref{eq:pcp}. For readability, assume \(|g| = 1\). The \ac{CASH} problem is defined as 
\begin{equation*}
	(\vec{A}, \vec{\lambda})^\star \in \argmin_{\vec{A} \in \mathcal{A}, \vec{\lambda} \in \Lambda} R \left(\mathcal{P}_{g, \vec{A}, \vec{\lambda}, D}, D \right) .
\end{equation*}

Let the choice which algorithm to use be treated as an additional categorical meta-hyperparameter \(\lambda_r\). Then the complete hyperparameter space for a single algorithm can be defined as
\begin{equation*}
	\Lambda = \Lambda_{A^{(1)}} \times \dots \Lambda_{A^{(n)}} \times \lambda_r
\end{equation*}
referred to as the \emph{configuration space}. This leads to the final \ac{CASH} minimization problem
\begin{equation}
\label{eq:cash}
	\vec{\lambda}^\star \in \argmin_{\vec{\lambda} \in \Lambda} R \left(\mathcal{P}_{g, \vec{\lambda}, D}, D \right) .
\end{equation}
This definition can be easily extended for \(|g| > 1\) by introducing a distinct \(\lambda_r\) for each node. For readability, let \( f( \vec{\lambda} ) = R \left(\mathcal{P}_{g, \vec{\lambda}, D}, D \right) \) be denoted as the \emph{objective function}.

It is important to note that Equation~\eqref{eq:cash} is not easily solvable as the search space is quite large and complex. As hyperparameters can be categorical and real-valued, Equation~\eqref{eq:cash} is a mixed-integer nonlinear optimization problem \shortcite{Belotti2013}. Furthermore, conditional dependencies between different hyperparameters exist. If for example the \(i\)th algorithm is selected, only \(\Lambda_{A^{(i)}}\) is relevant as all other hyperparameters do not influence the result. Therefore, \(\Lambda_{A^{(i)}}\) depends on \(\lambda_r = i\). Following \shortciteA{Hutter2009,Thornton2013,Swearingen2017} the hyperparameters \( \vec{\lambda} \in \Lambda_{A^{(i)}}\) can be aggregated in two groups: mandatory hyperparameters always have to be present while conditional hyperparameters depend on the selected value of another hyperparameter. A hyperparameter \(\lambda_i\) is conditional on another hyperparameter \(\lambda_j\), if and only if \(\lambda_i\) is relevant when \(\lambda_j\) takes values from a specific set \(V_i(j) \subset \Lambda_j\).

\begin{figure}
	\centering
	\includegraphics[width=0.9\linewidth]{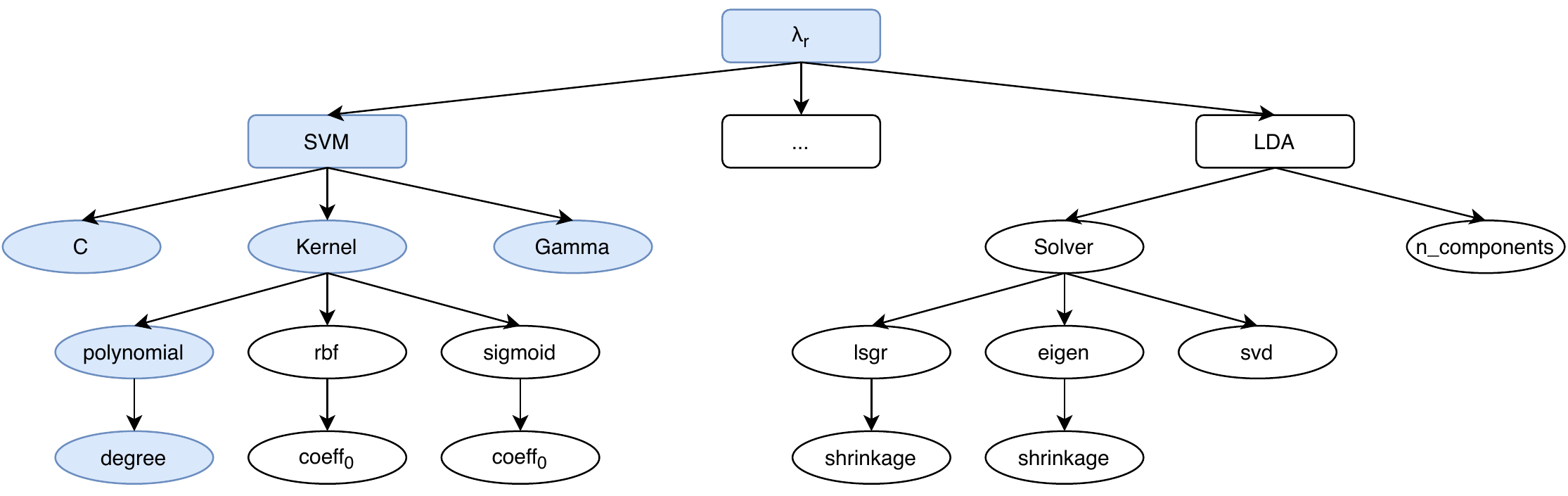}
	\caption{Incomplete representation of the structured configuration space for selecting and tuning a classification algorithm. Rectangle nodes represent the selection of an algorithm. Ellipse nodes represent tunable hyperparameters. Highlighted in blue is an active configuration to select and configure a \ac{SVM} with a polynomial kernel.}
	\label{fig:configuration_space}
\end{figure}

Using this notation, the configuration space can be interpreted as a tree as visualized in Figure~\ref{fig:configuration_space}. \(\lambda_r\) represents the root node with a child node for each algorithm. Each algorithm has the according mandatory hyperparameters as child nodes, all conditional hyperparameters are children of another hyperparameter. This tree structure can be used to significantly reduce the search space.

The rest of this section introduces different optimization strategies to solve Equation~\eqref{eq:cash}.

\subsection{Grid Search}
\label{sec:theory_grid_search}
The first approach to explore the configuration space systematically was grid search. As the name implies, grid search creates a grid of configurations and evaluates all of them. Even though grid search is easy to implement and parallelize \shortcite{BergstraJames2012}, it has two major drawbacks:
\begin{enumerate*}[label={\arabic*)}]
	\item it does not scale well for large configuration spaces, as the number of function evaluations grows exponentially with the number of hyperparameters \shortcite{LaValle2004} and
	\item the hierarchical hyperparameter structure is not considered, leading to many redundant configurations.
\end{enumerate*}

In the traditional version, grid search does not exploit knowledge of well performing regions. This drawback is partially eliminated by \emph{contracting} grid search \shortcite{Hsu2003,Hesterman2010}. At first, a coarse grid is fitted, next a finer grid is created centered around the best performing configuration. This iterative procedure is repeated \(k\) times converging to a local minimum.

\subsection{Random Search}
Another widely-known approach is random search \shortcite{Anderson1953}. A candidate configuration is generated by choosing a value for each hyperparameter randomly and independently of all others. Conditional hyperparameters can be handled implicitly by traversing the hierarchical dependency graph. Random search is straightforward to implement and parallelize and well suited for gradient-free functions with many local minima \shortcite{Solis1981}. Even though the convergence speed is faster than grid search \shortcite{BergstraJames2012}, still many function evaluations are necessary as no knowledge of well performing regions is exploited. As function evaluations are very expensive, random search requires a long optimization period.

\subsection{Sequential Model-based Optimization}
\label{sec:smbo}
\begin{figure}
	\centering
	\includegraphics[width=0.5\linewidth]{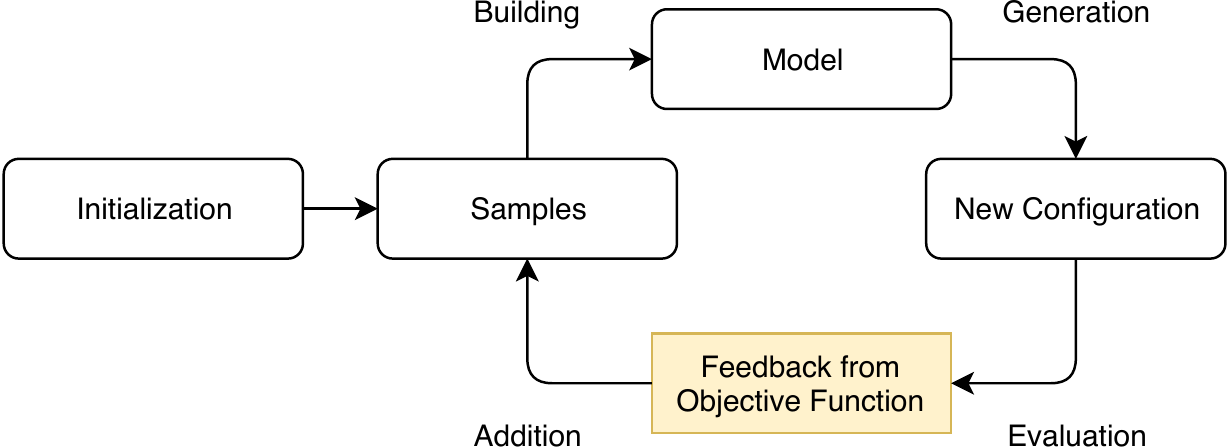}
	\caption{Schematic procedure of \ac{SMBO}.}
	\label{fig:smbo}
\end{figure}

The \ac{CASH} problem can be treated as a regression problem: \(f( \vec{\lambda} )\) can be approximated using standard regression methods based on the so-far tested hyperparameter configurations \(D_{1:n} = \left\{ \left( \vec{\lambda}_{1}, f( \vec{\lambda}_{1}) \right), \dots, \left( \vec{\lambda}_{n}, f( \vec{\lambda}_{n}) \right) \right\}\). This concept is captured by \ac{SMBO} \shortcite{Bergstra2011,Hutter2011,Bergstra2013} displayed in Figure~\ref{fig:smbo}.

The loss function is complemented by a probabilistic regression model \(M\) that acts as a surrogate for \(f\). The surrogate model \(M\), build using \(D_{1:n}\), allows predicting the performance of an arbitrary configuration \( \vec{\lambda} \) without evaluating the demanding objective function. A new configuration \(\vec{\lambda}_{n + 1} \in \Lambda\), obtained using a cheap acquisition function, is evaluated on the objective function \(f\) and the result added to \(D_{1:n}\). These steps are repeated until a fixed budget \(T\)---usually either a fixed number of iterations or a time limit---is exhausted. The initialization is often implemented by selecting a small number of random configurations.

Even though fitting a model and selecting a configuration introduces a computational overhead, the probability of testing badly performing configurations can be lowered significantly. As the actual function evaluation is usually way more expensive than these additional steps, better performing configurations can be found in a shorter time span in comparison to random or grid search.

To actually implement the surrogate model fitting and configuration selection, Bayesian optimization \shortcite{Brochu2010,Shahriari2016,Frazier2018} is used. It is an iterative optimization framework being well suited for expensive objective functions. A probabilistic model of the objective function \(f\) is obtained using Bayes' theorem
\begin{equation}
\label{eq:bayesian_optimisation}
	P\left(f \given D_{1:n} \right) \propto P \left(D_{1:n} \given f \right) P(f) .
\end{equation}
Bayesian optimization is very efficient concerning the number of objective function evaluations \shortcite{Brochu2010} as the acquisition function handles the trade-off between exploration and exploitation automatically. New regions with a high uncertainty are explored, preventing the optimization from being stuck in a local minimum. Well performing regions with a low uncertainty are exploited converging to a local minimum \shortcite{Brochu2010}. The surrogate model \(M\) corresponds to the posterior in Equation~\eqref{eq:bayesian_optimisation}. As the characteristics and shape of the loss function are in general unknown, the posterior has to be a non-parametric model.

The traditional surrogate models for Bayesian optimization are Gaussian processes \shortcite{Rasmussen2006}. The key idea is that any objective function \(f\) can be modeled using an infinite dimensional Gaussian distribution. A common drawback of Gaussian processes is the runtime complexity of \(\mathcal{O} (n^3)\) \shortcite{Rasmussen2006}. However, as long as multi-fidelity methods (see Section~\ref{sec:performance_improvements}) are not used, this is not relevant for \ac{AutoML} as evaluating a high number of configurations is prohibitively expensive. A more relevant drawback for \ac{CASH} is the missing native support of categorical input\footnote{
	Extensions for treating integer variables in Gaussian processes exist (\eg, \shortciteR{Levesque2017,Garrido-Merchan2018}).
} and utilization of the search space structure.

Random forest regression \shortcite{Breiman2001} is an ensemble method consisting of multiple regression trees \shortcite{Breiman1984}. Regression trees use recursive splitting of the training data to create groups of similar observations. Besides the ability to handle categorical variables natively, random forests are fast to train and even faster on evaluating new data while obtaining a good predictive power.

In contrast to the two previous surrogate models, a \ac{TPE} \shortcite{Bergstra2011} models the likelihood \(P(D_{1:n} \given f)\) instead of the posterior. Using a performance threshold \(f'\), all observed configurations are split into a well and badly performing set, respectively. Using \ac{KDE} \shortcite{Parzen1961}, those sets are transformed into two distributions. Regarding the tree structure, \acp{TPE} handle hierarchical search spaces natively by modeling each hyperparameter individually. These distributions are connected hierarchically representing the dependencies between the hyperparameters resulting in a pseudo multidimensional distribution.

\subsection{Evolutionary Algorithms}
An alternative to \ac{SMBO} are evolutionary algorithms \shortcite{Coello2007}. Evolutionary algorithms are a collection of various population-based optimization algorithms inspired by biological evolution. In general, evolutionary algorithms are applicable to a wide variety of optimization problems as no assumptions about the objective function are necessary.

\shortciteA{Escalante2009} and \shortciteA{Claesen2014} perform hyperparameter optimization using a particle swarm \shortcite{Reynolds1987}. Originally developed to simulate simple social behavior of individuals in a swarm, particle swarms can also be used as an optimizer \shortcite{Kennedy1995}. Inherently, a particle's position and velocity are defined by continuous vectors \(\vec{x}_i, \vec{v}_i \in \mathbb{R}^d\). Similar to Gaussian processes, all categorical and integer hyperparameters have to be mapped to continuous variables introducing a mapping error.

\subsection{Multi-armed Bandit Learning}
\label{sec:bandit_learning}
Many \ac{SMBO} methods suffer from the mixed and hierarchical search space. By performing grid search considering only the categorical hyperparameters, the configuration space can be split into a finite set of smaller configuration spaces---called a \emph{hyperpartition}---containing only continuous hyperparameters. Each hyperpartition can be optimized by standard Bayesian optimization methods. The selection of a hyperpartition can be modeled as a \emph{multi-armed bandit problem} \shortcite{Robbins1952}. Even though multi-armed bandit learning can also be applied to continuous optimization \shortcite{Munos2014}, in the context of \ac{AutoML} it is only used in a finite setting in combination with other optimization techniques \shortcite{Hoffman2014,Efimova2017,Gustafson2018,Dores2018}.

\subsection{Gradient Descent}
A very powerful optimization method is \emph{gradient descent}, an iterative minimization algorithm. If \(f\) is differentiable and its closed-form representation is known, the gradient \(\nabla f\) is computable. However, for \ac{CASH} the closed-form representation of \(f\) is not known and therefore gradient descent in general not applicable. By assuming some properties of \(f\)---and therefore limiting the applicability of this approach to specific problem instances---gradient descent can still be used \shortcite{Maclaurin2015,Pedregosa2016}. Due to the rigid constraints, gradient descent is not analyzed in more detail.

\section{Automatic Data Cleaning}
\label{sec:data_cleaning}
Data cleaning is an important aspect of building an \ac{ML} pipeline. The purpose of data cleaning is to improve the quality of a data set by removing data errors. Common error classes are missing values in the input data, redundant entries, invalid values or broken links between entries of multiple data sets \shortcite{Do2000}. In general, data cleaning is split into two tasks: error detection and error repairing \shortcite{Chu2016}. For over two decades semi-automatic, interactive systems existed to aid a data scientist in data cleaning \shortcite{Galhardas2000,Raman2001}. Yet, most current approaches still aim to assist a human data scientist instead of fully automated data cleaning, (\eg, \shortciteR{Krishnan2015,Khayyaty2015,Krishnan2016,Eduardo2016,Rekatsinas2017}). \shortciteA{Krishnan2019} proposed an automatic data cleaning procedure with minimal human interaction: based on a human defined \emph{data quality} function, data cleaning is treated similarly to pipeline structure search. Basic data cleaning operators are combined iteratively using greedy search to create sophisticated data cleaning.

Most existing \ac{AutoML} frameworks recognize the importance of data cleaning and include various data cleaning stages in the \ac{ML} pipeline (\eg, \shortciteR{Feurer2015,Swearingen2017,Parry2019}). However, these data cleaning steps are usually hard-coded and not generated based on some metric during an optimization period. These fixed data cleaning steps usually contain imputation of missing values, removing of samples with incorrect values, like infinity or outliers, and scaling features to a normalized range. In general, current \ac{AutoML} frameworks do not consider state-of-the-art data cleaning methods.

Sometimes, high requirements for specific data qualities are introduced by later stages in an \ac{ML} pipeline, \eg, \acp{SVM} require a numerical encoding of categorical features while random forests can handle them natively. These additional requirements can be detected by analyzing a candidate pipeline and matching the prerequisites of every stage with meta-features of each feature in the data set \shortcite{Gil2018,Nguyen2020}.

Incorporating domain knowledge during data cleaning increases the data quality significantly \shortcite{Jeffery2006,Messaoud2011,Salvador2016}. Using different representations of expert knowledge, like integrity constraints or first order logic, low quality data can be detected and corrected automatically \shortcite{Raman2001,Hellerstein2008,Chu2015,Chu2016}. However, these potentials are not used by current \ac{AutoML} frameworks as they aim to be completely data-agnostic to be applicable to a wide range of data sets. Advanced and domain specific data cleaning is conferred to the user.

\section{Automatic Feature Engineering}
\label{sec:feature_engineering}
Feature engineering is the process of generating and selecting features from a given data set for the subsequent modeling step. This step is crucial for the \ac{ML} pipeline, as the overall model performance highly depends on the available features. By building good features, the performance of an \ac{ML} pipeline can be increased many times over an identical pipeline without dedicated feature engineering \shortcite{Pyle1999}. Feature engineering can be split into three sub-tasks: feature extraction, feature construction and feature selection \shortcite{Motoda2002}. Feature engineering---especially feature construction---is highly domain specific and difficult to generalize. Even for data scientists assessing the impact of a feature is difficult, as domain knowledge is necessary. Consequently, feature engineering is a mainly manual and time-consuming task driven by trial and error. In the context of \ac{AutoML}, feature extraction and feature construction are usually aggregated as feature generation.

\subsection{Feature Generation}

Feature generation creates new features through a functional mapping of the original features (feature extraction) or discovering missing relationships between the original features (feature creation) \shortcite{Motoda2002}. In general, this step requires the most domain knowledge and is therefore the hardest to automate. Approaches to enhance automatic feature generation with domain knowledge (\eg, \shortciteR{Friedman2015,Smith2017}) are not considered as \ac{AutoML} aims to be domain-agnostic. Still, some features---like dates or addresses---can be transformed easily without domain knowledge to extract more meaningful features \shortcite{Chen2018}.

\begin{figure}[ht]
	\centering
	\includegraphics[width=0.6\linewidth]{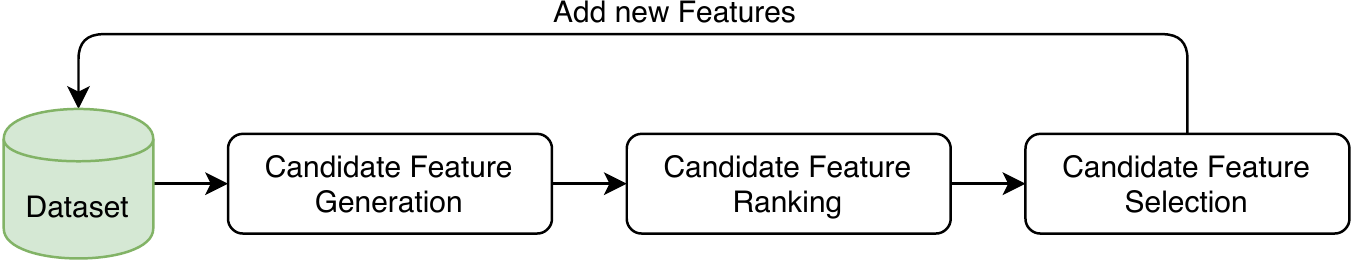}
	\caption{Iterative feature generation procedure.}
	\label{fig:feature_generation}
\end{figure}

Basically all automatic feature generation approaches follow the iterative schema displayed in Figure~\ref{fig:feature_generation}. Based on an initial data set, a set of candidate features is generated and ranked. Highly ranked features are evaluated and added to the data set potentially. These three steps are repeated several times.

New features are generated using a predefined set of operators transforming the original features \shortcite{Sondhi2009}:
\begin{description}
	\item[Unary] Unary operators transform a single feature, for example by discretizing or normalizing numerical features, applying rule-based expansions of dates or using unary mathematical operators like a logarithm.
	\item[Binary] Binary operators combine two features, \eg, via basic arithmetic operations. Using correlation tests and regression models, the correlation between two features can be expressed as a new feature \shortcite{Kaul2017}.
	\item[High-Order] High-order operators are usually build around the SQL \emph{Group By} operator: all records are grouped by one feature and then aggregated via minimum, maximum, average or count.
\end{description}
Similar to pipeline structure search, feature generation can be considered as a node selection problem in a \emph{transformation tree}: the root node represents the original features; each edge applies one specific operator leading to a transformed feature set \shortcite{Khurana2016,Lam2017}.

Many approaches augment feature selection with an \ac{ML} model to actually calculate the performance of the new feature set. Early approaches combined beam search in combination with different heuristics to explore the feature space in a best-first way \shortcite{Markovitch2002}. More recently, greedy search \shortcite{Dor2012,Khurana2016} and depth-first search \shortcite{Lam2017} in combination with feature selection have been used to create a sequence of operators. In each iteration, a random operation is applied to the currently best-performing data set until the performance improvement does converge. Another popular approach is combining features using genetic programming \shortcite{Smith2005,Tran2016}.

Instead of exploring the transformation tree iteratively, exhaustive approaches consider a fully expanded transformation tree up to a predefined depth \shortcite{Kanter2015,Katz2017}. Most of the candidate features do not contain meaningful information. Consequently, the set of candidate features has to be filtered. Yet, generating exponentially many features makes this approach prohibitively expensive in combination with an \ac{ML} model. Instead, the new features can be filtered without an actual evaluation (see Section~\ref{sec:feature_selection}) or ranked based on meta-features (see Section~\ref{sec:meta_learning}). Based on the meta-features of a candidate feature, the expected loss reduction after including this candidate can be predicted using a regression model \shortcite{Katz2017,Nargesian2017}, reinforcement learning \shortcite{Khurana2018} or stability selection \shortcite{Kaul2017}. The predictive model is created in an offline training phase. Finally, candidate features are selected by their ranking and the best features are added to the data set.

Some frameworks specialize on feature generation in relational databases \shortcite{Kanter2015,Lam2017}. \shortciteA{Wistuba2017} and \shortciteA{Chen2018} propose using stacked estimators. The predicted output is added as an additional feature such that later estimators can correct wrongly labeled data. Finally, \shortciteA{Khurana2018a} proposed to create an ensemble of sub-optimal feature sets (see Section~\ref{sec:ensemble_learning}).

Another approach for automatic feature generation is \emph{representation learning} \shortcite{Bengio2013,Goodfellow2016}. Representation learning aims to transform the input data into a latent representation space well suited for a---in the context of this survey---supervised learning task automatically. As this approach is usually used in combination with neural networks and unstructured data, it is not further evaluated.

\subsection{Feature Selection}
\label{sec:feature_selection}
Feature selection chooses a subset of the feature set to speed up the subsequent \ac{ML} model training and to improve its performance by removing redundant or misleading features \shortcite{Motoda2002}. Furthermore, the interpretability of the trained model is increased. Simple domain-agnostic filtering approaches for feature selection are based on information theory and statistics \shortcite{Pudil1994,Yang1997,Dash1997,Guyon2003}. Algorithms like univariate selection, variance threshold, feature importance, correlation matrices \shortcite{Saeys2007} or stability selection \shortcite{Meinshausen2010} are already integrated in modern \ac{AutoML} frameworks \shortcite{Thornton2013,Komer2014,Feurer2015,Olson2016,Swearingen2017,Parry2019} and selected via standard \ac{CASH} methods. More advanced feature selection methods are usually implemented in dedicated feature engineering frameworks.

In general, the feature set---and consequently also its power set---is finite. Feature selection via \emph{wrapper functions} searches for the best feature subset by testing its performance on a specific \ac{ML} algorithm. Simple approaches use random search or test the power set exhaustively \shortcite{Dash1997}. Heuristic approaches follow an iterative procedure by adding single features \shortcite{Kononenko1994}. \shortciteA{Margaritis2009} used a combination of forward and backward selection to select a feature-subset while \shortciteA{Gaudel2010} proposed to model the subset selection as a reinforcement problem. \shortciteA{Vafaie1992} used genetic programming in combination with a cheap prediction algorithm to obtain a well performing feature subset.

Finally, \emph{embedded} methods incorporate feature selection directly into the training process of an \ac{ML} model. Many \ac{ML} models provide some sort of feature ranking that can be utilized, \eg, \acp{SVM} \shortcite{Guyon2002,Rakotomamonjy2003}, perceptrons \shortcite{Mejia-Lavalle2006} or random forests \shortcite{Tuv2009}. Similarly, embedded methods can be used in combination with feature extraction and feature creation. \shortciteA{Tran2016} used genetic programming to construct new features. In addition, the information how often each feature was used during feature construction is re-used to obtain a feature importance. \shortciteA{Katz2017} proposed to calculate meta-features for each new feature, \eg, diversity of values or mutual information with the other features. Using a pre-trained classifier, the influence of a single feature can be predicted to select only promising features.

\section{Performance Improvements}
\label{sec:performance_improvements}
In the previous sections, various techniques for building an \ac{ML} pipeline have been presented. In this section, different performance improvements are introduced. These improvements cover multiple techniques to speed up the optimization procedure as well as improving the overall performance of the generated \ac{ML} pipeline.

\subsection{Multi-fidelity Approximations}
\label{sec:multi-fidelity}
The major problem for \ac{AutoML} and \ac{CASH} procedures is the extremely high turnaround time. Depending on the used data set, fitting a single model can take several hours, in extreme cases even up to several days \shortcite{Krizhevsky2012}. Consequently, optimization progress is very slow. A common approach to circumvent this limitation is the usage of multi-fidelity approximations \shortcite{Fernandez-Godino2016}. Data scientist often use only a subset of the training data or a subset of the available features \shortcite{Bottou2012}. By testing a configuration on this training subset, badly performing configurations can be discarded quickly and only well performing configurations have to be tested on the complete training set. The methods presented in this section aim to mimic this manual procedure to make it applicable for fully automated \ac{ML}.

A straight-forward approach to mimic expert behavior is choosing multiple random subsets of the training data for performance evaluation \shortcite{Nickson2014}. More sophisticated methods augment the black box optimization in Equation~\eqref{eq:pcp} by introducing an additional budget term \(s \in [0, 1]\) that can be freely selected by the optimization algorithm.

\begin{figure}
	\centering
	\includegraphics[width=0.6\linewidth]{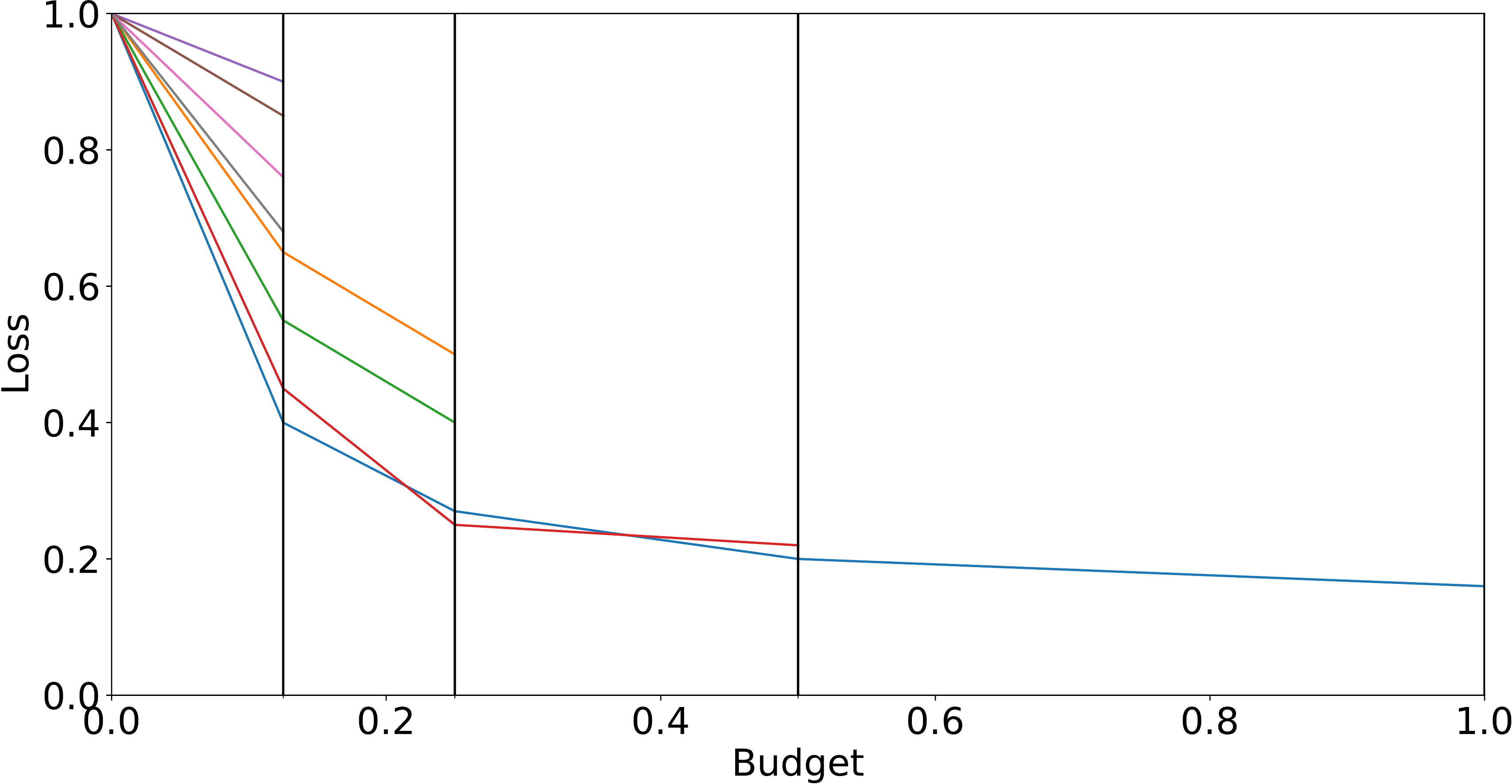}
	\caption{Schematic representation of \name{SuccessiveHalving} with 8 different configurations.}
	\label{fig:successive-halving}
\end{figure}

\name{SuccessiveHalving} \shortcite{Jamieson2016} solves the selection of \(s\) via bandit learning. The basic idea, as visualized in Figure~\ref{fig:successive-halving}, is simple: \name{SuccessiveHalving} randomly creates \(m\) configurations and tests each for the partial budget \(s_{0} = 1 / m\). The better half is transferred to the next iteration allocating twice the budget to evaluate each remaining configuration. This procedure is repeated until only one configuration remains \shortcite{Hutter2018a}. A crucial problem with \name{SuccessiveHalving} is the selection of \(m\) for a fixed budget: is it better to test many different configurations with a low budget or only a few configurations with a high budget?

\name{Hyperband} \shortcite{Li2016,Li2018} answers this question by selecting an appropriate number of configurations dynamically. It calculates the number of configurations and budget size based on some budget constraints. A descending sequence of configuration numbers \(m\) is calculated and passed to \name{SuccessiveHalving}. Consequently, no prior knowledge is required anymore for \name{SuccessiveHalving}.

\name{Fabolas} \shortcite{Klein2016} treats the budget \(s\) as an additional input parameter in the search space that can be freely chosen by the optimization procedure instead of being deterministically calculated. A Gaussian process is trained on the combined input \((\vec{\lambda}, s)\). Additionally, the acquisition function is enhanced by entropy search \shortcite{Hennig2012}. This allows predicting the performance of \(\vec{\lambda}_i\), tested with budget \(s_i\), for the full budget \(s = 1\).

It is important to note that all presented methods usually generate a budget in a fixed interval \([a, b]\) and the actual interpretation of this budget is conferred to the user. For instance, \name{Hyperband} and \name{SuccessiveHalving} have been used very successfully to select the number of training epochs in neural networks. Consequently, multi-fidelity approximations can be used for many problem instances.

\subsection{Early Stopping}
In contrast to using only a subset of the training data, several methods have been proposed to terminate the evaluation of unpromising configurations early. Many existing \ac{AutoML} frameworks (see Section~\ref{sec:selected_frameworks}) incorporate \(k\)-fold cross-validation to limit the effects of overfitting. A quite simple approximation is to abort the fitting after the first fold if the performance is significantly worse than the current incumbent \shortcite{Maron1993,Hutter2011}.

The training of an \ac{ML} model is often an iterative procedure converging to a local minimum. By observing the improvement in each iteration, the learning curve of an \ac{ML} model can be predicted \shortcite{Domhan2015,Klein2017a}. This allows discarding probably bad performing configurations without a complete training. By considering multiple configurations in an iterative procedure simultaneously, the most promising configuration can be optimized in each step \shortcite{Swersky2014}.

In non-deterministic scenarios, configurations usually have to be evaluated on multiple problem instances to obtain reliable performance measures. Some of these problem instances may be very unfavorable leading to drawn-out optimization periods \shortcite{Huberman1997}. By evaluating multiple problem instances in parallel, long running instances can be discarded early \shortcite{Weisz2018a,Li2020}.

\subsection{Scalability}
As previously mentioned, fitting an \ac{ML} pipeline is a time consuming and computational expensive task. A common strategy for solving a computational heavy problem is parallelization on multiple cores or within a cluster (\eg, \shortciteR{Buyya1999,Dean2008}). \name{scikit-learn} \shortcite{Pedregosa2011}, which is used by most evaluated frameworks (see Section~\ref{sec:selected_frameworks}), already implements optimizations to distribute workload on multiple cores on a single machine. As \ac{AutoML} normally has to fit many \ac{ML} models, distributing different fitting instances in a cluster is an obvious idea.

Most of the previously mentioned methods allow easy parallelization of single evaluations. Using grid search and random search, pipeline instances can be sampled independently of each other. Evolutionary algorithms allow a simultaneous evaluation of candidates in the same generation. However, \ac{SMBO} is---as the name already implies---a sequential procedure.

\ac{SMBO} procedures often contain a randomized component. Executing multiple \ac{SMBO} instances with different random seeds allows a simple parallelization \shortcite{Hutter2012}. However, this simple approach often does not allow sharing knowledge between the different instances. Alternatively, the surrogate model \(M\) can be handled by a single \textit{coordinator} while the evaluation of candidates is distributed to several \textit{workers}. Pending candidate evaluations can be either ignored---if sampling a new candidate depends on a stochastic process \shortcite{Bergstra2011,Kandasamy2018}--- or imputed with a constant \shortcite{Ginsbourger2010} or simulated performance \shortcite{Ginsbourger2010a,Snoek2012,Desautels2014}. This way, new configurations can be sampled from an approximated posterior while preventing the evaluation of the same configuration twice.

\begin{figure}
	\centering
	\includegraphics[width=0.5\linewidth]{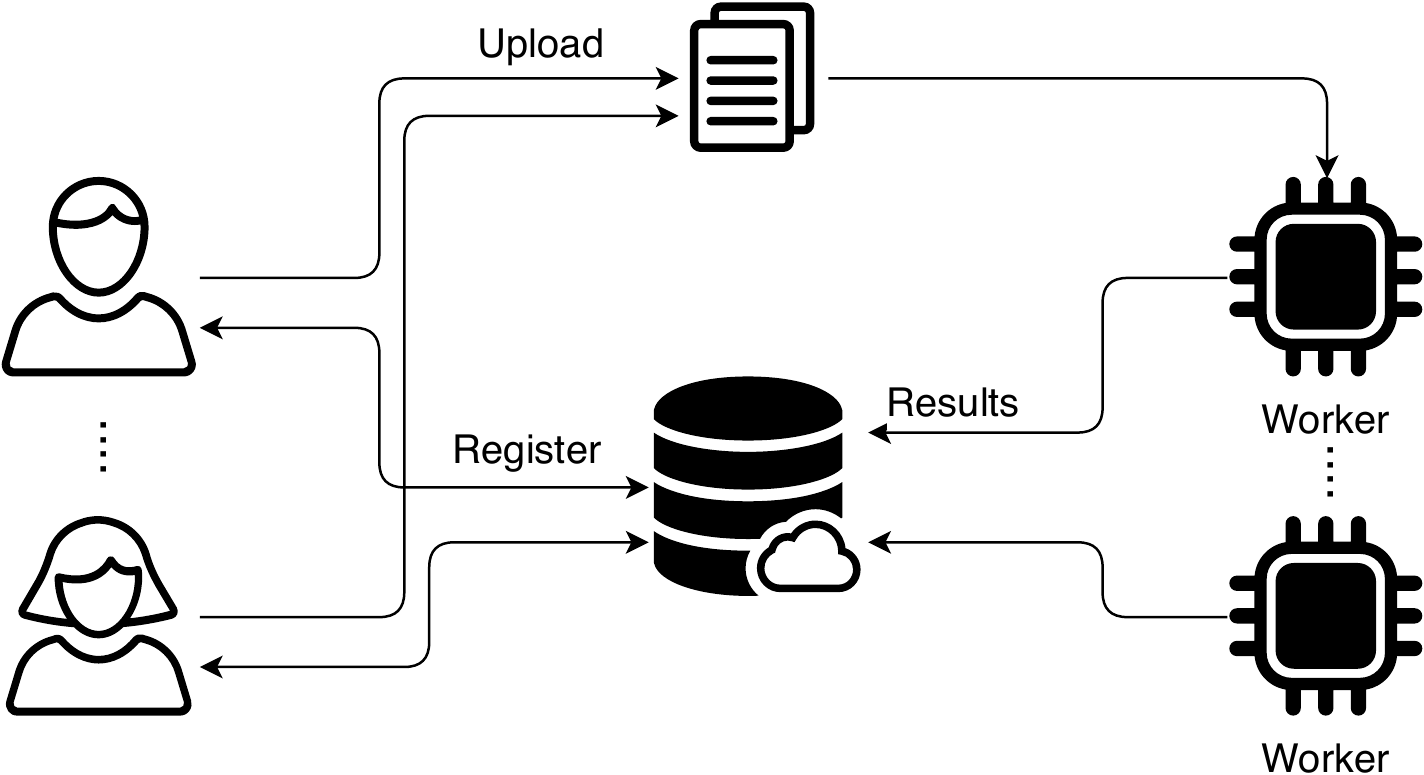}
	\caption{Components of an \ac{AutoML} service \shortcite{Swearingen2017}.}
	\label{fig:automl-aa-service}
\end{figure}

The scaling of \ac{AutoML} tasks to a cluster also allows the introduction of \ac{AutoML} services. Users can upload their data set and configuration space---called a \emph{study}---to a persistent storage. Workers in a cluster test different configurations of a study until a budget is exhausted. This procedure is displayed in Figure~\ref{fig:automl-aa-service}. As a result, users can obtain optimized \ac{ML} pipelines with minimal effort in a short timespan.

Various open-source designs for \ac{AutoML} services have been proposed (\eg, \shortciteR{Sparks2015,Chan2017,Swearingen2017,Koch2018}) but also several commercial solutions exist (\eg, \shortciteR{Golovin2017,Clouder2018,H2O.ai2018}). Some commercial solutions also focus on providing \ac{ML} without the need to write own code, enabling domain experts without programming skills to create optimized \ac{ML} workflows \shortcite{USU2018,Baidu2018,RapidMiner2018}.

\subsection{Ensemble Learning}
\label{sec:ensemble_learning}
A well-known concept in \ac{ML} is ensemble learning \shortcite{Opitz1999,Polikar2006,Rokach2010}. Ensemble methods combine multiple \ac{ML} models to create predictions. Depending on the diversity of the combined models, the overall accuracy of the predictions can be increased significantly. The cost of evaluating multiple \ac{ML} models is often neglectable considering the performance improvements.

During the search of a well performing \ac{ML} pipeline, \ac{AutoML} frameworks create a large number of different pipelines. Instead of only yielding the best performing configuration, the set of best performing configurations can be used to create an ensemble \shortcite{Lacoste2014,Feurer2015,Wistuba2017}. Similarly, automatic feature engineering often creates several different candidate data sets \shortcite{Khurana2016,Katz2017,Nargesian2017}. By using multiple data sets, various \ac{ML} pipelines can be constructed \shortcite{Khurana2018a}.

An interesting approach for ensemble learning is \emph{stacking} \shortcite{Wolpert1992}. A stacked \ac{ML} pipeline is generated in multiple layers, each layer being a \emph{normal} \ac{ML} pipeline. The predicted output of each previous layer is appended as a new feature to the training data of subsequent layers. This way, later layers have the chance to correct wrong predictions of previous layers \shortcite{Wistuba2017,Khurana2018a,Chen2018}.

\subsection{Meta-learning}
\label{sec:meta_learning}
Given a new unknown \ac{ML} task, \ac{AutoML} methods usually start from scratch to build an \ac{ML} pipeline. However, a human data scientist does not always start all over again but learns from previous tasks. Meta-learning is the science of learning how \ac{ML} algorithms learn. Based on the observation of various configurations on previous \ac{ML} tasks, meta-learning builds a model to construct promising configurations for a new unknown \ac{ML} task leading to faster convergence with less trial and error. \shortciteA{Vanschoren2019} provides a survey exclusively on meta-learning.

Meta-learning can be used in multiple stages of building an \ac{ML} pipeline automatically to increase the efficiency:

\paragraph{Search Space Refinements}
All presented \ac{CASH} methods require an underlying search space definition. Often these search spaces are chosen arbitrarily without any validation leading to either bloated spaces or spaces missing well-performing regions. In both cases the \ac{AutoML} procedure is unable to find optimal results. Meta-learning can be used to assess the importance of single hyperparameters allowing to remove unimportant hyperparameters from the configuration space \shortcite{Hutter2014,Wistuba2015,Rijn2018,Probst2019} or identify promising regions \shortcite{Wistuba2015a}. \shortciteA{Perrone2019} use transfer learning to automatically construct a minimal search space from the best configurations on related \ac{ML} tasks.

\paragraph{Candidate Configuration Suggestion} Many \ac{AutoML} procedures generate candidate configurations by selecting the configuration with the highest expected improvement. Meta-learning can be used as an additional criterion for selecting promising candidate configurations based on the predicted performance (\eg, \shortciteR{Alia2006,Wistuba2015a,Nargesian2017}) or ranking of the models (\eg, \shortciteR{Sohn1999,Gama2000}). Consequently, the risk of superfluous configuration evaluations is minimized.

\paragraph{Warm-Starting}
Basically all presented methods have an initialization phase where random configurations are selected. The same methods as for candidate suggestion can be applied to initialization. Warm-starting can also be used for many aspects of \ac{AutoML}, yet most research focuses on model selection and tuning \shortcite{Gomes2012,DeMiranda2012,Reif2012,Feurer2015,Feurer2015a,Wistuba2015a,Lindauer2018}.

\paragraph{Pipeline Structure}
Meta-learning is also applicable for pipeline structure search. \shortciteA{Feurer2015} use meta-features to warm-start the pipeline synthesis. Using information on which preprocessing and model combination performs well, potentially better performing pipelines can be favored \shortcite{Post2016,Bilalli2017,Schoenfeld2018}. \shortciteA{Gil2018} uses meta-features in the context of planning to select promising pipeline structures. Similarly, \shortciteA{Drori2019} and \shortciteA{Rakotoarison2019} use meta-features of the data set and pipeline candidate to predict the performance of the pipeline.

To actually apply meta-learning for any of these areas, \textit{meta-data} about a set of prior evaluations
\begin{equation*}
	\mathbf{P} = \bigcup_{t_j \in T, \vec{\lambda}_i \in \Lambda} R(\vec{\lambda}_i, t_j)~,
\end{equation*}
with \(T\) being the set of all known \ac{ML} tasks, is necessary. Meta-data usually comprises properties of the previous task in combination with the used configuration and resulting model evaluations \shortcite{Vanschoren2019}.

A simple task-independent approach for ranking configurations is sorting \(\mathbf{P}\) by performance. Configurations with higher performance are more favorable \shortcite{Vanschoren2019}. For configurations with similar performance, the training time can be used to prefer faster configurations \shortcite{vanRijn2015}. Yet, ignoring the task can lead to useless recommendations, for example a configuration performing well for a regression task may not be applicable to a classification problem.

An \ac{ML} task \(t_j\) can be described by a vector \(\vec{m}(t_j)\) of meta-features. Meta-features describe the training data set, \eg, number of instances or features, distribution of and correlation between features or measures from information theory. The actual usage of \(\vec{m}(t_j)\) highly depends on the meta-learning technique. For example, using the meta-features of a new task \(\vec{m}(t_{\mathrm{new}})\), a subset of \(\mathbf{P}' \subset \mathbf{P}\) with similar tasks can be obtained. \(\mathbf{P}'\) is then used similarly to task-independent meta-learning \shortcite{Vanschoren2019}.

\section{Selected Frameworks}
\label{sec:selected_frameworks}
This section provides an introduction to the evaluated \ac{AutoML} frameworks. Frameworks were selected based on their popularity, namely the number of citations and GitHub stars. Preferably, the frameworks cover a wide range of the methods presented in Section~\ref{sec:pipeline_structure}--\ref{sec:performance_improvements} without implementing the same approaches multiple times. Finally, all frameworks had to be open source.

Implementations of \ac{CASH} algorithms are presented and analyzed in Section~\ref{sec:cash_algorithms}. Frameworks for creating complete \ac{ML} pipelines are discussed in Section~\ref{sec:automl_algorithms}. In this section, all presented implementations are discussed qualitatively; experimental evaluation is provided in Section~\ref{sec:experiments}. A reference to the source code of each framework is provided in Appendix~\ref{app:source_code}.

\subsection{CASH Algorithms}
\label{sec:cash_algorithms}
At first, popular implementations of methods for algorithm selection and \ac{HPO} are discussed. The mathematical foundation for all discussed implementations was provided in Section~\ref{sec:cash} and Section~\ref{sec:performance_improvements}. A summary including the most important properties is available in Table~\ref{tbl:overview_cash_solver}.

\begin{table}[ht]
\center

\renewcommand{\arraystretch}{1.2}
\begin{tabular}{@{} l @{\hskip 7mm} l l l l l l @{} }
	\toprule
	Algorithm			& Solver						& \(\Lambda\)	& Parallel	& Time	& Cat.	\\
	\midrule
	\name{Dummy}			& --								& no	& no		& no	& no	\\
	\name{Random Forest}	& --								& no	& no		& no	& no	\\	
	
	Grid Search			& Grid Search							& no	& Local		& no	& yes	\\
	Random Search		& Random Search							& no	& Local		& no	& yes	\\
	\name{RoBO}			& \ac{SMBO} with Gaussian process		& no	& no		& no	& no	\\
	\name{BTB}			& Bandit learning and Gaus. process		& yes	& no		& no	& yes	\\
	\name{hyperopt}		& \ac{SMBO} with \ac{TPE}				& yes 	& Cluster	& no	& yes	\\
	\name{SMAC}			& \ac{SMBO} with random forest			& yes	& Local		& yes	& yes	\\
	\name{BOHB}			& Bandit learning and \ac{TPE}			& yes	& Cluster	& yes	& yes	\\
	\name{Optunity}		& Particle Swarm Optimization			& yes	& Local		& no	& no	\\
	\bottomrule
\end{tabular}

\caption{
	Comparison of different \ac{CASH} algorithms. Reported are the used solver, whether the search space structure is considered (\(\Lambda\)), if parallelization is implemented (Parallel), whether a timeout for a single evaluation exists (Time) and if categorical variables are natively supported (Cat.).
}
\label{tbl:overview_cash_solver}

\end{table}

\paragraph{Baseline Methods}
To assess the effectiveness of the different \ac{CASH} algorithms, two baseline methods are used: a dummy classifier and a random forest. The dummy classifier uses stratified sampling to create random predictions. The \name{scikit-learn} \shortcite{Pedregosa2011} implementations with default hyperparameters are used for both methods.

\paragraph{Grid Search}
A custom implementation based on \name{GridSearchCV} from \name{scikit-learn} \shortcite{Pedregosa2011} is used. \name{GridSearchCV} is extended to support algorithm selection via a distinct \name{GridSearchCV} instance for each \ac{ML} algorithm. To ensure fair results, a mechanism for stopping the optimization after a fixed number of iterations has been added.

\paragraph{Random Search}
Similar to grid search, a custom implementation of random search based on the \name{scikit-learn} implementation \name{RandomizedSearchCV} is used. \name{RandomizedSearchCV} is extended to support algorithm selection.

\paragraph{RoBO}
\name{RoBO} \shortcite{Klein2017} is a generic framework for general purpose Bayesian optimization. In the context of this work, \name{RoBO} is configured to use \ac{SMBO} with a Gaussian process as a surrogate model. The hyperparameters of the Gaussian process are tuned automatically using Markov chain Monte Carlo sampling. Categorical hyperparameters are not supported. \name{RoBO} is evaluated in version 0.3.1.

\paragraph{BTB}
\name{BTB} \shortcite{Gustafson2018} combines multi-armed bandit learning with Gaussian processes. Categorical hyperparameters are selected via bandit learning and the remaining continuous hyperparameters are selected via Bayesian optimization. In the context of this work \textit{upper confidence bound} is used as the policy. \name{BTB} is evaluated in version 0.2.5.

\paragraph{Hyperopt}
\name{hyperopt} \shortcite{Bergstra2011} is a \ac{CASH} solver based on \ac{SMBO} with \acp{TPE} as surrogate models. \name{hyperopt} is evaluated in version 0.2.

\paragraph{SMAC}
\name{SMAC} \shortcite{Hutter2011} was the first framework explicitly supporting categorical variables for configuration selection based on \ac{SMBO}, making it especially suited for \ac{CASH}. The performance of all previous configurations is modeled using random forest regression. \name{SMAC} automatically terminates single configuration evaluations after a fixed timespan. This way, very unfavorable configurations are discarded quickly without slowing the complete optimization down. \name{SMAC} is evaluated in version 0.10.0.

\paragraph{BOHB}
\name{BOHB} \shortcite{Falkner2018} combines Bayesian optimization with \name{Hyperband} \shortcite{Li2018} for \ac{CASH} optimization. A limitation of \name{Hyperband} is the random generation of the tested configurations. \name{BOHB} replaces this random selection by a \ac{SMBO} procedure based on \acp{TPE}. For each function evaluation, \name{BOHB} passes the current budget and a configuration instance to the objective function. In the context of this evaluation, the budget is treated as the fraction of training data used for training. \name{BOHB} is evaluated in version 0.7.4.

\paragraph{Optunity}
\name{Optunity} \shortcite{Claesen2014} is a generic framework for \ac{CASH} with a set of different solvers. In the context of this paper, only the particle swarm optimization is used. Based on a heuristic, a suited number of particles and generations is selected for a given number of evaluations. \name{Optunity} is evaluated in version 1.0.0.

\subsection{AutoML Frameworks}
\label{sec:automl_algorithms}
This section presents the \ac{AutoML} frameworks capable of building complete \ac{ML} pipelines based on the methods provided in Section~\ref{sec:pipeline_structure}, \ref{sec:data_cleaning}, and \ref{sec:feature_engineering}. For algorithm selection and \ac{HPO}, implementations from Section~\ref{sec:cash_algorithms} are used. A summary is available in Table~\ref{tbl:overview_automl_frameworks}.

\begin{table}
\center

\renewcommand{\arraystretch}{1.2}
\begin{tabular}{l l l l l l l}
	\toprule
	Framework 					& \ac{CASH} Solver & Structure & Ensem. & Cat. & Parallel & Time \\
	\midrule
	\name{Dummy}					& --					& Fixed		& no	& no	& no			& no 	\\
	\name{Random Forest}		& --					& Fixed		& no	& no	& no			& no	\\
	
	\name{TPOT}					& Genetic Prog.		& Variable	& no	& no	& Local		& yes	\\
	\name{hpsklearn}			& \name{hyperopt}	& Fixed		& no	& yes	& no			& yes	\\
	\name{auto-sklearn}			& \name{SMAC}		& Fixed		& yes	& Enc.	& Cluster		& yes	\\
	\name{Random Search}		& Random Search		& Fixed		& no	& Enc.	& Cluster		& yes	\\
	\name{ATM}					& \name{BTB}			& Fixed		& no	& yes	& Cluster		& no	\\
	\name{H2O AutoML}			& Grid Search		& Fixed		& yes	& yes	& Cluster		& yes	\\
	\bottomrule
\end{tabular}

\caption{
	Comparison of different \ac{AutoML} frameworks. Reported are the used \ac{CASH} solver and pipeline structure. It is listed whether ensemble learning (Ensem.), categorical input (Cat.), parallel evaluation of pipelines or a timeout for evaluations are supported (Time).
}
\label{tbl:overview_automl_frameworks}

\end{table}

\paragraph{Baseline Methods}
To assess the effectiveness of the different \ac{AutoML} algorithms, two baseline methods are added:
\begin{enumerate*}[label={\arabic*)}]
 \item a dummy classifier using stratified sampling to create random predictions and 
 \item a simple pipeline consisting of an imputation of missing values and a random forest.
\end{enumerate*}
For both baseline methods the \name{scikit-learn} \shortcite{Pedregosa2011} implementation is used.

\paragraph{TPOT}
\name{TPOT} \shortcite{Olson2016,Olson2016b} is a framework for building and tuning flexible classification and regression pipelines based on genetic programming. Regarding \ac{HPO}, \name{TPOT} can only handle categorical parameters; similar to grid search all continuous hyperparameters have to be discretized. \name{TPOT}'s ability to create arbitrary complex pipelines makes it very prone for overfitting. To compensate this, \name{TPOT} optimizes a combination of high performance and low pipeline complexity. Therefore, pipelines are selected from a Pareto front using a multi-objective selection strategy. \name{TPOT} supports basically all popular \name{scikit-learn} preprocessing, classification and regression methods. It is evaluated in version 0.10.2.

\paragraph{Hyperopt-Sklearn}
\name{hyperopt-sklearn} or \name{hpsklearn} \shortcite{Komer2014} is a framework for fitting classification and regression pipelines based on \name{hyperopt}. The pipeline structure is fixed to exactly one preprocessor and one classification or regression algorithm; all algorithms are based on \name{scikit-learn}. \name{hpsklearn} only provides a thin wrapper around \name{hyperopt} by introducing the fixed pipeline structure and adding a configuration space definition. A parallelization of the configuration evaluation is not available. It supports only a rudimentary data preprocessing, namely \ac{PCA}, standard or min-max scaling and normalization. Additionally, the most popular \name{scikit-learn} classification and regression methods are supported. \name{hpsklearn} is evaluated in version 0.0.3.

\paragraph{Auto-Sklearn}
\name{auto-sklearn} \shortcite{Feurer2015,Feurer2018} is a tool for building classification and regression pipelines. All pipeline candidates have a semi-fixed structure: at first, a fixed set of data cleaning steps---including optional categorical encoding, imputation, removing variables with low variance and optional scaling---is executed. Next, an optional preprocessing and mandatory modeling algorithm are selected and tuned via \name{SMAC}. As the name already implies, \name{auto-sklearn} uses \name{scikit-learn} for all \ac{ML} algorithms. The sister package \name{Auto-WEKA} \shortcite{Thornton2013,Kotthoff2016} provides very similar functionality for the \name{WEKA} library.

In contrast to the other \ac{AutoML} frameworks presented in this section, \name{auto-sklearn} does incorporate many different performance improvements. Testing pipeline candidates is improved via parallelization on a single computer or in a cluster and each evaluation is limited by a time budget. \name{auto-sklearn} uses meta-learning to initialize the optimization procedure. Additionally, ensemble learning is implemented by combining the best pipelines. \name{auto-sklearn} is evaluated in version 0.5.2.

\paragraph{Random Search}
Random search is added as additional baseline method with tuned hyperparameters based on \name{auto-sklearn}. Instead of using \name{SMAC}, configurations are generated randomly. Additionally, ensemble building and meta-learning are disabled.

\paragraph{ATM}
\name{ATM} \shortcite{Swearingen2017} is a collaborative service for building optimized classification pipelines based on \name{BTB}. Currently, \name{ATM} uses a simple pipeline structure with an optional \ac{PCA}, an optional scaling followed by a tunable classification algorithm. All algorithms are based on \name{scikit-learn} and popular classification algorithms are supported.

An interesting feature of \name{ATM} is the so-called \name{ModelHub}. This central database stores information about data sets, tested configurations and their performances. By combining the performance evaluations with, currently not stored, meta-features of the data sets, a valuable foundation for meta-learning could be created. This catalog of examples could grow with every evaluated configuration enabling a continuously improving meta-learning. Yet, currently this potential is not utilized. \name{ATM} is evaluated in version 0.2.2.

\paragraph{H2O AutoML}
\name{H2O} \shortcite{H2O.ai2019} is a distributed \ac{ML} framework to assist data scientists. In the context of this paper, only the \name{H2O AutoML} component is considered. \name{H2O AutoML} is able to select and tune a classification algorithm without preprocessing automatically. Available algorithms are tested in a fixed order with either expert-defined or via randomized grid-search selected hyperparameters. In the end, the best performing configurations are aggregated to create an ensemble. In contrast to all other evaluated frameworks, \name{H2O} is developed in Java with Python bindings and does not use \name{scikit-learn}. \name{H2O} is evaluated in version 3.26.0.8.

\section{Experiments}
\label{sec:experiments}
This section provides empirical evaluations of different \ac{CASH} and pipeline building frameworks. At first, the comparability of the results is discussed and the methodology of the benchmarks is explained. Next, the usage of synthetic data sets is shortly discussed. Finally, all selected frameworks are evaluated empirically on real data.

\subsection{Comparability of Results}
A reliable and fair comparison of different \ac{AutoML} algorithms and frameworks is difficult due to different preconditions. Starting from incompatible interfaces, for example stopping the optimization after a fixed number of iterations or after a fixed timespan, to implementation details, like refitting a model on the complete data set after cross-validation, many design decisions can skew the performance comparison heavily. Moreover, the scientific papers that propose the algorithms often use different data sets for benchmarking purposes. Using agreed-on data sets with standardized search spaces for benchmarking, like it is done in other fields of research (\eg, \shortciteR{Geiger2012}), would increase the comparability.

To solve some of these problems, the \name{ChaLearn} \ac{AutoML} challenge \shortcite{Guyon2015,Guyon2016,Guyon2018} has been introduced. The \name{ChaLearn} \ac{AutoML} challenge is an online competition for \ac{AutoML}\footnote{
	Available at \url{http://automl.chalearn.org/}.
} established in 2015. It focuses on solving supervised learning tasks, namely classification and regression, using data sets from a wide range of domains without any human interaction. The challenge is designed such that participants upload \ac{AutoML} code that is going to be evaluated on a task. A task contains a training and validation data set, both unknown to the participant. Given a fixed timespan on standardized hardware, the submitted code trains a model and the performance is measured using the validation data set and a fixed loss function. The tasks are chosen such that the underlying data sets cover a wide variety of complications, \eg, skewed data distributions, imbalanced training data, sparse representations, missing values, categorical input or irrelevant features.

The \name{ChaLearn} \ac{AutoML} challenge provides a good foundation for a fair and reproducible comparison of state-of-the-art \ac{AutoML} frameworks. However, its focus on competition between various teams makes this challenge unsuited for initial development of new algorithm. The black-box evaluation and missing knowledge of the used data sets make reproducing and debugging failing optimization runs impossible. Even though the competitive concept of this challenge can boost the overall progress of \ac{AutoML}, additional measures are necessary for daily usage.

\name{HPOlib} \shortcite{Eggensperger2013} aims to provide standardized data sets for the evaluation of \ac{CASH} algorithms. Therefore, benchmarks using synthetic objective functions (see Section~\ref{sec:synthetic_test_functions}) and real data sets (see Section~\ref{sec:real_datasets}) have been defined. Each benchmark defines an objective function, a training and validation data set along with a configuration space. This way, the benchmark data set is decoupled from the algorithm under development and can be reused by other researchers leading to more comparable evaluations.

Recently, an open-source \ac{AutoML} benchmark has been published by \shortciteA{Gijsbers2019}. By integrating \ac{AutoML} frameworks via simple adapters, a fair comparison under standardized conditions is possible. Currently only four different \ac{AutoML} frameworks and no \ac{CASH} algorithms at all are integrated. Yet, this approach is very promising to provide an empirical basis for \ac{AutoML} in the future.

\subsection{Benchmarking Methodology}
All experiments are conducted using \emph{n1-standard-8} virtual machines from Google Cloud Platform equipped with Intel Xeon E5 processors with \(8\) cores and \(30\) GB memory\footnote{
	For more information see \url{https://cloud.google.com/compute/docs/machine-types}.
}. Each virtual machine uses \name{Ubuntu} 18.04.02, \name{Python} 3.6.7 and \name{scikit-learn} 0.21.3. To eliminate the effects of non-determinism, all experiments are repeated ten times with different random seeds and results are averaged. Three different types of experiments with different setups are conducted:

\begin{enumerate}
	\item Synthetic test functions (see Section~\ref{sec:synthetic_test_functions}) are limited to exactly \(250\) iterations. The performance is defined as the minimal absolute distance
    \begin{equation*}
	    \min_{\vec{\lambda}_i \in \Lambda} \lvert f(\vec{\lambda}_i) - f(\vec{\lambda}^\star) \rvert
    \end{equation*}
    between the considered configurations \(\vec{\lambda}_i\) and the global optimum \(\vec{\lambda}^\star\).

	\item \ac{CASH} solvers (see Section~\ref{sec:cash_performance}) are limited to exactly \(325\) iterations. Preliminary evaluations have shown that all algorithms basically always converge before hitting this iteration limit. The model fitting in each iteration is limited to a cut-off time of ten minutes. Configurations violating this time limit are assigned the worst possible performance. The performance of each configuration is determined using a \(4\)-fold cross-validation with three folds passed to the optimizer and using the last fold to calculate a test-performance. As loss function, the accuracy
\begin{equation}
\label{eq:accuracy}
	\mathcal{L}_{\text{Acc}} ( \hat{y}, y ) = \dfrac{1}{ |y| } \sum\limits_{i = 1}^{ |y| } \mathbbm{1} (\hat{y}_i = y_i)
\end{equation}
is used, with \(\mathbbm{1}\) being an indicator function.

	\item \ac{AutoML} frameworks (see Section~\ref{sec:automl_performance}) are limited by a soft-limit of \(1\) hour and a hard-limit of \(1.25\) hours. Fitting of single configurations is aborted after ten minutes if the framework supports a cut-off time. The performance of each configuration is determined using a \(4\)-fold cross-validation with three folds passed to the \ac{AutoML} framework\footnote{
	Internally, the \ac{AutoML} frameworks may implement different methods to prevent overfitting, \eg, a nested cross-validation or a hold-out data set.
} and using the last fold to calculate a test-performance. As loss function, the accuracy in Equation~\eqref{eq:accuracy} is used.
\end{enumerate}

The evaluation timeout of ten minutes cancels roughly \(1.4\%\) of all evaluations. Consequently, the influence on the final results is negligible while the overall runtime is reduced by orders of magnitude. Preliminary tests revealed that all algorithms are limited by CPU power and not available memory. Therefore, the memory consumption is not further considered. Frameworks supporting parallelization are configured to use eight threads. Furthermore, frameworks supporting memory limits are configured to use at most \(4096\)~MB memory per thread. The source code used for the benchmarks is available online\footnote{
	Available at \url{https://github.com/Ennosigaeon/automl_benchmark}.
}.

For the third experiment, we also tested cut-off timeouts of \(4\) and \(8\) hours on ten randomly selected data sets. The performance after \(4\) or even \(8\) hours did only marginally improve in comparison to \(1\) hour and is therefore not further considered.

\subsection{Synthetic Test Functions}
\label{sec:synthetic_test_functions}

A common strategy applied for many years is using synthetic test functions for benchmarking (\eg, \shortciteR{Snoek2012,Eggensperger2015,Klein2017}). Due to the closed-form representation, the synthetic loss for a given configuration can be computed in constant time. Synthetic test functions do not allow a simulation of categorical hyperparameters leading to an unrealistic, completely unstructured configuration space. Consequently, these functions are only suited to simulate \ac{HPO} without algorithm selection. The circumvention of real data also prevents the evaluation of data cleaning and feature engineering steps. Finally, all synthetic test functions have a continuous and smooth surface. These properties do not hold for real response surfaces \shortcite{Eggensperger2015}. This implies that synthetic test functions are not suited for \ac{CASH} benchmarking. A short evaluation of the presented \ac{CASH} algorithms on selected synthetic test functions is given in Appendix~\ref{app:synthetic_test_functions}.

\subsection{Empirical Performance Models}

In the previous section it was shown that synthetic test functions are not suited for benchmarking. Using real data sets as an alternative is very inconvenient. Even though they provide the most realistic way to evaluate an \ac{AutoML} algorithm, the time for fitting a single model can become prohibitively large. In order to lower the turnaround time for testing a single configuration significantly, \acp{EPM} have been introduced \shortcite{Eggensperger2015,Eggensperger2017}.

An \ac{EPM} is a surrogate for a real data set that models the response surface of a specific loss function. By sampling the performance of many different configurations, a regression model of the response surface is created. In general, the training of an \ac{EPM} is very expensive as several thousand models with different configurations have to be trained. The benefit of this computational heavy setup phase is that the turnaround time of testing new configurations proposed by an \ac{AutoML} algorithm is reduced significantly. Instead of training an expensive model, the performance can be retrieved in quasi constant time from the regression model.

In theory, \acp{EPM} can be used for \ac{CASH} as well as complete pipeline creation. However, due to the quasi exhaustive analysis of the configuration space, \acp{EPM} suffer heavily from the curse of dimensionality. Consequently, no \acp{EPM} are available to test the performance of a complete \ac{ML} pipeline. In the context of this work \acp{EPM} have not been evaluated. Instead, real data sets have been used directly.

\subsection{Real Data Sets}
\label{sec:real_datasets}

All previously introduced methods for performance evaluations only consider selecting and tuning a modeling algorithm. Data cleaning and feature engineering are ignored completely even though those two steps have a significant impact on the final performance of an \ac{ML} pipeline \shortcite{Chu2016}. The only possibility to capture and evaluate all aspects of \ac{AutoML} algorithms is using real data sets. However, real data sets also introduce a significant evaluation overhead, as for each pipeline multiple \ac{ML} models have to be trained. Depending on the complexity and size of the data set, testing a single pipeline can require several hours of wall clock time. In total, multiple months of CPU time were necessary to conduct all evaluations with real data sets presented in this benchmark.

As explained in Section~\ref{sec:problem_formulation}, the performance of an \ac{AutoML} algorithm depends on the tested data set. Consequently, it is not useful to evaluate the performance on only a few data sets in detail but instead the performance is evaluated on a wide range of different data sets. To ensure reproducibility of the results, only publicly available data sets from \name{OpenML} \shortcite{Vanschoren2014}, a collaborative platform for sharing data sets in a standardized format, have been selected. More specifically, a combination of the curated benchmarking suites \name{OpenML100}\footnote{
	Available at \url{https://www.openml.org/s/14}.
} \shortcite{Bischl2017}, \name{OpenML-CC18}\footnote{
	Available at \url{https://www.openml.org/s/99}.
} \shortcite{Bischl2019} and \name{AutoML Benchmark}\footnote{
	Available at \url{https://www.openml.org/s/218}.
} \shortcite{Gijsbers2019} is used. The combination of these benchmarking suits contains \(137\) classification tasks with high-quality data sets having between \(500\) and \(600,000\) samples and less than \(7,500\) features. High-quality does not imply that no preprocessing of the data is necessary as, for example, some data sets contain missing values. A complete list of all data sets with some basic meta-features is provided in Appendix~\ref{app:evaluated_data_sets}. All \ac{CASH} algorithm and most \ac{AutoML} frameworks do not support categorical features. Therefore, categorical features of all data sets are transformed using one hot encoding. Furthermore, data sets are shuffled to remove potential impacts of ordered data.

\subsubsection{CASH Algorithms}
\label{sec:cash_performance}
All previously mentioned \ac{CASH} algorithms are tested on all data sets. Therefore, a hierarchical configuration space containing \(13\) classifiers with a total number of \(58\) hyperparameters is created. This configuration space---listed in Table~\ref{tbl:classification_config_space} and Appendix~\ref{app:complete_config_space}---is used by all \ac{CASH} algorithms. Algorithms not supporting hierarchical configuration spaces use a configuration space without conditional dependencies. Furthermore, if no categorical or integer hyperparameters are supported, these parameters are transformed to continuous variables. Some algorithms only support \ac{HPO} without algorithm selection. For those algorithms, an optimization instance is created for each \ac{ML} algorithm. The number of iterations per estimator is limited to \(25\) such that the total number of iterations still equals \(325\).

\begin{table}[hb]
\centering

\renewcommand{\arraystretch}{1}
\begin{tabular}{l @{\hskip 10mm} l l l}
	\toprule
	Algorithm & \(\# \lambda\) & Cat. & Con. \\
	\midrule
	
	Bernoulli na\"ive Bayes		& 2		& 1	& 1	\\
	Multinomial na\"ive Bayes		& 2		& 1	& 1	\\
	Decision Tree					& 4		& 1	& 3	\\
	Extra Trees						& 5		& 2	& 3	\\
	Gradient Boosting				& 8		& 1	& 5	\\
	Random Forest					& 5		& 2	& 4	\\
	K Nearest Neighbors				& 3		& 2	& 1	\\
	LDA									& 4		& 1	& 3	\\
	QDA									& 1		& 0	& 1	\\
	Linear SVM						& 4		& 2	& 2	\\
	Kernel SVM						& 7		& 2	& 5	\\
	Passive Aggressive				& 4		& 2	& 2	\\
	Linear Classifier with SGD	& 10	& 4	& 6	\\
	
	\bottomrule
\end{tabular}

\caption{
	Configuration space for classification algorithms. In total, 13 different algorithms with 58 hyperparameters are available. The number of categorical (Cat.), continuous (Con.) and total number of hyperparameters (\(\# \lambda\)) is listed.
}
\label{tbl:classification_config_space}

\end{table}

For grid search, each continuous hyperparameter is split into two distinct values leading to \(6,206\) different configurations. As the number of evaluations is limited to \(325\) configurations, only the first \(10\) classifiers are tested completely, \textit{Kernel \ac{SVM}} only partially, \textit{Passive Aggressive} and \textit{SGD} not at all. 

Table~\ref{tbl:results_evaluation_cash} in Appendix~\ref{app:raw_experiment_results} contains the raw results of the evaluation. It reports the average accuracies over all trials per data set. \(23\) of the evaluated data sets contain missing values. As no algorithm in the configuration space is able to handle missing values, all evaluations on these data sets failed and are not further considered.

In the following, accuracy scores are normalized to an interval between zero and one to obtain data set independent evaluations. Zero represents the performance of the dummy classifier and one the performance of the random forest. Algorithms outperforming the random forest baseline obtain results greater than one.

\begin{figure}[ht]
	\centering
	\includegraphics[width=1\linewidth]{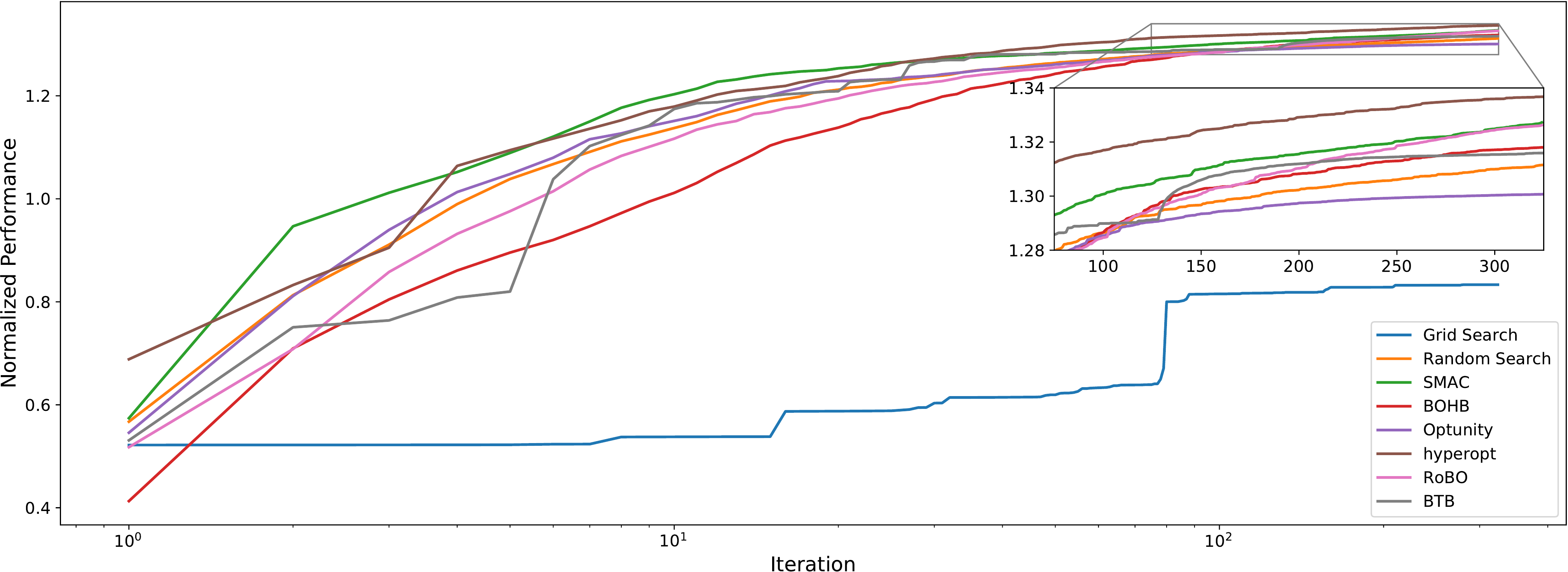}
	
	\caption{Normalized performance of the incumbent per iteration. Results are averaged over all data sets and \(10\) repetitions.}
	\label{fig:performance_cash_incumbent}

\end{figure}

Figure~\ref{fig:performance_cash_incumbent} shows the performance of the best incumbent per iteration averaged over all data sets. It is important to note that the results for the very first iterations are slightly skewed due to the parallel evaluation of candidate configurations. Iterations are recorded in order of finished evaluation timestamps, meaning that \(8\) configurations started in parallel are recorded as \(8\) distinct iterations.

\begin{table}[h]
\center

\newrobustcmd{\B}{\fontseries{b}\selectfont}

\renewcommand{\arraystretch}{1.2}
\begin{tabular}{@{} l l l l l l @{\hskip 2mm} l @{\hskip 2mm} l l @{}}
	\toprule
			& Grid	& Random & \name{SMAC} & \name{BOHB} & \name{Optunity} & \name{hyperopt} & \name{RoBO} &\name{BTB}\\
	\midrule
	Rep.	 &	0.0656	& 0.0428	& 0.0395	& 0.0414	& 0.0514	& 0.0483	& 0.0421	& 0.0535	\\
	Data Set & 0.7655	& 1.1004	& 1.1420	& 1.1478	& 1.0732	& 1.1206	& 1.1334	& 1.1302	\\
	\bottomrule
\end{tabular}

\caption{
	Standard deviation of the normalized performance of the final incumbent averaged over ten repetitions (Rep.) and all data sets (Data Set).
}
\label{tbl:stability_cash_incumbent}

\end{table}

\begin{figure}[h]
	\centering
	\includegraphics[width=1\linewidth]{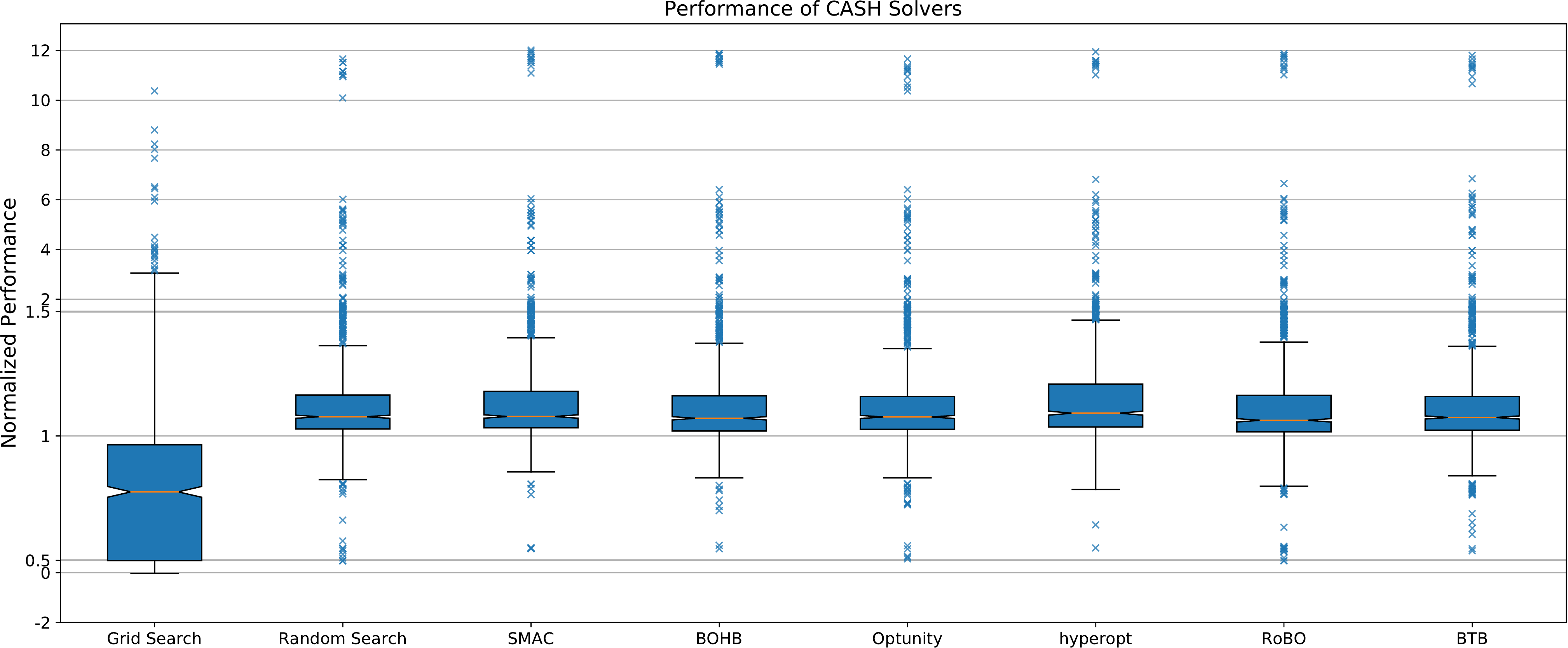}
	
	\caption{Normalized performance of the final incumbent per \ac{CASH} solvers. For better readability, performances between \(0.5\) and \(1.5\) are stretched out.}
	\label{fig:performance_cash}

\end{figure}

It is apparent that all methods except grid search are able to outperform the random forest baseline within roughly \(10\) iterations. After \(325\) iterations, all algorithms converge to similar performance measures. The individual performances after \(325\) iterations are also displayed in Figure~\ref{fig:performance_cash}. Table~\ref{tbl:stability_cash_incumbent} contains the standard deviation of the normalized performance of the final incumbent after the optimization. Values averaged over ten repetitions and all data sets are shown. It is apparent that the normalized performance heavily depends on the used data set.

A pair-wise comparison of the performances of the final incumbent is displayed in Table~\ref{tbl:pair_wise_cash_results}. It is apparent that \name{hyperopt} outperforms all other optimizers and grid search is basically always outperformed. Yet, a more detailed comparison of performances, provided in Figure~\ref{fig:pair_wise_automl_results} in Appendix~\ref{app:raw_experiment_results}, reveals that absolute performance differences are small.

\begin{table}[h]
\center

\newrobustcmd{\B}{\fontseries{b}\selectfont}

\renewcommand{\arraystretch}{1.2}
\begin{tabular}{@{} l @{\hskip 2mm} l l @{\hskip 2mm} l l l @{\hskip 2mm} l @{\hskip 2mm} l l @{}}
	\toprule
				& Grid	& Random 	& \name{SMAC}	& \name{BOHB}	& \name{Optunity}	& \name{hyperopt}	& \name{RoBO}	& \name{BTB} \\
	\midrule
Grid				& --	& 0.0263	& 0.0175	& 0.0175	& 0.0263		& 0.0175		& 0.0175	& 0.0263 \\
Random				& 0.9561	& ---	& 0.3771	& 0.6403	& 0.5175		& 0.0614		& 0.5964	& 0.5614 \\
\name{SMAC}			& 0.9649	& 0.5614	& ---	& 0.8508	& 0.6228		& 0.2192		& 0.7105	& 0.6403 \\
\name{BOHB}			& 0.9649	& 0.2807	& 0.0877	& ---	& 0.3596		& 0.0877		& 0.4385	& 0.3859 \\
\name{Optunity}		& 0.9561	& 0.4385	& 0.3245	& 0.5877	& ---		& 0.1403		& 0.5263	& 0.5087 \\
\name{hyperopt}		& 0.9649	& 0.8684	& 0.7368	& 0.8684	& 0.8157		& ---		& 0.7894	& 0.8947 \\
\name{RoBO}			& 0.9649	& 0.3596	& 0.2456	& 0.5087	& 0.4385		& 0.1491		& ---	& 0.3947 \\
\name{BTB}			& 0.9561	& 0.3859	& 0.3070	& 0.5701	& 0.4385		& 0.0614		& 0.5614	& --- \\
	\midrule
Avg. Rank	& 7.7280	& 3.9210	& 3.0964	& 5.0438	& 4.2192		& 1.7368		& 4.6403	& 4.4122 \\
 	\bottomrule
\end{tabular}

\caption{
	Fraction of data sets on which the \ac{CASH} solvers in each row performed better than the framework in each column. As \ac{CASH} solvers can yield identical performances, the according fractions do not have to add up to \(1\). Additionally, the rank of each \ac{CASH} solver is given.
}
\label{tbl:pair_wise_cash_results}

\end{table}

\begin{figure}[h]
	\centering
	\includegraphics[width=1\linewidth]{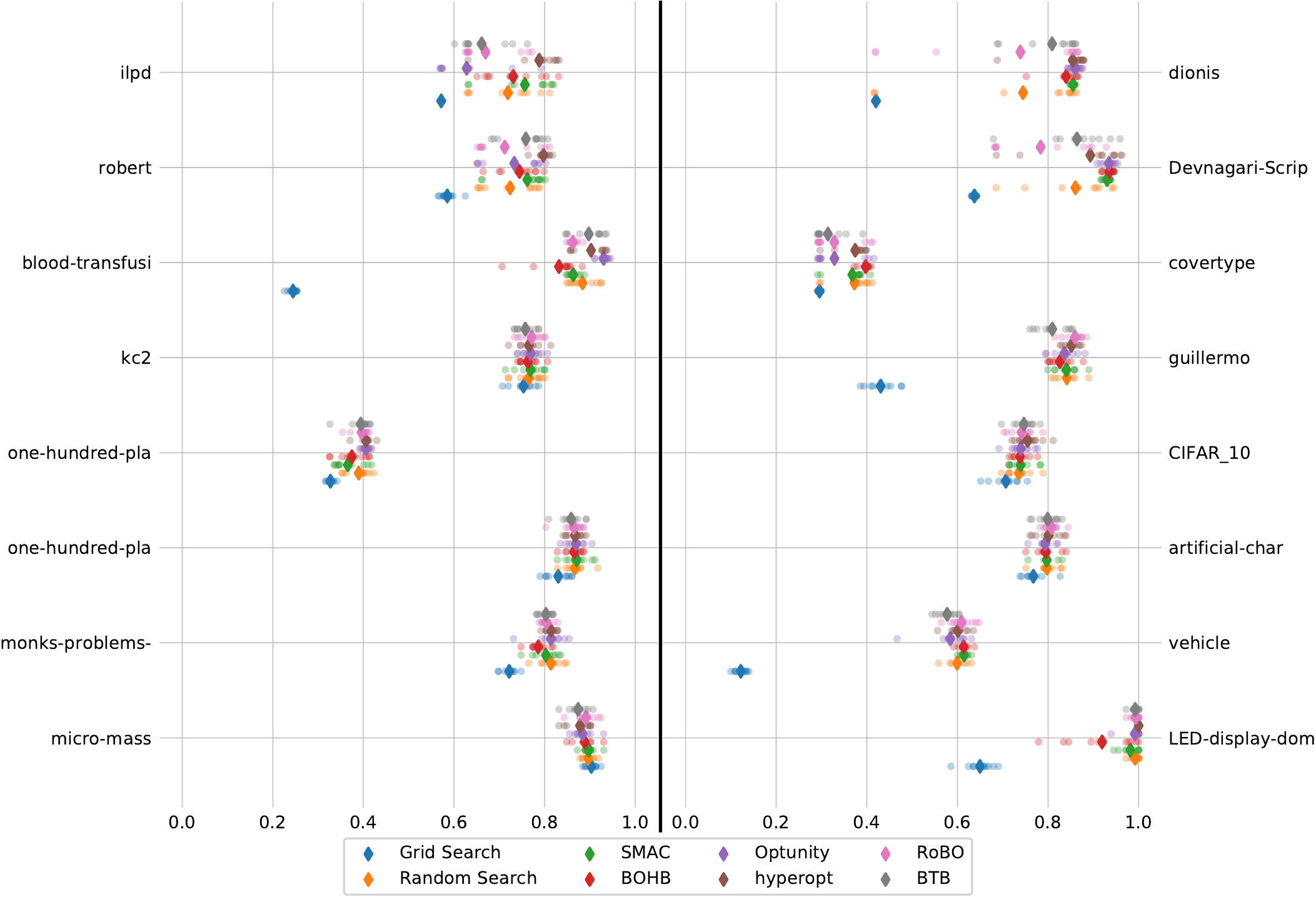}
	
	\caption{Raw and averaged accuracy of all \ac{CASH} solvers on selected data sets.}
	\label{fig:performance_ds_cash}

\end{figure}

Figure~\ref{fig:performance_ds_cash} shows the raw scores for each \ac{CASH} framework over \(10\) repetitions for \(16\) data sets. Those data sets were selected as they show the highest deviation of the scores over ten repetitions. The remaining data sets yielded very consistent results. We do not know which data set properties are responsible for the unstable results.

Next, we examine the similarity of the proposed configurations per data set. Therefore, numerical hyperparameters are normalized by their according search space, categorical hyperparameters are not transformed. We decided to only compare configurations having the same classification algorithm. For each classification algorithm, all configuration vectors are aggregated using mean shift clustering \shortcite{Fukunaga1975} with a bandwidth \(h = 0.25\). To account for the mixed-type vector representations, the Gower distance \shortcite{Gower1971} is used as the distance metric between two configurations. To assess the quality of the resulting clusters---and therefore also the overall configuration similarity---, the silhouette coefficient \shortcite{Rousseeuw1987} is computed.

\begin{figure}[h]
	\centering
	\includegraphics[width=1\linewidth]{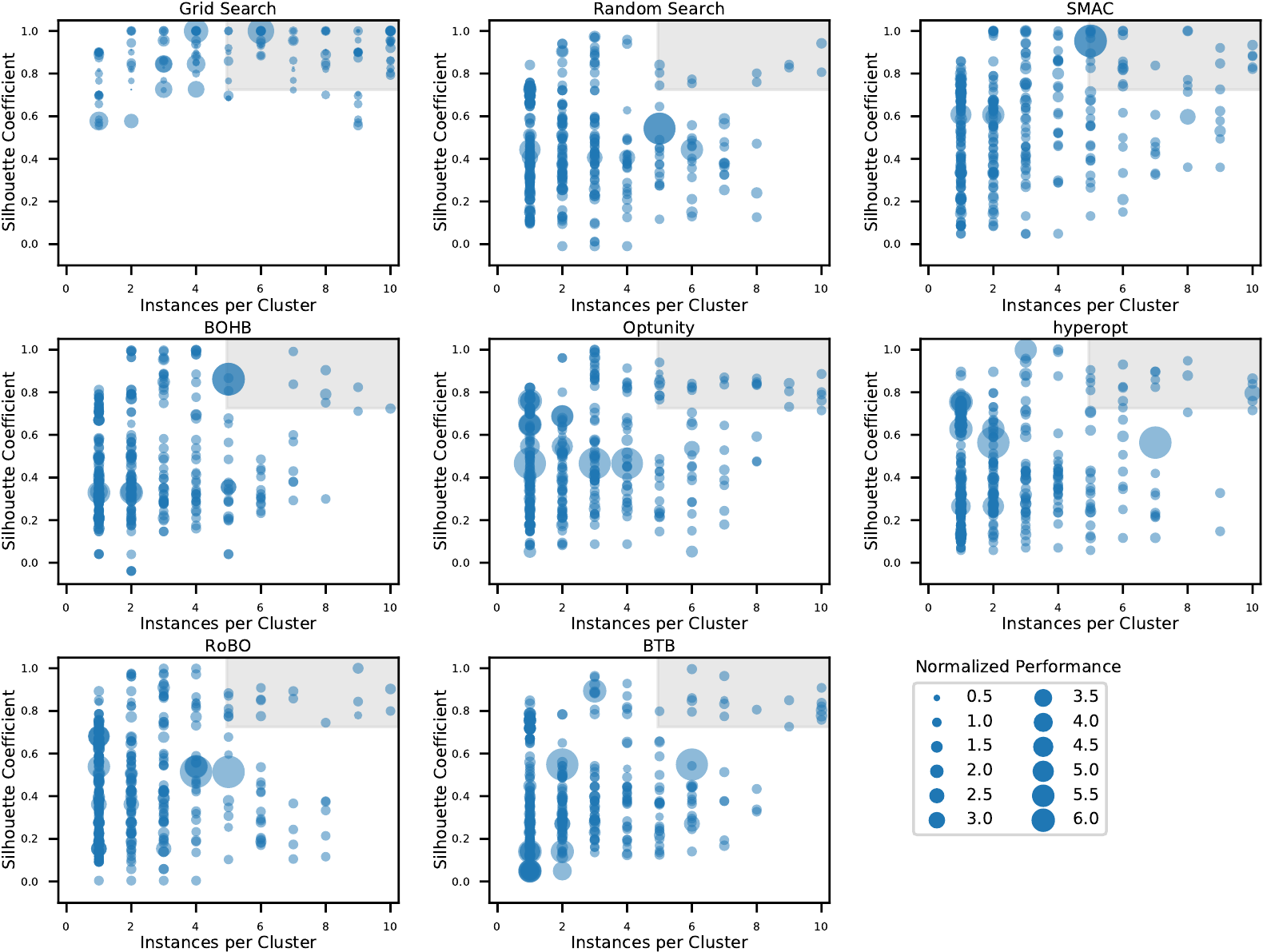}
	
	\caption{Similarity of configurations versus number of instances per cluster. Each marker represents the similarity of configurations for a single data set and single classification algorithm. The marker size indicates the normalized accuracy (larger equals higher accuracy). Clusters in the highlighted area are considered to contain similar configurations. Each subplot considers only configurations yielded by the stated \ac{CASH} algorithm.}
	\label{fig:config_cimilarity_cash}

\end{figure}

Figure~\ref{fig:config_cimilarity_cash} shows the silhouette coefficient versus number of instances per cluster. Displayed are clusters of all configurations aggregated per \ac{CASH} algorithm. On average, each \ac{CASH} algorithm yields \(3.0670 \pm 2.3772\) different classification algorithms. Most clusters contain only a few configurations with a low silhouette coefficient indicating that the resulting hyperparameters have a high variance.

We require clusters to contain at least \(5\) configurations to be considered as similar. In addition, the silhouette coefficient has to be greater than \(0.75\). In total, \(106\) of \(114\) data sets contain at least one cluster with similar configurations. However, most of those clusters are created by grid search which usually yields identical configurations for each trial. \(11\) data sets yield configurations with a high similarity for at least half of the \ac{CASH} algorithms. However, for most data sets configurations are very dissimilar. It is not apparent which meta-features are responsible for those results. In summary, most \ac{CASH} procedures yield highly different hyperparameters on most data sets depending on the random seed.

\begin{figure}[h]
	\centering
	\includegraphics[width=1\linewidth]{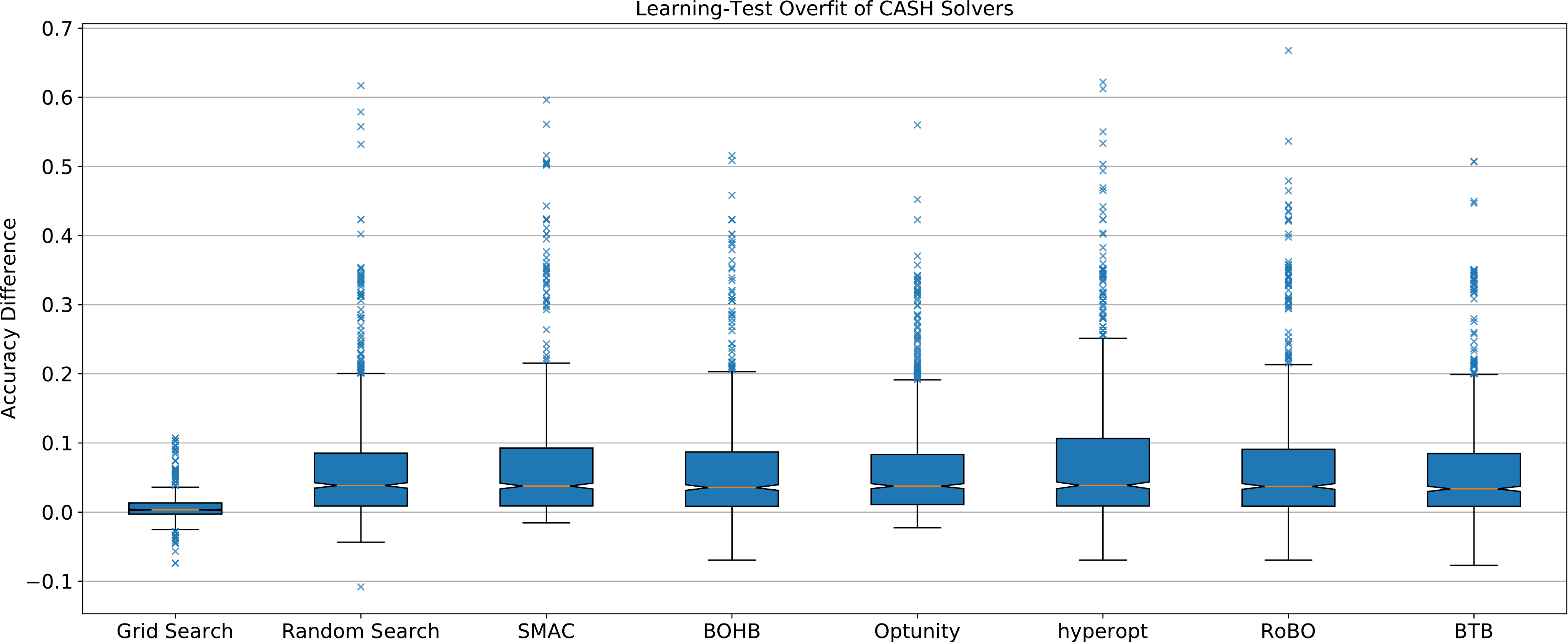}
	
	\caption{Overfit estimation between the learning and testing data set. Displayed are the raw differences between the accuracy scores. Larger values indicate higher overfitting.}
	\label{fig:overfitting_cash}

\end{figure}

Finally, we examine the known tendency of \ac{AutoML} tools to overfit \shortcite{Fabris2019}. In Figure~\ref{fig:overfitting_cash}, an estimate of the overfitting tendency of the different \ac{CASH} solvers is given. Displayed are the differences between the accuracy on the training and test data set. It is apparent that on average, all evaluated methods---except grid search---have a similar tendency to overfit. For single instances, all \ac{CASH} methods, again with the exception of grid search, suffer heavily from overfitting.

\subsubsection{AutoML Frameworks}
\label{sec:automl_performance}
Next, \ac{AutoML} frameworks capable of building complete \ac{ML} pipelines are evaluated. Therefore, all data sets from the \name{AutoML Benchmark} suite are used. Additionally, all data sets from the \name{OpenML100} and \name{OpenML-CC18} suites unable to be processed by \ac{CASH} procedures---namely data sets containing missing values---are selected. The final list of all \(73\) selected data sets is provided in Table~\ref{tbl:results_evaluation_frameworks} in Appendix~\ref{app:raw_experiment_results}.

\name{ATM} does not provide the possibility to abort configuration evaluations after a fixed time and therefore often exceeds the total time budget. To enforce the time budget, all configuration evaluations are manually aborted after \(1.25\) hours. \name{Random Search} uses \name{auto-sklearn} with a random configuration generation. Meta-learning and ensemble support are deactivated. As \name{hyperopt-sklearn} does not support parallelization, only single-threaded evaluations of configurations are used. Furthermore, \name{hyperopt-sklearn} was manually extended to support a time budget instead of number of iterations. The remaining optimizers and all unmentioned parameters are used with their default parameters. 

Table~\ref{tbl:results_evaluation_frameworks} in Appendix~\ref{app:raw_experiment_results} contains the raw results of the evaluation. The average accuracy over all trials per data set is reported. In contrast to the \ac{CASH} algorithms, the \ac{AutoML} frameworks struggle with various data sets. \name{ATM} drops samples with missing values in the training set. Data sets \(38\), \(1111\), \(1112\), \(1114\) and \(23380\) contain missing values for every single sample. Consequently, \name{ATM} uses an empty training set and crashes. \name{hyperopt-sklearn} is very fragile, especially regarding missing values. If the very first configuration evaluation of a data set fails, \name{hyperopt-sklearn} aborts the optimization. To compensate this issue, the very first evaluation is repeated upto \(100\) times. Furthermore, the optimization often does not stop after the soft-timeout for no apparent reason. \name{TPOT} sometimes crashes with a segmentation fault. For multiple data sets \name{TPOT} times out after first generation. Consequently, only random search without genetic programming is performed. Data sets \(40923\), \(41165\) and \(41167\) time out consistently with no result. \name{auto-sklearn} and \name{random search} both violated the memory constraints on the data sets \(40927\), \(41159\) and \(41167\). Finally, for \name{H2O AutoML} the Java server consistently crashes for no apparent reason on the data sets \(40978\), \(41165\), \(41167\) and \(41169\). Data set \(41167\) is the largest evaluated data set. This could explain why so many frameworks are struggling with this specific data set. In the following analysis, these failing data sets are ignored.

\begin{figure}[h]
	\centering
	\includegraphics[width=1\linewidth]{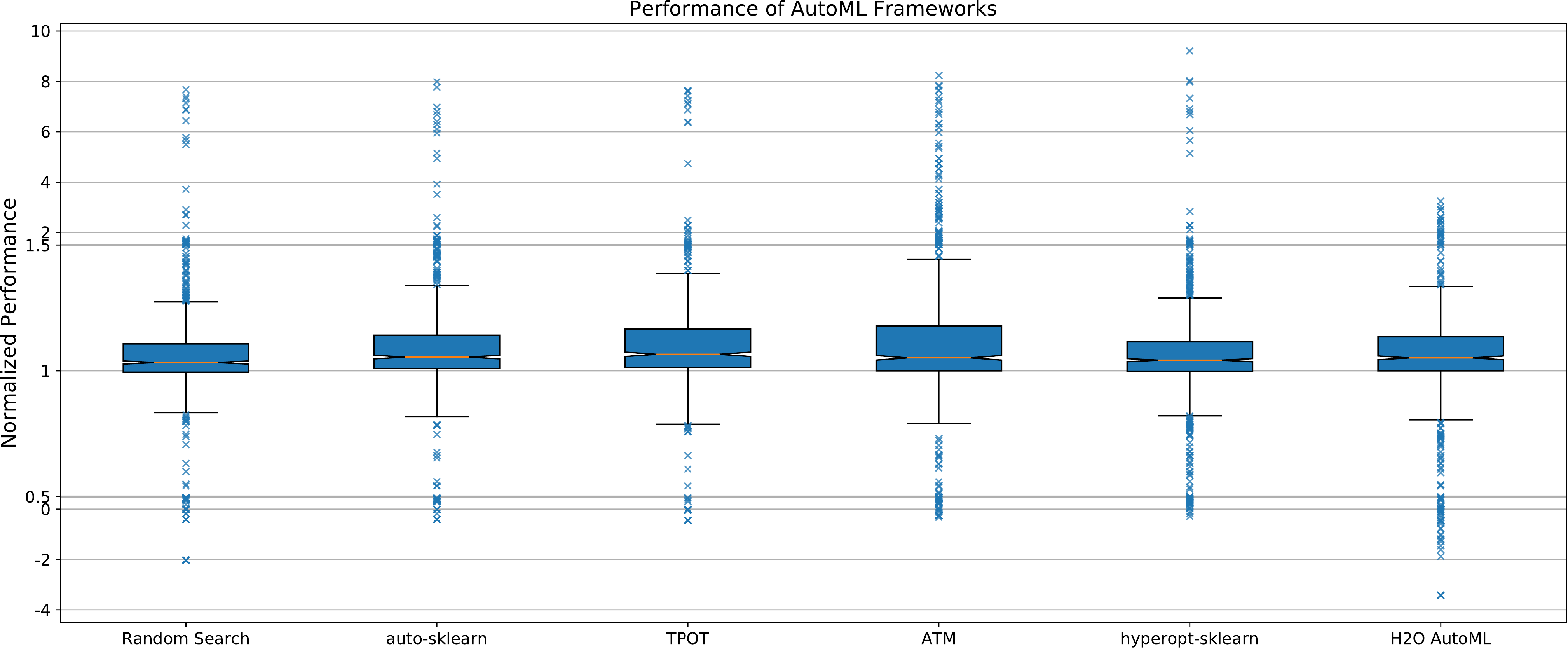}
	
	\caption{Normalized performance of the final pipeline per \ac{AutoML} framework. For better readability, performances between \(0.5\) and \(1.5\) are stretched out.}
	\label{fig:performance_automl_frameworks}

\end{figure}

Figure~\ref{fig:performance_automl_frameworks} contains the normalized performances of all \ac{AutoML} frameworks averaged over all data sets. It is apparent that all frameworks are able to outperform the random forest baseline on average. However, single results vary significantly. Table~\ref{tbl:pair_wise_automl_results} compares all framework pairs and lists the average rank for each framework. It is apparent that \name{TPOT} outperforms the most frameworks averaged over all data sets. A detailed pair-wise comparison including the absolute performance differences is provided in Figure~\ref{fig:pair_wise_automl_results} in Appendix~\ref{app:raw_experiment_results}.

\begin{table}[h]
\center

\newrobustcmd{\B}{\fontseries{b}\selectfont}

\renewcommand{\arraystretch}{1.2}
\begin{tabular}{@{} l @{\hskip 10mm} l l l l l l @{}}
	\toprule
					& \name{TPOT}	& \name{hpsklearn}	& \name{auto-sklearn}	& Random 		& \name{ATM} 		& \name{H2O} \\
	\midrule
\name{TPOT}			& ---			& 0.7571 			& 0.6086 				& 0.8529		& 0.6000 			& 0.5000 \\
\name{hpsklearn}	& 0.2285 		& ---				& 0.2816 				& 0.5571 		& 0.4117 			& 0.2898 \\
\name{auto-sklearn}	& 0.3623 		& 0.7042 			& ---					& 0.8000 		& 0.4848			& 0.5294 \\
Random 				& 0.1323 		& 0.4428			& 0.2000 				& ---			& 0.3846 			& 0.3283 \\
\name{ATM} 			& 0.3692 		& 0.5735 			& 0.4848 				& 0.6153 		& ---				& 0.4687 \\

\name{H2O} 			& 0.4705 		& 0.7101 			& 0.4558 				& 0.6716 		& 0.5156 			& --- \\
	\midrule
Avg. Rank		& 2.6027		& 4.0410			& 2.9863				& 4.4109 		& 3.4931 			& 3.1643 \\
	\bottomrule
\end{tabular}

\caption{
	Fraction of data sets on which the framework in each row performed better than the framework in each column. As frameworks can yield identical performances, the according fractions do not have to add up to \(1\). Additionally, the rank of each framework averaged over all frameworks is given.
}
\label{tbl:pair_wise_automl_results}

\end{table}

\begin{figure}
	\centering
	\includegraphics[width=1\linewidth]{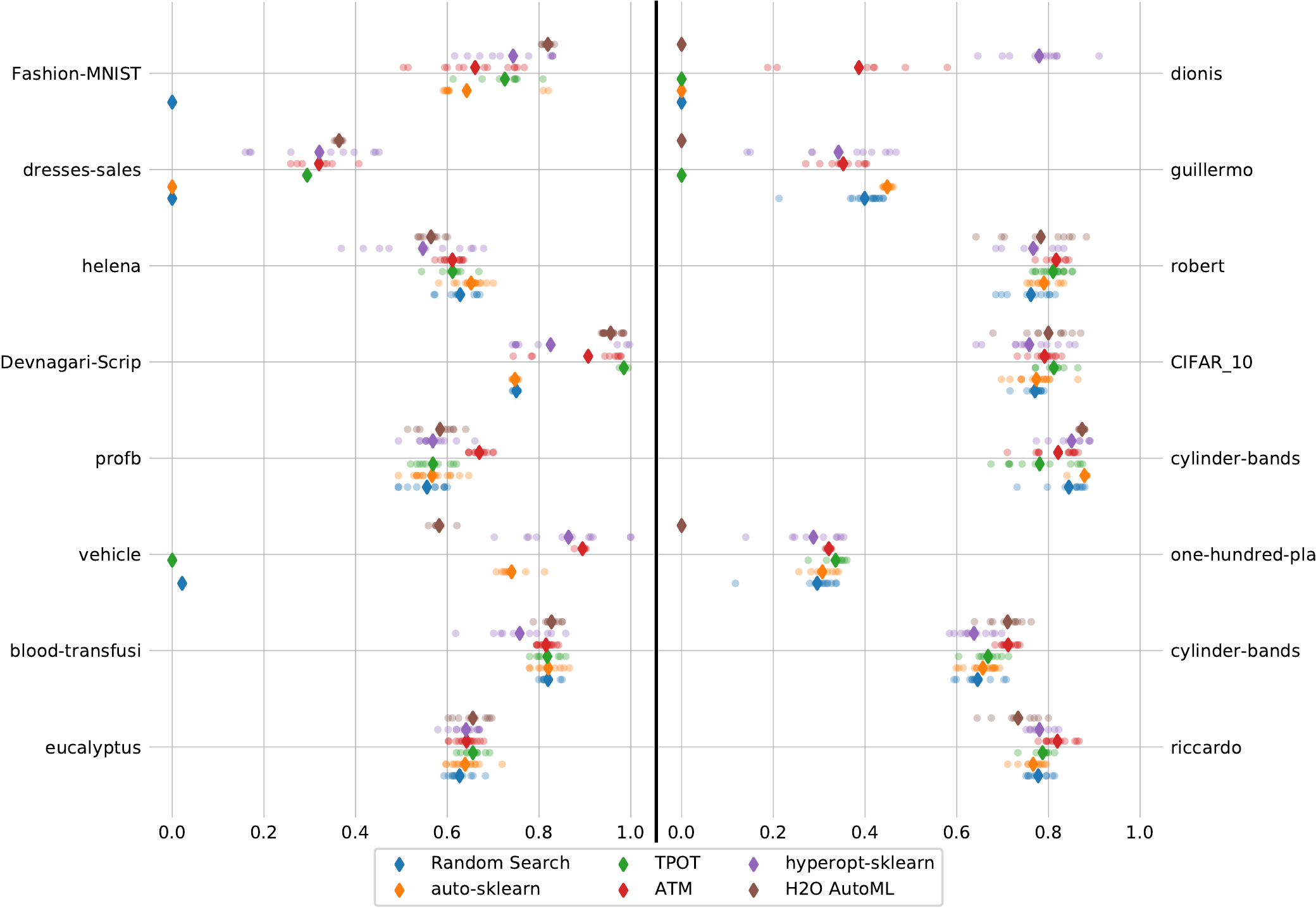}
	
	\caption{Raw and averaged accuracy of all \ac{AutoML} frameworks on selected data sets.}
	\label{fig:performance_ds_framworks}

\end{figure}

Figure~\ref{fig:performance_ds_framworks} shows raw scores for each \ac{AutoML} framework over ten trials for \(16\) data sets. Those data sets were selected as they show the highest deviation of the scores over the ten trials. About \(50\%\) of all evaluated data sets show a high variance in the obtained results. The remaining data sets yield very consistent performances. It is not clear which data set features are responsible for this separation. Table~\ref{tbl:stability_automl_incumbent} contains the standard deviation of the normalized performance of the final pipeline after the optimization. Shown are averaged values over ten repetitions and all data sets. In comparison with the \ac{CASH} solvers, the stability within ten iterations has decreased while the stability across data sets has increased.

\begin{table}
\center

\newrobustcmd{\B}{\fontseries{b}\selectfont}

\renewcommand{\arraystretch}{1.2}
\begin{tabular}{@{} l l l l l l l @{}}
	\toprule
		& \name{TPOT}	& \name{hpsklearn}	& \name{auto-sklearn}	& Random	& \name{ATM}	& \name{H2O} \\
	\midrule
	Rep.	 & 0.0761	& 0.1508			& 0.0843				& 0.0955	& 0.0963		& 0.0993	\\
	Data Set & 0.7343	& 0.7004			& 0.6772				& 0.6956	& 0.8938		& 0.2526	\\
	\bottomrule
\end{tabular}

\caption{
	Standard deviation of the normalized performance of the final pipeline averaged over ten repetitions (Rep.) and all data sets (Data Set).
}
\label{tbl:stability_automl_incumbent}

\end{table}

\begin{figure}[h]
	\centering
	\includegraphics[width=1\linewidth]{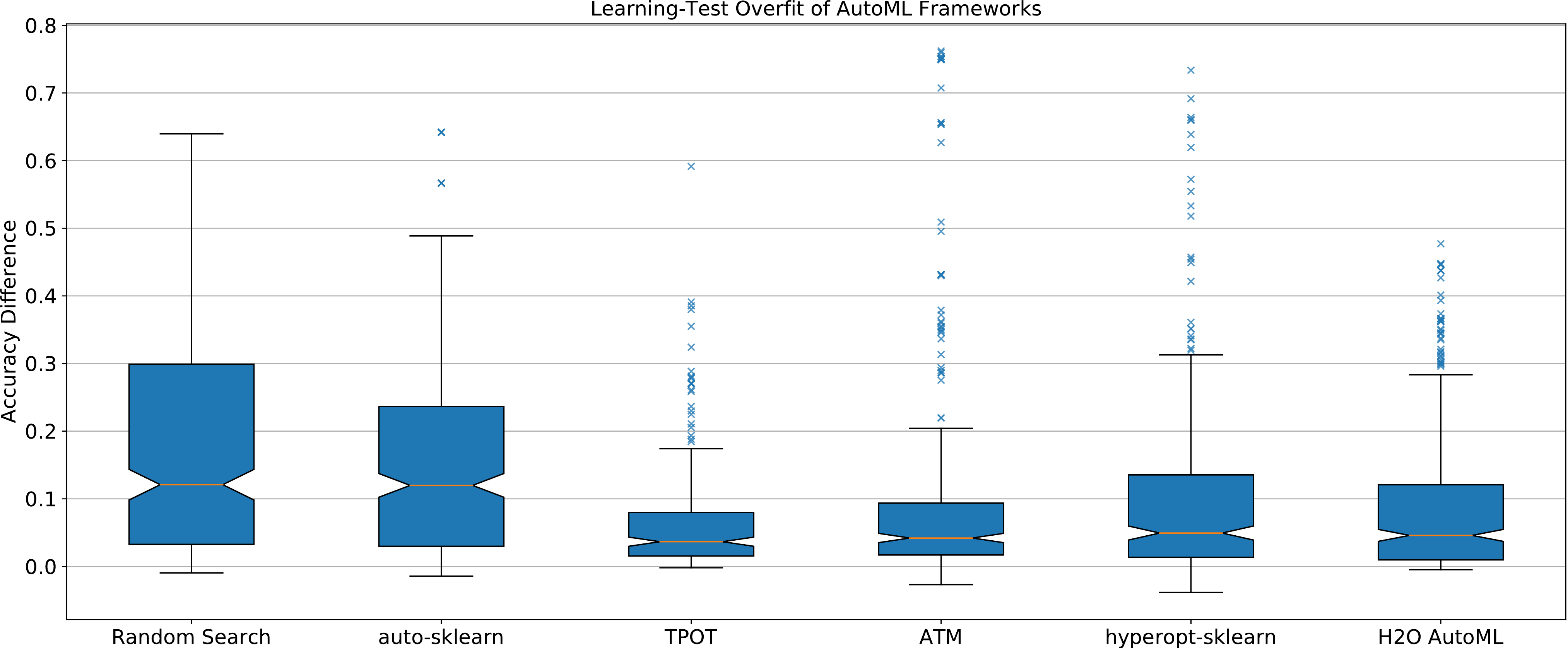}
	
	\caption{Overfit estimation between the learning and testing data set. Displayed are the raw differences between the accuracy scores. Larger values indicate higher overfitting.}
	\label{fig:overfitting_frameworks}

\end{figure}

Figure~\ref{fig:overfitting_frameworks} shows an estimate of the test-training overfit for all evaluated frameworks. In general, the \ac{AutoML} frameworks, especially random search and \name{auto-sklearn}, appear to be more prone to overfitting than \ac{CASH} solvers. All tested frameworks overfit strongly for single instances.

\begin{figure}
	\centering
	\includegraphics[width=1\linewidth]{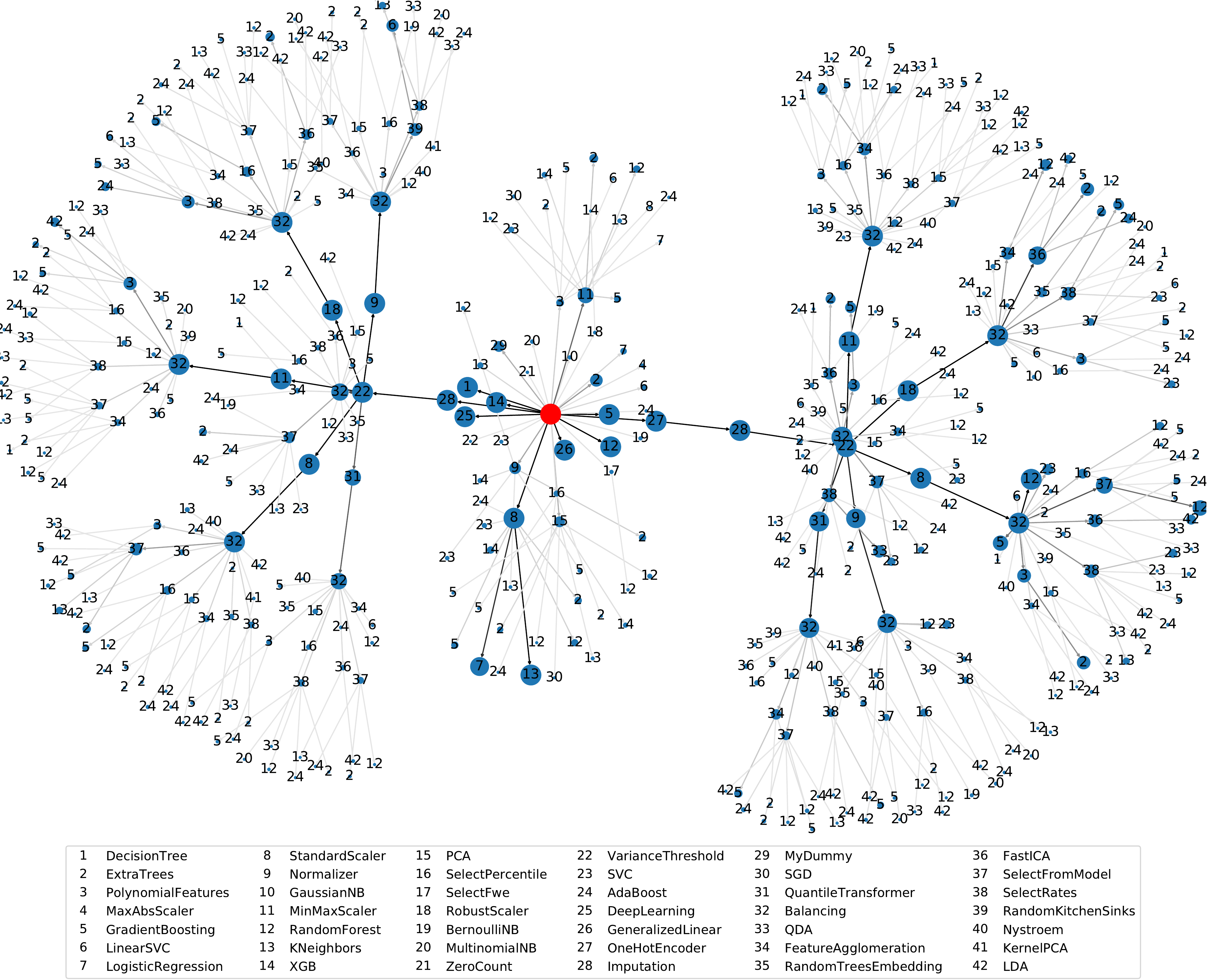}
	
	\caption{Overview of constructed \ac{ML} pipelines. The node size and edge color indicate the popularity of specific (sub-)pipelines. The red node represents the root node. Pipelines are created by following the graph from the root to a leaf node.}
	\label{fig:overview_pipelines}

\end{figure}

Figure~\ref{fig:overview_pipelines} provides an overview of often constructed pipelines. For readability, pipelines were required to be constructed at least thrice to be included in the graph. Ensembles of pipelines are treated as distinct pipelines. \name{TPOT}, \name{ATM}, \name{hyperopt-sklearn} and \name{H2O AutoML} produce on average pipelines with less than two steps. Consequently, the cluster of pipelines around the root node is created by those \ac{AutoML} frameworks. Basically all pipelines in the left and right sub-graph were created by the two \name{auto-sklearn} variants.

\begin{table}
\center

\newrobustcmd{\B}{\fontseries{b}\selectfont}

\renewcommand{\arraystretch}{1.2}
\begin{tabular}{@{} l l l l l l l @{}}
	\toprule
				& \name{TPOT}	& \name{hpsklearn}	& \name{auto-sklearn}	& Random	& \name{ATM}	& \name{H2O} \\
	\midrule
	
	\name{TPOT}			& \(0.1190\)	& \(0.1106\)	& \(0.0379\)	& \(0.0356\)	& \(0.0519\)	& \(0.1165\)	\\
	\name{hpsklearn}	& \(0.1106\)	& \(0.1926\)	& \(0.0517\)	& \(0.0461\)	& \(0.0828\)	& \(0.1414\)	\\
	\name{auto-sklearn} & \(0.0379\)	& \(0.0517\)	& \(0.5996\)	& \(0.5542\)	& \(0.0557\)	& \(0.0202\)	\\
	Rand. Search		& \(0.0356\)	& \(0.0461\)	& \(0.5542\)	& \(0.5307\)	& \(0.0329\)	& \(0.0266\)	\\
	\name{ATM}			& \(0.0519\)	& \(0.0828\)	& \(0.0557\)	& \(0.0329\)	& \(0.4591\)	& \(0.0\)		\\
	\name{H2O}			& \(0.1165\)	& \(0.1414\)	& \(0.0202\)	& \(0.0266\)	& \(0.0\)		& \(0.3135\)	\\
	
	\bottomrule
\end{tabular}

\caption{
	Averaged pair-wise Levenshtein ratio on original \ac{ML} pipelines.
}
\label{tbl:pipeline_similarity}

\end{table}

\begin{table}
\center

\newrobustcmd{\B}{\fontseries{b}\selectfont}

\renewcommand{\arraystretch}{1.2}
\begin{tabular}{@{} l l l l l l l @{}}
	\toprule
				& \name{TPOT}	& \name{hpsklearn}	& \name{auto-sklearn} & Random	& \name{ATM}	& \name{H2O} \\
	\midrule
	
	\name{TPOT}			& \(0.7784\)	& \(0.7330\)	& \(0.3300\)	& \(0.3674\)	& \(0.7234\)	& \(0.8595\)	\\
	\name{hpsklearn}	& \(0.7330\)	& \(0.7995\)	& \(0.4048\)	& \(0.4377\)	& \(0.8208\)	& \(0.7877\)	\\
	\name{auto-sklearn} & \(0.3300\)	& \(0.4048\)	& \(0.9104\)	& \(0.8790\)	& \(0.4164\)	& \(0.2803\)	\\
	Rand. Search		& \(0.3674\)	& \(0.4377\)	& \(0.8790\)	& \(0.8423\)	& \(0.4490\)	& \(0.3272\)	\\
	\name{ATM}			& \(0.7234\)	& \(0.8208\)	& \(0.4164\)	& \(0.4490\)	& \(0.8524\)	& \(0.7769\)	\\
	\name{H2O}			& \(0.8595\)	& \(0.7877\)	& \(0.2803\)	& \(0.3272\)	& \(0.7769\)	& \(1.0\)		\\
	
	\bottomrule
\end{tabular}

\caption{
	Averaged pair-wise Levenshtein ratio on generalized \ac{ML} pipelines.
}
\label{tbl:generalized_pipeline_similarity}

\end{table}

To further assess the similarity of the resulting \ac{ML} pipelines, we transform each pipeline to a string by mapping each algorithm to a distinct letter. The similarity between two pipelines is expressed by the Levenshtein ratio \shortcite{Levenshtein1966,Ratcliff1988}. Table~\ref{tbl:pipeline_similarity} shows the averaged pair-wise Levenshtein ratio of all pipelines per \ac{AutoML} framework. It is apparent that random search and \name{auto-sklearn} have a high similarity with each other and themselves. This can be explained by the long (semi-)fixed pipeline structure. All other \ac{AutoML} frameworks yield very low similarity ratios. This can be explained partially by the different search spaces, \ie, the \ac{AutoML} frameworks do not support identical base algorithms. Therefore, we also consider a generalized representation of the \ac{ML} pipelines, \eg, replacing all classification algorithms with an identical symbol. Table~\ref{tbl:generalized_pipeline_similarity} shows that \name{TPOT}, \name{hyperopt-sklearn}, \name{ATM} and \name{H2O} build similar pipelines. \name{auto-sklearn} and random search build pipelines that differ strongly from the remaining frameworks but are still very similar to each other.

\subsubsection{Comparison with Human Experts}

Finally, all \ac{AutoML} frameworks are compared with human experts. Unfortunately, it is not possible to reuse the same data sets, as human evaluations for those data sets are not available. Instead, we decided to use two publicly available data sets from \name{kaggle}, namely \textit{Otto Group Product Classification Challenge}\footnote{Available at \url{https://www.kaggle.com/c/otto-group-product-classification-challenge}.} and \textit{Santander Customer Satisfaction}\footnote{Available at \url{https://www.kaggle.com/c/santander-customer-satisfaction}.}. Even though the evaluation of just two data sets provides only limited generalization, it can still be used to get a feeling for the competitiveness of \ac{AutoML} tools with human experts.

The experimental setup from Section~\ref{sec:automl_performance} is reused. Only the loss function is adapted to reflect the loss function used by the two challenges---\textit{logarithmic loss} for \textit{Otto} and \textit{ROC AUC} for \textit{Santander}. If any framework does not support the respective loss function, we continued to use accuracy.

\begin{table}[h]
\center

\newrobustcmd{\B}{\fontseries{b}\selectfont}

\renewcommand{\arraystretch}{1.2}
\begin{tabular}{@{} l | l l l | l l l @{}}
	\toprule
						& \multicolumn{3}{c}{\textit{Otto}}						& \multicolumn{3}{|c}{\textit{Santander}}	 \\
						& Validation	& Test			& Ranking		& Validation	& Test			& Ranking \\
	\midrule

	Human				& ---			& \(0.38055\)	& ---			& ---			& \(0.84532\)	& ---	\\
	\name{TPOT}			& \(0.81066\)	& \(1.05085\)	& \(0.7908\)	& \(0.83279\)	& \(0.83100\)	& \(0.6827\) \\
	\name{hpsklearn}	& \(0.81177\)	& \(0.58701\)	& \(0.6216\)	& \(0.66170\)	& \(0.64493\)	& \(0.8789\) \\
	\name{auto-sklearn}	& \(0.55469\)	& \(0.55081\)	& \(0.5155\)	& \(0.83547\)	& \(0.83346\)	& \(0.6543\) \\
	Random				& \(0.88702\)	& \(0.89943\)	& \(0.7777\)	& \(0.82806\)	& \(0.82427\)	& \(0.7235\) \\
	\name{ATM}			& \(0.74912\)	& \(2.43115\)	& \(0.8459\)	& \(0.68721\)	& \(0.69043\)	& \(0.8653\) \\
	\name{H2O}			& \(0.45523\)	& \(0.49628\)	& \(0.3774\)	& \(0.83406\)	& \(0.83829\)	& \(0.5329\) \\

	\bottomrule
\end{tabular}

\caption{
	Comparison with human experts for two data sets. Displayed are the validation and test score. Additionally, the fraction of human submissions that have yielded better results is given (Ranking). For \textit{Otto} smaller validation and test values are better while for \textit{Santander} higher values are better.
}
\label{tbl:comparison_human}

\end{table}

Table~\ref{tbl:comparison_human} compares all \ac{AutoML} frameworks with the best human performance. For both data sets, all algorithms are able to achieve mediocre results that are outperformed by human experts clearly. A detailed look at the leaderboard reveals that human experts required on average \(8.57\) hours to refine their initial pipeline to outperform the best \ac{AutoML} framework. Obviously, this duration does not incorporate the time spend to craft the initial solution. Considering that all frameworks spend only one hour, the results are still remarkable.

\section{Discussion and Opportunities for Future Research}
\label{sec:future_research}

The experiments in Section~\ref{sec:cash_performance} revealed that all \ac{CASH} algorithms, except grid search, perform on average very similarly. Surprisingly, random search did not perform worse than the other algorithms. The performance differences of the final configurations are not significant for most data sets with \(67.18\)\% of all configurations not being significantly worse than the best result. Mean absolute differences are less than \(1.9\%\) accuracy per data set. Consequently, a ranking of \ac{CASH} algorithms on pure performance measures is not reasonable. Other aspects like scalability or method overhead should also be considered.

On average, all \ac{AutoML} frameworks appear to perform quite similarly with a maximum performance difference of only \(2.2\)\% and three frameworks yielding no significantly worse results than the best framework. Yet, the global average conceals that for each individual data set the performance differs by \(6.7\)\% accuracy averaged over all frameworks. Only \(43.61\)\% of the final pipelines are not significantly worse than the best pipeline. In addition, the \ac{CASH} algorithms performed better than the \ac{AutoML} frameworks on \(48\)\% of the shared data sets (see Table~\ref{tbl:results_evaluation_cash} and \ref{tbl:results_evaluation_frameworks} in Appendix~\ref{app:raw_experiment_results}). This is also a surprising result as each \ac{CASH} algorithm spends on average only \(12\) minutes optimizing a single data set in contrast to the \(1\) hour of \ac{AutoML} frameworks. Possible explanations for both observations could be the significantly larger search spaces of \ac{AutoML} frameworks, a smaller number of evaluated configurations due to internal overhead, \eg, cross-validations, or the tendency of \ac{AutoML} frameworks to overfit stronger than \ac{CASH} solvers. Further evaluations are necessary to explain this behavior.

Currently, \ac{AutoML} frameworks build pipelines with an average length of less than \(2.5\) components. This is partly caused by frameworks with a short, fixed pipeline layout. Yet, also \name{TPOT} yields pipelines with less than \(1.5\) components on average. Consequently, the potential of specialized pipelines is currently not utilized at all. A benchmarking of other frameworks capable of building flexible pipelines, \eg, \name{ML-Plan} \shortcite{Mohr2018,Wever2018} or \name{P4ML} \shortcite{Gil2018}, in combination with longer optimization periods is desirable to understand the capabilities of creating adaptable pipelines better.

Currently, \ac{AutoML} is completely focused on supervised learning. Even though some methods may be applicable for unsupervised or reinforcement learning, researchers always test their proposed approaches for supervised learning. Dedicated research for unsupervised or reinforcement learning could boost the development of \ac{AutoML} framework for currently uncovered learning problems. Additionally, specialized methods could improve the performance for those tasks.

The majority of all publications currently treats the \ac{CASH} problem either by introducing new solvers or adding performance improvements to existing approaches. A possible explanation could be that \ac{CASH} is completely domain-agnostic and therefore comparatively easier to automate. However, \ac{CASH} is only a small piece of the puzzle to build an \ac{ML} pipeline automatically. Data scientists usually spend 60--80\% of their time with cleaning a data set and feature engineering and only 4\% with fine tuning of algorithms \shortcite{Press2016}. This distribution is currently not reflected in research efforts. We have not been able to find any literature covering advanced data cleaning methods in the context of \ac{AutoML}. Regarding feature creation, most methods combine predefined operators with features naively. For building flexible pipelines, currently only a few different approaches have been proposed. Further research in any of these three areas can improve the overall performance of an automatically created \ac{ML} pipeline highly.

So far, researchers have focused on a single point of the pipeline creation process. Combining flexibly structured pipelines with automatic feature engineering and sophisticated \ac{CASH} methods has the potential to beat the frameworks currently available. However, the complexity of the search space is raised to a whole new level, probably requiring new methods for efficient search. Nevertheless, the long term goal should be to build complete pipelines with every single component optimized automatically.

\ac{AutoML} aims to automate the creation of an \ac{ML} pipeline completely to enable domain experts to use \ac{ML}. Except very few publications (\eg, \shortciteR{Friedman2015,Smith2017}) current \ac{AutoML} algorithms are designed as a black-box. Even though this may be convenient for an inexperienced user, this approach has two major drawbacks:
\begin{enumerate}
	\item A domain expert has a profound knowledge about the data set. Using this knowledge, the search space can be reduced significantly.
	\item Interpretability of \ac{ML} has become more important in recent years \shortcite{Doshi-Velez2017}. Users want to be able to understand how a model has been obtained. When using hand-crafted \ac{ML} models, the reasoning of the model is often already unknown to the user. By automating the creation, the user has basically no chance to understand why a specific pipeline has been selected.
\end{enumerate}
Even though methods like \textit{feature attribution} \shortcite{GoogleLLC2019} or \textit{rule-extraction} \shortcite{Alaa2018} have already been used in combination with \ac{AutoML}, the black-box problem still prevails. Human-guided \ac{ML} \shortcite{Langevin2018,Gil2019} aims to present simple questions to the domain expert to guide the exploration of the search space. Domain experts would be able to guide model creation by their experience. Further research in this area may lead to more profound models depicting the real-world dependencies closer. Simultaneously, the domain expert would have the chance to understand the reasoning of the \ac{ML} model better. This could increase the acceptance of the proposed pipeline.

\ac{AutoML} frameworks usually introduce their own hyperparameters that can be tuned. Yet, this is basically the same problem that \ac{AutoML} tried to solve in the first place. Research leading to frameworks with less hyperparameters is desirable \shortcite{Feurer2018a}.

The experiments revealed that some data sets are better suited for \ac{AutoML} than others. Currently, we can not explain which data set meta-features are responsible for this behavior. A better understanding of the relation between data set meta-features and \ac{AutoML} algorithms may enable \ac{AutoML} for the failing data sets and boost meta-learning.

Following the \acs{CRISP-DM} \shortcite{Shearer2000}, \ac{AutoML} currently focuses only the modeling stage. However, to conduct an \ac{ML} project successfully, all stages in the \acs{CRISP-DM} should be considered. To make \ac{AutoML} truly available to novice users, integration of data acquisition and deployment measures are necessary. In general, \ac{AutoML} currently does not consider lifecycle management at all.

\section{Conclusion}
\label{sec:conclusion}
In this paper, we have provided a theoretical and empirical introduction to the current state of \ac{AutoML}. We provided the first empirical evaluation of \ac{CASH} algorithms on \(114\) publicly available real-world data sets. Furthermore, we conducted the largest evaluation of \ac{AutoML} frameworks in terms of considered frameworks as well as number of data sets. Important techniques used by those frameworks are introduced and summarized theoretically. This way, we presented the most important research for automating each step of creating an \ac{ML} pipeline. Finally, we extended current problem formulations to cover the complete process of building \ac{ML} pipelines.

The topic \ac{AutoML} has come a long way since its beginnings in the 1990s. Especially in the last ten years, it has received a lot of attention from research, enterprises and the media. Current state-of-the-art frameworks enable domain experts to build reasonably well performing \ac{ML} pipelines without knowledge about \ac{ML} or statistics. Seasoned data scientists can profit from the automation of tedious manual tasks, especially model selection and \ac{HPO}. However, automatically generated pipelines are still very basic and are not able to beat human experts yet \shortcite{Guyon2016}. It is likely that \ac{AutoML} will continue to be a hot research topic leading to even better, holistic \ac{AutoML} frameworks in the future.

\acks{%
	This work is partially supported by the Federal Ministry of Transport and Digital Infrastructure within the mFUND research initiative and the Ministry of Economic Affairs, Labour and Housing of the state Baden-W\"urttemberg within the KI-Fortschrittszentrum ``Lernende Systeme'', Grant No. 036-170017.
}

\appendix

\section{Framework Source Code}
\label{app:source_code}

Table~\ref{tbl:framework_source_code} lists the repositories of all evaluated open-source \ac{AutoML} tools. Some methods are still under active development and may differ significantly from the evaluated versions.


\begin{normalsize}
\renewcommand{\arraystretch}{0.9}
\begin{longtable}{@{} l @{\hskip 7mm} l l @{} }
	\toprule
	Algorithm			& Type				& Source Code	\\
	\midrule
	Custom 				& Both	 			& \url{https://github.com/Ennosigaeon/automl_benchmark}	\\
	
	\name{RoBO}			& \ac{CASH}			& \url{https://github.com/automl/RoBO}	\\
	\name{BTB}			& \ac{CASH}			& \url{https://github.com/HDI-Project/BTB}	\\
	\name{hyperopt}		& \ac{CASH}			& \url{https://github.com/hyperopt/hyperopt} 	\\
	\name{SMAC}			& \ac{CASH}			& \url{https://github.com/automl/SMAC3}	\\
	\name{BOHB}			& \ac{CASH}			& \url{https://github.com/automl/HpBandSter}	\\
	\name{Optunity}		& \ac{CASH}			& \url{https://github.com/claesenm/optunity}	\\
	
	\name{TPOT}			& \ac{AutoML}		& \url{https://github.com/EpistasisLab/tpot}	\\
	\name{hpsklearn}	& \ac{AutoML}		& \url{https://github.com/hyperopt/hyperopt-sklearn}		\\
	\name{auto-sklearn}	& \ac{AutoML}		& \url{https://github.com/automl/auto-sklearn}		\\
	\name{ATM}			& \ac{AutoML}		& \url{https://github.com/HDI-Project/ATM}		\\
	\name{H2O AutoML}	& \ac{AutoML}		& \url{https://github.com/h2oai/h2o-3}		\\
	\bottomrule

\caption{
	Source code repositories for all used \ac{CASH} and \ac{AutoML} frameworks.
}
\label{tbl:framework_source_code}

\end{longtable}
\end{normalsize}

\section{Synthetic Test Functions}
\label{app:synthetic_test_functions}

All \ac{CASH} algorithms from Section~\ref{sec:selected_frameworks} are tested on various synthetic test functions. Grid search and random search are used as base line algorithms. Table~\ref{tbl:results_cash_solver} contains the performance of each algorithm after the completed optimization. Over all benchmarks, \name{RoBO} was able to consistently outperform or yield equivalent results compared to all competitors.

\begin{table}[h]
\center

\newrobustcmd{\B}{\fontseries{b}\selectfont}

\renewcommand{\arraystretch}{1.2}

\resizebox{\textwidth}{!}{%
\begin{tabular}{@{} l l l @{\hskip 1mm} l l l @{\hskip 1mm} l l l @{}}
	\toprule
	Benchmark		& Grid		& Random		& \name{RoBO}	& \name{BTB}	& \name{hyperopt}	& \name{SMAC}	& \name{BOHB}	& \name{Optunity}	\\
	\midrule

	Levy			& 0.00089	& 0.00102		& \B 0.00000	& 0.19588		& 0.00010			& 0.00058		& 0.02430		& 0.00013			\\
	Branin			& 0.24665	& 0.28982		& \B 0.00065	& \ul{0.00077}	& 0.05011			& 0.10191		& 0.39143		& 0.03356			\\
	Hartmann6		& 1.04844	& 0.66960		& \B 0.06575	& 0.27107		& 0.44905			& 0.27262		& 0.35435		& \ul{0.22289}		\\
	Rosenbrock10	& 9.00000	& 45.8354		& \B 4.43552	& 19.4919		& 22.4746			& 38.1581		& 34.4457		& 36.3984			\\
	Camelback		& 0.94443	& 0.45722		& \ul{0.02871}	& \ul{0.07745}	& 0.07594			& 0.18440		& 0.38247		& \B 0.01754		\\

	\bottomrule
\end{tabular}}

\caption{
	Results of all tested \ac{CASH} solvers after \(100\) iterations. For each synthetic benchmark the mean performance over \(10\) trials is reported. Bold face represents the best mean value for each benchmark. Results not significantly worse than the best result---according to a Wilcoxon signed-rank test---are underlined.
}
\label{tbl:results_cash_solver}

\end{table}

\section{Evaluated Data Sets}
\label{app:evaluated_data_sets}

\begin{footnotesize}

\renewcommand{\arraystretch}{0.9}
\newcommand{\rotation}{60}

\begin{longtable}{@{} l @{\hskip 0mm} r @{\hskip 5mm} l l l l l l r @{}}
	\toprule
	\rotatebox{\rotation}{Data Set} &	& \rotatebox{\rotation}{Classes} & \rotatebox{\rotation}{Samples} & \rotatebox{\rotation}{Numeric Feat.} & \rotatebox{\rotation}{Categorical Feat.} & \rotatebox{\rotation}{Missing Values} & \rotatebox{\rotation}{Incom. Samples} & \rotatebox{\rotation}{Minority \%} \\
	\midrule

kr-vs-kp        & (3) &        2 &     3196 &        0 &       37 &        0 &        0 & 47.78 \\ 
letter          & (6) &       26 &    20000 &       16 &        1 &        0 &        0 & 3.67 \\ 
balance-scale   & (11) &        3 &      625 &        4 &        1 &        0 &        0 & 7.84 \\ 
mfeat-factors   & (12) &       10 &     2000 &      216 &        1 &        0 &        0 & 10.00 \\ 
mfeat-fourier   & (14) &       10 &     2000 &       76 &        1 &        0 &        0 & 10.00 \\ 
breast-w        & (15) &        2 &      699 &        9 &        1 &       16 &       16 & 34.48 \\ 
mfeat-karhunen  & (16) &       10 &     2000 &       64 &        1 &        0 &        0 & 10.00 \\ 
mfeat-morpholog & (18) &       10 &     2000 &        6 &        1 &        0 &        0 & 10.00 \\ 
mfeat-pixel     & (20) &       10 &     2000 &        0 &      241 &        0 &        0 & 10.00 \\ 
car             & (21) &        4 &     1728 &        0 &        7 &        0 &        0 & 3.76 \\ 
mfeat-zernike   & (22) &       10 &     2000 &       47 &        1 &        0 &        0 & 10.00 \\ 
cmc             & (23) &        3 &     1473 &        2 &        8 &        0 &        0 & 22.61 \\ 
mushroom        & (24) &        2 &     8124 &        0 &       23 &     2480 &     2480 & 48.20 \\ 
optdigits       & (28) &       10 &     5620 &       64 &        1 &        0 &        0 & 9.86 \\ 
credit-approval & (29) &        2 &      690 &        6 &       10 &       67 &       37 & 44.49 \\ 
credit-g        & (31) &        2 &     1000 &        7 &       14 &        0 &        0 & 30.00 \\ 
pendigits       & (32) &       10 &    10992 &       16 &        1 &        0 &        0 & 9.60 \\ 
segment         & (36) &        7 &     2310 &       19 &        1 &        0 &        0 & 14.29 \\ 
diabetes        & (37) &        2 &      768 &        8 &        1 &        0 &        0 & 34.90 \\ 
sick            & (38) &        2 &     3772 &        7 &       23 &     6064 &     3772 & 6.12 \\ 
soybean         & (42) &       19 &      683 &        0 &       36 &     2337 &      121 & 1.17 \\ 
spambase        & (44) &        2 &     4601 &       57 &        1 &        0 &        0 & 39.40 \\ 
splice          & (46) &        3 &     3190 &        0 &       61 &        0 &        0 & 24.04 \\ 
tic-tac-toe     & (50) &        2 &      958 &        0 &       10 &        0 &        0 & 34.66 \\ 
vehicle         & (54) &        4 &      846 &       18 &        1 &        0 &        0 & 23.52 \\ 
waveform-5000   & (60) &        3 &     5000 &       40 &        1 &        0 &        0 & 33.06 \\ 
electricity     & (151) &        2 &    45312 &        7 &        2 &        0 &        0 & 42.45 \\ 
satimage        & (182) &        6 &     6430 &       36 &        1 &        0 &        0 & 9.72 \\ 
eucalyptus      & (188) &        5 &      736 &       14 &        6 &      448 &       95 & 14.27 \\ 
isolet          & (300) &       26 &     7797 &      617 &        1 &        0 &        0 & 3.82 \\ 
vowel           & (307) &       11 &      990 &       10 &        3 &        0 &        0 & 9.09 \\ 
scene           & (312) &        2 &     2407 &      294 &        6 &        0 &        0 & 17.91 \\ 
monks-problems- & (333) &        2 &      556 &        0 &        7 &        0 &        0 & 50.00 \\ 
monks-problems- & (334) &        2 &      601 &        0 &        7 &        0 &        0 & 34.28 \\ 
monks-problems- & (335) &        2 &      554 &        0 &        7 &        0 &        0 & 48.01 \\ 
JapaneseVowels  & (375) &        9 &     9961 &       14 &        1 &        0 &        0 & 7.85 \\ 
synthetic\_contr & (377) &        6 &      600 &       60 &        2 &        0 &        0 & 16.67 \\ 
irish           & (451) &        2 &      500 &        2 &        4 &       32 &       32 & 44.40 \\ 
analcatdata\_aut & (458) &        4 &      841 &       70 &        1 &        0 &        0 & 6.54 \\ 
analcatdata\_dmf & (469) &        6 &      797 &        0 &        5 &        0 &        0 & 15.43 \\ 
profb           & (470) &        2 &      672 &        5 &        5 &     1200 &      666 & 33.33 \\ 
collins         & (478) &       15 &      500 &       20 &        4 &        0 &        0 & 1.20 \\ 
mnist\_784      & (554) &       10 &    70000 &      784 &        1 &        0 &        0 & 9.02 \\ 
sylva\_agnostic & (1036) &        2 &    14395 &      216 &        1 &        0 &        0 & 6.15 \\ 
gina\_agnostic  & (1038) &        2 &     3468 &      970 &        1 &        0 &        0 & 49.16 \\ 
ada\_agnostic   & (1043) &        2 &     4562 &       48 &        1 &        0 &        0 & 24.81 \\ 
mozilla4        & (1046) &        2 &    15545 &        5 &        1 &        0 &        0 & 32.86 \\ 
pc4             & (1049) &        2 &     1458 &       37 &        1 &        0 &        0 & 12.21 \\ 
pc3             & (1050) &        2 &     1563 &       37 &        1 &        0 &        0 & 10.24 \\ 
jm1             & (1053) &        2 &    10885 &       21 &        1 &       25 &        5 & 19.35 \\ 
kc2             & (1063) &        2 &      522 &       21 &        1 &        0 &        0 & 20.50 \\ 
kc1             & (1067) &        2 &     2109 &       21 &        1 &        0 &        0 & 15.46 \\ 
pc1             & (1068) &        2 &     1109 &       21 &        1 &        0 &        0 & 6.94 \\ 
KDDCup09\_appete & (1111) &        2 &    50000 &      192 &       39 &  8024152 &    50000 & 1.78 \\ 
KDDCup09\_churn & (1112) &        2 &    50000 &      192 &       39 &  8024152 &    50000 & 7.34 \\ 
KDDCup09\_upsell & (1114) &        2 &    50000 &      192 &       39 &  8024152 &    50000 & 7.36 \\ 
MagicTelescope  & (1120) &        2 &    19020 &       11 &        1 &        0 &        0 & 35.16 \\ 
airlines        & (1169) &        2 &   539383 &        3 &        5 &        0 &        0 & 44.54 \\ 
artificial-char & (1459) &       10 &    10218 &        7 &        1 &        0 &        0 & 5.87 \\ 
bank-marketing  & (1461) &        2 &    45211 &        7 &       10 &        0 &        0 & 11.70 \\ 
banknote-authen & (1462) &        2 &     1372 &        4 &        1 &        0 &        0 & 44.46 \\ 
blood-transfusi & (1464) &        2 &      748 &        4 &        1 &        0 &        0 & 23.80 \\ 
cardiotocograph & (1466) &       10 &     2126 &       35 &        1 &        0 &        0 & 2.49 \\ 
climate-model-s & (1467) &        2 &      540 &       20 &        1 &        0 &        0 & 8.52 \\ 
cnae-9          & (1468) &        9 &     1080 &      856 &        1 &        0 &        0 & 11.11 \\ 
eeg-eye-state   & (1471) &        2 &    14980 &       14 &        1 &        0 &        0 & 44.88 \\ 
first-order-the & (1475) &        6 &     6118 &       51 &        1 &        0 &        0 & 7.94 \\ 
gas-drift       & (1476) &        6 &    13910 &      128 &        1 &        0 &        0 & 11.80 \\ 
har             & (1478) &        6 &    10299 &      561 &        1 &        0 &        0 & 13.65 \\ 
hill-valley     & (1479) &        2 &     1212 &      100 &        1 &        0 &        0 & 50.00 \\ 
ilpd            & (1480) &        2 &      583 &        9 &        2 &        0 &        0 & 28.64 \\ 
madelon         & (1485) &        2 &     2600 &      500 &        1 &        0 &        0 & 50.00 \\ 
nomao           & (1486) &        2 &    34465 &       89 &       30 &        0 &        0 & 28.56 \\ 
ozone-level-8hr & (1487) &        2 &     2534 &       72 &        1 &        0 &        0 & 6.31 \\ 
phoneme         & (1489) &        2 &     5404 &        5 &        1 &        0 &        0 & 29.35 \\ 
one-hundred-pla & (1491) &      100 &     1600 &       64 &        1 &        0 &        0 & 1.00 \\ 
one-hundred-pla & (1492) &      100 &     1600 &       64 &        1 &        0 &        0 & 1.00 \\ 
one-hundred-pla & (1493) &      100 &     1599 &       64 &        1 &        0 &        0 & 0.94 \\ 
qsar-biodeg     & (1494) &        2 &     1055 &       41 &        1 &        0 &        0 & 33.74 \\ 
wall-robot-navi & (1497) &        4 &     5456 &       24 &        1 &        0 &        0 & 6.01 \\ 
semeion         & (1501) &       10 &     1593 &      256 &        1 &        0 &        0 & 9.73 \\ 
steel-plates-fa & (1504) &        2 &     1941 &       33 &        1 &        0 &        0 & 34.67 \\ 
tamilnadu-elect & (1505) &       20 &    45781 &        2 &        2 &        0 &        0 & 3.05 \\ 
wdbc            & (1510) &        2 &      569 &       30 &        1 &        0 &        0 & 37.26 \\ 
micro-mass      & (1515) &       20 &      571 &     1300 &        1 &        0 &        0 & 1.93 \\ 
wilt            & (1570) &        2 &     4839 &        5 &        1 &        0 &        0 & 5.39 \\ 
adult           & (1590) &        2 &    48842 &        6 &        9 &     6465 &     3620 & 23.93 \\ 
covertype       & (1596) &        7 &   581012 &       10 &       45 &        0 &        0 & 0.47 \\ 
Bioresponse     & (4134) &        2 &     3751 &     1776 &        1 &        0 &        0 & 45.77 \\ 
Bioresponse     & (4134) &        2 &     3751 &     1776 &        1 &        0 &        0 & 45.77 \\ 
Amazon\_employee & (4135) &        2 &    32769 &        0 &       10 &        0 &        0 & 5.79 \\ 
PhishingWebsite & (4534) &        2 &    11055 &        0 &       31 &        0 &        0 & 44.31 \\ 
PhishingWebsite & (4534) &        2 &    11055 &        0 &       31 &        0 &        0 & 44.31 \\ 
GesturePhaseSeg & (4538) &        5 &     9873 &       32 &        1 &        0 &        0 & 10.11 \\ 
MiceProtein     & (4550) &        8 &     1080 &       77 &        5 &     1396 &      528 & 9.72 \\ 
cylinder-bands  & (6332) &        2 &      540 &       18 &       22 &      999 &      263 & 42.22 \\ 
cylinder-bands  & (6332) &        2 &      540 &       18 &       22 &      999 &      263 & 42.22 \\ 
cjs             & (23380) &        6 &     2796 &       32 &        3 &    68100 &     2795 & 9.80 \\ 
dresses-sales   & (23381) &        2 &      500 &        1 &       12 &      835 &      401 & 42.00 \\ 
higgs           & (23512) &        2 &    98050 &       28 &        1 &        9 &        1 & 47.14 \\ 
numerai28.6     & (23517) &        2 &    96320 &       21 &        1 &        0 &        0 & 49.48 \\ 
LED-display-dom & (40496) &       10 &      500 &        7 &        1 &        0 &        0 & 7.40 \\ 
texture         & (40499) &       11 &     5500 &       40 &        1 &        0 &        0 & 9.09 \\ 
Australian      & (40509) &        2 &      690 &       14 &        1 &        0 &        0 & 44.49 \\ 
SpeedDating     & (40536) &        2 &     8378 &       59 &       64 &    18372 &     7330 & 16.47 \\ 
connect-4       & (40668) &        3 &    67557 &        0 &       43 &        0 &        0 & 9.55 \\ 
dna             & (40670) &        3 &     3186 &        0 &      181 &        0 &        0 & 24.01 \\ 
shuttle         & (40685) &        7 &    58000 &        9 &        1 &        0 &        0 & 0.02 \\ 
churn           & (40701) &        2 &     5000 &       16 &        5 &        0 &        0 & 14.14 \\ 
Devnagari-Scrip & (40923) &       46 &    92000 &     1024 &        1 &        0 &        0 & 2.17 \\ 
CIFAR\_10       & (40927) &       10 &    60000 &     3072 &        1 &        0 &        0 & 10.00 \\ 
MiceProtein     & (40966) &        8 &     1080 &       77 &        5 &     1396 &      528 & 9.72 \\ 
car             & (40975) &        4 &     1728 &        0 &        7 &        0 &        0 & 3.76 \\ 
Internet-Advert & (40978) &        2 &     3279 &        3 &     1556 &        0 &        0 & 14.00 \\ 
mfeat-pixel     & (40979) &       10 &     2000 &      240 &        1 &        0 &        0 & 10.00 \\ 
Australian      & (40981) &        2 &      690 &        6 &        9 &        0 &        0 & 44.49 \\ 
steel-plates-fa & (40982) &        7 &     1941 &       27 &        1 &        0 &        0 & 2.83 \\ 
wilt            & (40983) &        2 &     4839 &        5 &        1 &        0 &        0 & 5.39 \\ 
segment         & (40984) &        7 &     2310 &       19 &        1 &        0 &        0 & 14.29 \\ 
climate-model-s & (40994) &        2 &      540 &       20 &        1 &        0 &        0 & 8.52 \\ 
Fashion-MNIST   & (40996) &       10 &    70000 &      784 &        1 &        0 &        0 & 10.00 \\ 
jungle\_chess\_2p & (41027) &        3 &    44819 &        6 &        1 &        0 &        0 & 9.67 \\ 
APSFailure      & (41138) &        2 &    76000 &      170 &        1 &  1078695 &    75244 & 1.81 \\ 
christine       & (41142) &        2 &     5418 &     1599 &       38 &        0 &        0 & 50.00 \\ 
jasmine         & (41143) &        2 &     2984 &        8 &      137 &        0 &        0 & 50.00 \\ 
sylvine         & (41146) &        2 &     5124 &       20 &        1 &        0 &        0 & 50.00 \\ 
albert          & (41147) &        2 &   425240 &       26 &       53 &  2734000 &   425159 & 50.00 \\ 
MiniBooNE       & (41150) &        2 &   130064 &       50 &        1 &        0 &        0 & 28.06 \\ 
guillermo       & (41159) &        2 &    20000 &     4296 &        1 &        0 &        0 & 40.02 \\ 
riccardo        & (41161) &        2 &    20000 &     4296 &        1 &        0 &        0 & 25.00 \\ 
dilbert         & (41163) &        5 &    10000 &     2000 &        1 &        0 &        0 & 19.13 \\ 
fabert          & (41164) &        7 &     8237 &      800 &        1 &        0 &        0 & 6.09 \\ 
robert          & (41165) &       10 &    10000 &     7200 &        1 &        0 &        0 & 9.58 \\ 
volkert         & (41166) &       10 &    58310 &      180 &        1 &        0 &        0 & 2.33 \\ 
dionis          & (41167) &      355 &   416188 &       60 &        1 &        0 &        0 & 0.21 \\ 
jannis          & (41168) &        4 &    83733 &       54 &        1 &        0 &        0 & 2.01 \\ 
helena          & (41169) &      100 &    65196 &       27 &        1 &        0 &        0 & 0.17 \\ 

	\bottomrule
	
	\caption{
	List of all tested data sets. Listed are the (abbreviated) name and \name{OpenML} id for each data set together with the number of classes, the number of samples, the number of numeric and categorical features per sample, how many values are missing in total (Missing values), how many samples contain at least one missing value (Incomp. Samples) and the percentage of samples belonging to the least frequent class (Minority \%).
}
\end{longtable}
\end{footnotesize}

\section{Configuration Space for CASH Solvers}
\label{app:complete_config_space}

{
\renewcommand{\arraystretch}{0.9}
\begin{longtable}{@{} l l l l @{}}
	\toprule
	Classifier & Hyperparameter & Type & Values \\
	\midrule
	Bernoulli na\"ive Bayes		& alpha & con & \([0.01, 100]\) \\
								& fit\_prior & cat & [false, true] \\
	\hline
	Multinomial na\"ive Bayes	& alpha & con & \([0.01, 100]\) \\
								& fit\_prior & cat & [false, true] \\
	\hline
	Decision Tree				& criterion & cat & [entropy, gini] \\
								& max\_depth & int & [1, 10] \\
								& min\_samples\_leaf & int & [1, 20] \\
								& min\_samples\_split & int & [2, 20] \\
	\hline
	Extra Trees					& bootstrap & cat & [false, true] \\
								& criterion & cat & [entropy, gini] \\
								& max\_features & con & [0.0, 1.0] \\
								& min\_samples\_leaf & int & [1, 20] \\
								& min\_samples\_split & int & [2, 20] \\
	\hline
	Gradient Boosting			& learning\_rate & con & [0.01, 1.0] \\
								& criterion & cat & [friedman\_mse, mae, mse] \\
								& max\_depth & int & [1, 10] \\
								& min\_samples\_split & int & [2, 20] \\
								& min\_samples\_leaf & int & [1, 20] \\
								& n\_estimators & int & [50, 500] \\
	\hline
	Random Forest				& bootstrap & cat & [false, true] \\
								& criterion & cat & [entropy, gini] \\
								& max\_features & con & [0.0, 1.0] \\
								& min\_samples\_split & int & [2, 20] \\
								& min\_samples\_leaf & int & [1, 20] \\
								& n\_estimators & int & [2, 100] \\
	\hline
	\(k\) Nearest Neighbors		& n\_neighbors & int & [1, 100] \\
								& p & int & [1, 2] \\
								& weights & cat & [distance, uniform] \\
	\hline
	LDA							& n\_components & cat & [1, 250] \\
								& shrinkage & con & [0.0, 1.0] \\
								& solver & cat & [eigen, lsgr, svd] \\
								& tol & con & [0.00001, 0.1] \\
	\hline
	QDA							& reg\_param & con & [0.0, 1.0] \\
	\hline
	Linear SVM					& C & con & [0.01, 10000] \\
								& loss & cat & [hinge, squared\_hinge] \\
								& penalty & cat & [l1, l2] \\
								& tol & con & [0.00001, 0.1] \\
	\hline
	Kernel SVM					& C & con & [0.01, 10000] \\
								& coef0 & con & [-1, 1] \\
								& degree & int & [2, 5] \\
								& gamma & con & [1, 10000] \\
								& kernel & cat & [poly, rbf, sigmoid] \\
								& shrinking & cat & [false, true] \\
								& tol & con & [0.00001, 0.1] \\
	\hline
	Passive Aggressive			& average & cat & [false, true] \\
								& C & con & [0.00001, 10] \\
								& loss & cat & [hinge, squared\_hinge] \\
								& tol & con & [0.00001, 0.1] \\
	\hline
	SGD							& alpha & con & \([0.0000001, 0.1]\) \\
								& average & cat & [false, true] \\
								& epsilon & con & [0.00001, 0.1] \\
								& eta0 & con & \([0.0000001, 0.11]\) \\
								& learning\_rate & cat & [constant, invscaling, optimal] \\
								& loss & cat & [hinge, log, modified\_huber] \\ 
								& l1\_ratio & con & \([0.0000001, 1]\) \\
								& penalty & cat & [elasticnet, l1, l2] \\
								& power\_t & con & [0.00001, 1] \\
								& tol & con & [0.00001, 0.1] \\
	
	\bottomrule
	
	\caption{
	Complete configuration space used for \ac{CASH} benchmarking. Hyperparameter names equal the used names in \name{scikit-learn}. \emph{cat} are categorical, \emph{con} are continuous and \emph{int} integer hyperparameters.
}
\end{longtable}
}

\section{Raw Experiment Results}
\label{app:raw_experiment_results}

\begin{footnotesize}
\centering

\newrobustcmd{\B}{\fontseries{b}\selectfont}

\renewcommand{\arraystretch}{0.9}

\begin{longtable}{@{} l l l l l @{\hskip 2mm} l l l @{\hskip 2mm} l @{\hskip 2mm} l l @{}}
	\toprule
	Data Set	& Dummy	& RF	& Grid	& Random	& SMAC	& BOHB	& Optunity	& hyperopt	& RoBO	& BTB	\\ 
	\midrule

\(3^+\) 	&	0.4991 	&	0.9830 	&	0.8488 	&	\ul{0.9985} 	&	\ul{0.9983} 	&	\ul{0.9980} 	&	0.9979 	&	\B 0.9989 	&	0.9975 	&	\ul{0.9979} 	\\ 
\(6\) 	&	0.0396 	&	0.9315 	&	0.5482 	&	0.9471 	&	\B 0.9613 	&	\ul{0.9525} 	&	0.9459 	&	\ul{0.9609} 	&	0.9438 	&	0.9472 	\\ 
\(11\) 	&	0.4394 	&	0.8170 	&	0.8718 	&	0.9920 	&	0.9867 	&	0.9473 	&	0.9660 	&	\B 1.0000 	&	0.9862 	&	\ul{0.9957} 	\\ 
\(12^+\) 	&	0.0997 	&	0.9468 	&	0.8542 	&	\ul{0.9808} 	&	\B 0.9835 	&	0.9818 	&	0.9800 	&	\ul{0.9832} 	&	\ul{0.9833} 	&	\ul{0.9807} 	\\ 
\(14\) 	&	0.1065 	&	0.7940 	&	0.7498 	&	\ul{0.8613} 	&	0.8560 	&	0.8485 	&	\ul{0.8625} 	&	\B 0.8678 	&	\ul{0.8635} 	&	\ul{0.8612} 	\\ 
\(16\) 	&	0.0982 	&	0.8955 	&	0.8442 	&	\ul{0.9825} 	&	\ul{0.9815} 	&	\ul{0.9798} 	&	\ul{0.9793} 	&	\B 0.9827 	&	\ul{0.9813} 	&	\ul{0.9807} 	\\ 
\(18\) 	&	0.0988 	&	0.7073 	&	0.6788 	&	\ul{0.7370} 	&	\ul{0.7443} 	&	\ul{0.7470} 	&	\ul{0.7378} 	&	\B 0.7478 	&	0.7303 	&	\ul{0.7343} 	\\ 
\(20\) 	&	0.1023 	&	0.9512 	&	0.9212 	&	\ul{0.9838} 	&	\ul{0.9843} 	&	\ul{0.9832} 	&	\ul{0.9823} 	&	\B 0.9855 	&	\ul{0.9823} 	&	0.9783 	\\ 
\(21\) 	&	0.5414 	&	0.9536 	&	0.7582 	&	0.9961 	&	0.9940 	&	0.9771 	&	\B 0.9988 	&	\ul{0.9965} 	&	0.9882 	&	0.9821 	\\ 
\(22\) 	&	0.0995 	&	0.7455 	&	0.7050 	&	0.8367 	&	0.8360 	&	0.8272 	&	0.8345 	&	\ul{0.8463} 	&	\B 0.8503 	&	\ul{0.8402} 	\\ 
\(23^+\) 	&	0.3597 	&	0.5043 	&	0.5063 	&	0.5647 	&	0.5622 	&	0.5656 	&	\ul{0.5636} 	&	\B 0.5853 	&	0.5695 	&	0.5624 	\\ 
\(28\) 	&	0.0992 	&	0.9607 	&	0.9057 	&	\ul{0.9898} 	&	\B 0.9906 	&	\ul{0.9898} 	&	\ul{0.9897} 	&	\ul{0.9900} 	&	\ul{0.9901} 	&	\ul{0.9902} 	\\ 
\(31^+\) 	&	0.5837 	&	0.7043 	&	0.7053 	&	\ul{0.7690} 	&	\ul{0.7697} 	&	\ul{0.7610} 	&	\ul{0.7743} 	&	\B 0.7753 	&	\ul{0.7617} 	&	\ul{0.7593} 	\\ 
\(32\) 	&	0.1006 	&	0.9847 	&	0.8008 	&	0.9925 	&	\ul{0.9938} 	&	\ul{0.9933} 	&	0.9924 	&	\B 0.9939 	&	\ul{0.9936} 	&	\ul{0.9933} 	\\ 
\(36\) 	&	0.1414 	&	0.9694 	&	0.4338 	&	\ul{0.9818} 	&	\ul{0.9818} 	&	0.9746 	&	\ul{0.9838} 	&	\B 0.9857 	&	0.9788 	&	0.9794 	\\ 
\(37\) 	&	0.5403 	&	0.7385 	&	0.6489 	&	0.7762 	&	\ul{0.7883} 	&	\ul{0.7827} 	&	\ul{0.7823} 	&	\B 0.7996 	&	\ul{0.7861} 	&	\ul{0.7840} 	\\ 
\(44\) 	&	0.5206 	&	0.9411 	&	0.8888 	&	\ul{0.9552} 	&	\ul{0.9542} 	&	0.9505 	&	\ul{0.9566} 	&	\B 0.9581 	&	0.9503 	&	0.9511 	\\ 
\(46\) 	&	0.3814 	&	0.9106 	&	0.8361 	&	0.9580 	&	0.9580 	&	0.9529 	&	\ul{0.9619} 	&	\B 0.9654 	&	0.9479 	&	0.9595 	\\ 
\(50\) 	&	0.5354 	&	0.9128 	&	0.6451 	&	\B 1.0000 	&	\ul{0.9983} 	&	0.9778 	&	\ul{0.9972} 	&	\B 1.0000 	&	\ul{0.9962} 	&	\ul{0.9979} 	\\ 
\(54^+\) 	&	0.2492 	&	0.7287 	&	0.4307 	&	\ul{0.8413} 	&	\ul{0.8406} 	&	0.8260 	&	\ul{0.8362} 	&	\ul{0.8516} 	&	\B 0.8594 	&	0.8094 	\\ 
\(60\) 	&	0.3369 	&	0.8136 	&	0.7111 	&	0.8692 	&	\ul{0.8709} 	&	\ul{0.8696} 	&	\B 0.8713 	&	\ul{0.8701} 	&	\ul{0.8697} 	&	\ul{0.8697} 	\\ 
\(151\) 	&	0.5106 	&	0.8863 	&	0.5935 	&	0.9275 	&	0.9183 	&	0.9125 	&	\ul{0.9302} 	&	\B 0.9377 	&	0.8852 	&	0.9303 	\\ 
\(182\) 	&	0.1923 	&	0.8966 	&	0.7091 	&	0.9138 	&	\ul{0.9171} 	&	0.9125 	&	\B 0.9186 	&	\ul{0.9164} 	&	0.9073 	&	\ul{0.9136} 	\\ 
\(300\) 	&	0.0370 	&	0.8979 	&	0.8432 	&	\ul{0.9676} 	&	\ul{0.9683} 	&	\ul{0.9683} 	&	0.9654 	&	\B 0.9718 	&	0.9578 	&	\ul{0.9705} 	\\ 
\(307\) 	&	0.0882 	&	0.9000 	&	0.2633 	&	0.9690 	&	\ul{0.9822} 	&	\ul{0.9737} 	&	0.9731 	&	0.9704 	&	\B 0.9902 	&	0.9764 	\\ 
\(312\) 	&	0.7105 	&	0.8874 	&	0.9303 	&	\ul{0.9881} 	&	\ul{0.9881} 	&	\ul{0.9881} 	&	\ul{0.9876} 	&	\B 0.9906 	&	\ul{0.9893} 	&	\ul{0.9905} 	\\ 
\(333\) 	&	0.4934 	&	0.9641 	&	0.7413 	&	\B 1.0000 	&	\B 1.0000 	&	\B 1.0000 	&	\B 1.0000 	&	\B 1.0000 	&	\B 1.0000 	&	\B 1.0000 	\\ 
\(334\) 	&	0.5464 	&	0.8597 	&	0.6497 	&	0.9923 	&	0.9818 	&	0.9193 	&	\ul{0.9917} 	&	\B 1.0000 	&	\ul{0.9934} 	&	0.9923 	\\ 
\(335\) 	&	0.4976 	&	0.9695 	&	0.7431 	&	\ul{0.9874} 	&	\ul{0.9868} 	&	\ul{0.9838} 	&	\ul{0.9868} 	&	\B 0.9898 	&	\B 0.9898 	&	\ul{0.9850} 	\\ 
\(375\) 	&	0.1144 	&	0.9472 	&	0.4545 	&	0.9677 	&	\B 0.9849 	&	0.9664 	&	0.9733 	&	\ul{0.9791} 	&	0.9686 	&	0.9706 	\\ 
\(377\) 	&	0.1689 	&	0.9522 	&	0.1706 	&	0.9928 	&	\ul{0.9944} 	&	\ul{0.9928} 	&	0.9922 	&	\ul{0.9956} 	&	\B 0.9967 	&	\ul{0.9900} 	\\ 
\(458\) 	&	0.3229 	&	0.9830 	&	0.9783 	&	0.9976 	&	\ul{0.9988} 	&	\ul{0.9984} 	&	\ul{0.9984} 	&	\B 0.9992 	&	\ul{0.9988} 	&	\ul{0.9988} 	\\ 
\(469^-\) 	&	0.1692 	&	0.1896 	&	0.2325 	&	\ul{0.2579} 	&	\ul{0.2612} 	&	\ul{0.2650} 	&	\ul{0.2621} 	&	\B 0.2692 	&	\ul{0.2596} 	&	\ul{0.2633} 	\\ 
\(478\) 	&	0.0893 	&	0.7187 	&	0.6093 	&	\ul{0.9987} 	&	\ul{0.9920} 	&	0.9747 	&	0.9867 	&	\B 1.0000 	&	\ul{0.9953} 	&	\ul{0.9920} 	\\ 
\(554\) 	&	0.1010 	&	0.9442 	&	0.8331 	&	0.9477 	&	0.9445 	&	0.9376 	&	0.9357 	&	\B 0.9578 	&	0.9403 	&	0.9468 	\\ 
\(1036\) 	&	0.8842 	&	0.9871 	&	0.9911 	&	\ul{0.9950} 	&	0.9948 	&	0.9944 	&	\B 0.9952 	&	\ul{0.9948} 	&	\ul{0.9945} 	&	\ul{0.9941} 	\\ 
\(1038\) 	&	0.5014 	&	0.9065 	&	0.8012 	&	0.9376 	&	0.9375 	&	0.9335 	&	\ul{0.9423} 	&	\B 0.9516 	&	0.9302 	&	0.9418 	\\ 
\(1043\) 	&	0.6270 	&	0.8297 	&	0.7879 	&	\ul{0.8521} 	&	\ul{0.8524} 	&	\ul{0.8500} 	&	\ul{0.8517} 	&	\ul{0.8565} 	&	0.8486 	&	\B 0.8568 	\\ 
\(1046\) 	&	0.5582 	&	0.9492 	&	0.9353 	&	\ul{0.9583} 	&	\ul{0.9580} 	&	0.9533 	&	\ul{0.9583} 	&	\B 0.9605 	&	0.9538 	&	0.9555 	\\ 
\(1049\) 	&	0.7779 	&	0.8975 	&	0.8747 	&	\ul{0.9178} 	&	\ul{0.9185} 	&	\ul{0.9153} 	&	\ul{0.9187} 	&	\B 0.9235 	&	\ul{0.9121} 	&	\ul{0.9151} 	\\ 
\(1050\) 	&	0.8158 	&	0.8893 	&	0.8663 	&	\ul{0.9053} 	&	\ul{0.9068} 	&	\ul{0.9053} 	&	\ul{0.9053} 	&	\B 0.9100 	&	\ul{0.8983} 	&	\ul{0.9051} 	\\ 
\(1063\) 	&	0.6828 	&	0.8127 	&	0.8299 	&	\ul{0.8669} 	&	\B 0.8707 	&	\ul{0.8650} 	&	\ul{0.8688} 	&	\ul{0.8669} 	&	\ul{0.8643} 	&	\ul{0.8586} 	\\ 
\(1067^+\) 	&	0.7409 	&	0.8504 	&	0.8509 	&	\ul{0.8649} 	&	\ul{0.8660} 	&	\ul{0.8621} 	&	\ul{0.8640} 	&	\ul{0.8687} 	&	\ul{0.8657} 	&	\B 0.8727 	\\ 
\(1068\) 	&	0.8670 	&	\ul{0.9330} 	&	0.9261 	&	\ul{0.9396} 	&	\ul{0.9402} 	&	\ul{0.9363} 	&	\ul{0.9381} 	&	\ul{0.9432} 	&	\B 0.9438 	&	\ul{0.9372} 	\\ 
\(1120\) 	&	0.5455 	&	0.8664 	&	0.6491 	&	\ul{0.8790} 	&	\ul{0.8797} 	&	\ul{0.8766} 	&	\ul{0.8802} 	&	\B 0.8819 	&	0.8714 	&	\ul{0.8794} 	\\ 
\(1169^-\) 	&	0.5060 	&	0.6144 	&	0.5545 	&	\ul{0.6650} 	&	\ul{0.6655} 	&	\ul{0.6635} 	&	0.6639 	&	\B 0.6655 	&	0.6627 	&	\ul{0.6627} 	\\ 
\(1459\) 	&	0.1017 	&	0.8557 	&	0.2446 	&	0.8834 	&	0.8631 	&	0.8315 	&	\B 0.9303 	&	0.9023 	&	0.8623 	&	0.8973 	\\ 
\(1461^+\) 	&	0.7935 	&	0.8991 	&	0.8687 	&	\ul{0.9079} 	&	\ul{0.9078} 	&	0.9070 	&	\B 0.9084 	&	\ul{0.9071} 	&	0.9052 	&	0.9044 	\\ 
\(1462\) 	&	0.5056 	&	0.9925 	&	0.8451 	&	\B 1.0000 	&	\B 1.0000 	&	\B 1.0000 	&	\ul{0.9995} 	&	\B 1.0000 	&	\B 1.0000 	&	\ul{0.9995} 	\\ 
\(1464^-\) 	&	0.6418 	&	0.7329 	&	0.7676 	&	\ul{0.7978} 	&	\ul{0.7973} 	&	\ul{0.7951} 	&	\ul{0.7938} 	&	\ul{0.8009} 	&	\B 0.8076 	&	\ul{0.7991} 	\\ 
\(1466\) 	&	0.1530 	&	0.9983 	&	\B 1.0000 	&	\B 1.0000 	&	\B 1.0000 	&	\B 1.0000 	&	\B 1.0000 	&	\B 1.0000 	&	\B 1.0000 	&	\B 1.0000 	\\ 
\(1467\) 	&	0.8438 	&	0.9037 	&	0.9111 	&	\ul{0.9179} 	&	\ul{0.9198} 	&	\ul{0.9167} 	&	\ul{0.9173} 	&	\B 0.9284 	&	\ul{0.9204} 	&	\ul{0.9247} 	\\ 
\(1468^+\) 	&	0.1139 	&	0.8985 	&	\ul{0.9586} 	&	0.9571 	&	\B 0.9630 	&	\ul{0.9614} 	&	0.9562 	&	\ul{0.9599} 	&	\ul{0.9617} 	&	\ul{0.9537} 	\\ 
\(1471\) 	&	0.5074 	&	0.8915 	&	0.5519 	&	0.9522 	&	\B 0.9741 	&	\ul{0.9729} 	&	0.9541 	&	\ul{0.9726} 	&	0.9414 	&	0.9459 	\\ 
\(1475^+\) 	&	0.2441 	&	0.5822 	&	0.3670 	&	\ul{0.6082} 	&	0.6003 	&	0.5969 	&	0.6068 	&	\B 0.6209 	&	0.6031 	&	0.5984 	\\ 
\(1476\) 	&	0.1773 	&	0.9919 	&	0.2300 	&	0.9927 	&	0.9931 	&	0.9907 	&	0.9920 	&	\B 0.9948 	&	0.9933 	&	0.9912 	\\ 
\(1478\) 	&	0.1684 	&	0.9650 	&	0.8509 	&	0.9893 	&	\ul{0.9908} 	&	\ul{0.9896} 	&	0.9857 	&	\B 0.9916 	&	0.9873 	&	0.9885 	\\ 
\(1479\) 	&	0.5074 	&	0.5459 	&	0.7857 	&	0.9354 	&	\ul{0.9558} 	&	\B 0.9566 	&	0.9321 	&	\ul{0.9492} 	&	\ul{0.9511} 	&	0.9431 	\\ 
\(1480\) 	&	0.5909 	&	0.7034 	&	0.7069 	&	\ul{0.7354} 	&	\ul{0.7394} 	&	\ul{0.7383} 	&	\ul{0.7400} 	&	\B 0.7550 	&	\ul{0.7417} 	&	\ul{0.7469} 	\\ 
\(1485\) 	&	0.4991 	&	0.6191 	&	0.5922 	&	\ul{0.8351} 	&	0.8340 	&	0.8232 	&	0.8171 	&	\B 0.8484 	&	0.8194 	&	0.8367 	\\ 
\(1486^-\) 	&	0.5927 	&	0.9640 	&	0.8404 	&	0.9662 	&	0.9645 	&	0.9655 	&	0.9655 	&	\B 0.9683 	&	0.9634 	&	0.9646 	\\ 
\(1487\) 	&	0.8837 	&	\ul{0.9435} 	&	0.9351 	&	\ul{0.9460} 	&	\ul{0.9468} 	&	\ul{0.9447} 	&	\ul{0.9466} 	&	\ul{0.9482} 	&	\B 0.9501 	&	\ul{0.9470} 	\\ 
\(1489^-\) 	&	0.5838 	&	0.8873 	&	0.7588 	&	\ul{0.9004} 	&	\ul{0.9002} 	&	\ul{0.8946} 	&	\ul{0.8986} 	&	\B 0.9028 	&	\ul{0.8990} 	&	0.8949 	\\ 
\(1491\) 	&	0.0100 	&	0.6177 	&	\B 0.8252 	&	0.8096 	&	0.8144 	&	0.7929 	&	0.8117 	&	0.8094 	&	\ul{0.8100} 	&	0.8010 	\\ 
\(1492^-\) 	&	0.0100 	&	0.5135 	&	0.1219 	&	0.5994 	&	\B 0.6146 	&	\ul{0.6137} 	&	0.5842 	&	\ul{0.6012} 	&	\ul{0.6094} 	&	0.5773 	\\ 
\(1493\) 	&	0.0104 	&	0.6412 	&	0.7217 	&	\ul{0.8135} 	&	\ul{0.8025} 	&	0.7858 	&	\B 0.8138 	&	\B 0.8138 	&	\ul{0.8037} 	&	\ul{0.8027} 	\\ 
\(1494\) 	&	0.5634 	&	0.8492 	&	0.7924 	&	\ul{0.8814} 	&	\B 0.8893 	&	\ul{0.8795} 	&	\ul{0.8823} 	&	\ul{0.8849} 	&	\ul{0.8760} 	&	\ul{0.8804} 	\\ 
\(1497\) 	&	0.3356 	&	0.9908 	&	0.5913 	&	\ul{0.9979} 	&	\ul{0.9971} 	&	0.9962 	&	\ul{0.9977} 	&	\B 0.9983 	&	\ul{0.9966} 	&	\ul{0.9975} 	\\ 
\(1501\) 	&	0.1008 	&	0.8690 	&	0.8559 	&	\ul{0.9475} 	&	\ul{0.9513} 	&	0.9433 	&	0.9406 	&	\B 0.9536 	&	0.9333 	&	0.9416 	\\ 
\(1504\) 	&	0.5528 	&	0.9758 	&	\B 1.0000 	&	\B 1.0000 	&	\B 1.0000 	&	\B 1.0000 	&	\B 1.0000 	&	\B 1.0000 	&	\B 1.0000 	&	\B 1.0000 	\\ 
\(1505\) 	&	0.0550 	&	0.9900 	&	0.1339 	&	\B 1.0000 	&	\B 1.0000 	&	\B 1.0000 	&	\B 1.0000 	&	\B 1.0000 	&	\B 1.0000 	&	\B 1.0000 	\\ 
\(1510\) 	&	0.5485 	&	0.9474 	&	0.8936 	&	\ul{0.9713} 	&	\ul{0.9713} 	&	\ul{0.9719} 	&	\B 0.9749 	&	\ul{0.9719} 	&	\ul{0.9731} 	&	\ul{0.9737} 	\\ 
\(1515\) 	&	0.0599 	&	0.7971 	&	\B 0.9029 	&	\ul{0.8959} 	&	\ul{0.8971} 	&	\ul{0.8884} 	&	0.8837 	&	0.8779 	&	0.8913 	&	0.8738 	\\ 
\(1570\) 	&	0.8988 	&	0.9814 	&	0.9450 	&	\ul{0.9857} 	&	\ul{0.9863} 	&	\ul{0.9853} 	&	\ul{0.9841} 	&	\B 0.9864 	&	\ul{0.9848} 	&	\ul{0.9851} 	\\ 
\(1596^-\) 	&	0.3771 	&	\B 0.9388 	&	0.6375 	&	0.8603 	&	0.9303 	&	\ul{0.9356} 	&	\ul{0.9344} 	&	\ul{0.8933} 	&	0.7836 	&	0.8638 	\\ 
\(4134^+\) 	&	0.5109 	&	0.7586 	&	0.6604 	&	\ul{0.7967} 	&	\ul{0.8017} 	&	\ul{0.7956} 	&	\ul{0.7937} 	&	\B 0.8058 	&	0.7942 	&	\ul{0.7969} 	\\ 
\(4134^+\) 	&	0.5023 	&	0.7674 	&	0.6660 	&	0.7950 	&	0.7955 	&	0.7856 	&	0.7948 	&	\B 0.8139 	&	0.7901 	&	0.8026 	\\ 
\(4135^-\) 	&	0.8914 	&	0.9441 	&	0.9413 	&	0.9480 	&	0.9477 	&	0.9458 	&	0.9473 	&	\B 0.9501 	&	\ul{0.9488} 	&	0.9475 	\\ 
\(4534^+\) 	&	0.5062 	&	0.9696 	&	0.9097 	&	\ul{0.9695} 	&	\ul{0.9701} 	&	\ul{0.9692} 	&	\ul{0.9712} 	&	\B 0.9724 	&	0.9658 	&	0.9694 	\\ 
\(4534^+\) 	&	0.5018 	&	0.9688 	&	0.9115 	&	\ul{0.9708} 	&	\ul{0.9698} 	&	\ul{0.9682} 	&	\ul{0.9711} 	&	\B 0.9726 	&	0.9646 	&	\ul{0.9699} 	\\ 
\(4538^-\) 	&	0.2374 	&	0.5936 	&	0.3597 	&	0.6505 	&	\B 0.6876 	&	0.6674 	&	0.6405 	&	\ul{0.6755} 	&	0.6349 	&	0.6469 	\\ 
\(23517^+\) 	&	0.4987 	&	0.5031 	&	0.5140 	&	\ul{0.5220} 	&	\ul{0.5225} 	&	\ul{0.5230} 	&	\ul{0.5221} 	&	\ul{0.5215} 	&	\ul{0.5236} 	&	\B 0.5236 	\\ 
\(40496\) 	&	0.0947 	&	0.7000 	&	\ul{0.7533} 	&	\ul{0.7653} 	&	\ul{0.7687} 	&	\ul{0.7627} 	&	\ul{0.7693} 	&	\ul{0.7653} 	&	\B 0.7713 	&	\ul{0.7573} 	\\ 
\(40499\) 	&	0.0888 	&	0.9622 	&	0.2067 	&	\ul{0.9981} 	&	0.9981 	&	0.9977 	&	\ul{0.9976} 	&	\B 0.9988 	&	\ul{0.9979} 	&	\ul{0.9981} 	\\ 
\(40509\) 	&	0.5145 	&	0.8667 	&	\ul{0.8831} 	&	\ul{0.8937} 	&	\ul{0.8932} 	&	\ul{0.8903} 	&	\ul{0.8932} 	&	\B 0.8947 	&	\ul{0.8903} 	&	\ul{0.8889} 	\\ 
\(40668^-\) 	&	0.5035 	&	0.7868 	&	0.6364 	&	0.8012 	&	0.8023 	&	0.7968 	&	0.7986 	&	\B 0.8084 	&	0.8027 	&	\ul{0.8019} 	\\ 
\(40670^-\) 	&	0.3855 	&	0.9182 	&	0.9449 	&	\ul{0.9635} 	&	\ul{0.9621} 	&	\ul{0.9616} 	&	\ul{0.9655} 	&	\B 0.9656 	&	0.9552 	&	\B 0.9656 	\\ 
\(40685^-\) 	&	0.6439 	&	\ul{0.9997} 	&	0.8191 	&	0.9995 	&	\ul{0.9997} 	&	0.9995 	&	0.9996 	&	\B 0.9998 	&	0.9996 	&	0.9994 	\\ 
\(40701^+\) 	&	0.7529 	&	0.9476 	&	0.8601 	&	\ul{0.9591} 	&	\ul{0.9603} 	&	0.9585 	&	\ul{0.9592} 	&	\B 0.9618 	&	0.9531 	&	\ul{0.9561} 	\\ 
\(40923^-\) 	&	0.0213 	&	\ul{0.7779} 	&	0.5717 	&	\ul{0.7187} 	&	\ul{0.7562} 	&	\ul{0.7308} 	&	0.6277 	&	\B 0.7879 	&	0.6694 	&	0.6610 	\\ 
\(40927^+\) 	&	0.0994 	&	0.3510 	&	0.2956 	&	\ul{0.3726} 	&	0.3680 	&	\B 0.3974 	&	0.3285 	&	0.3744 	&	0.3282 	&	0.3142 	\\ 
\(40975^+\) 	&	0.5395 	&	0.9563 	&	0.7597 	&	\ul{0.9881} 	&	\ul{0.9911} 	&	0.9723 	&	\ul{0.9956} 	&	\B 0.9963 	&	0.9873 	&	\ul{0.9913} 	\\ 
\(40978^+\) 	&	0.7520 	&	0.9735 	&	0.9685 	&	\ul{0.9780} 	&	\ul{0.9778} 	&	\ul{0.9754} 	&	\ul{0.9771} 	&	\B 0.9792 	&	0.9738 	&	\ul{0.9744} 	\\ 
\(40979^+\) 	&	0.0962 	&	0.9522 	&	0.9185 	&	\ul{0.9822} 	&	\ul{0.9825} 	&	\ul{0.9810} 	&	\ul{0.9823} 	&	\B 0.9865 	&	0.9777 	&	0.9785 	\\ 
\(40981^+\) 	&	0.5150 	&	0.8459 	&	0.8657 	&	\ul{0.8865} 	&	\ul{0.8845} 	&	\ul{0.8792} 	&	\ul{0.8816} 	&	\B 0.8942 	&	\B 0.8942 	&	\ul{0.8845} 	\\ 
\(40982^+\) 	&	0.2310 	&	0.7448 	&	0.4407 	&	\ul{0.7861} 	&	\ul{0.8005} 	&	\ul{0.7913} 	&	\ul{0.7962} 	&	\B 0.8014 	&	0.7772 	&	\ul{0.7878} 	\\ 
\(40983^+\) 	&	0.8981 	&	0.9791 	&	0.9451 	&	\ul{0.9851} 	&	\ul{0.9864} 	&	\ul{0.9860} 	&	\ul{0.9853} 	&	\B 0.9874 	&	0.9842 	&	\ul{0.9857} 	\\ 
\(40984^-\) 	&	0.1423 	&	0.9222 	&	0.4307 	&	\ul{0.9335} 	&	0.9325 	&	0.9261 	&	\ul{0.9349} 	&	\B 0.9408 	&	\ul{0.9355} 	&	\ul{0.9394} 	\\ 
\(40994^+\) 	&	0.8469 	&	0.9191 	&	0.9185 	&	\ul{0.9673} 	&	\B 0.9710 	&	0.9611 	&	\ul{0.9630} 	&	\ul{0.9648} 	&	\ul{0.9617} 	&	0.9586 	\\ 
\(40996^-\) 	&	0.1014 	&	0.8571 	&	0.7158 	&	0.8526 	&	\ul{0.8610} 	&	0.8543 	&	0.8570 	&	\B 0.8656 	&	0.8520 	&	0.8487 	\\ 
\(41027^-\) 	&	0.4247 	&	0.7878 	&	0.6166 	&	\ul{0.8697} 	&	0.8610 	&	0.8550 	&	\ul{0.8698} 	&	\B 0.8759 	&	0.8473 	&	\ul{0.8605} 	\\ 
\(41142^-\) 	&	0.4954 	&	0.6806 	&	0.6603 	&	\ul{0.7299} 	&	\ul{0.7294} 	&	0.7256 	&	\ul{0.7294} 	&	\B 0.7363 	&	\ul{0.7346} 	&	\ul{0.7315} 	\\ 
\(41143^+\) 	&	0.5030 	&	0.7769 	&	0.7510 	&	\ul{0.8248} 	&	\B 0.8253 	&	0.8192 	&	\ul{0.8229} 	&	\ul{0.8247} 	&	0.8160 	&	0.8184 	\\ 
\(41146^-\) 	&	0.5004 	&	0.9300 	&	0.5080 	&	\ul{0.9516} 	&	\ul{0.9501} 	&	0.9464 	&	\ul{0.9518} 	&	\B 0.9527 	&	0.9441 	&	0.9445 	\\ 
\(41150^-\) 	&	0.5962 	&	0.9238 	&	0.7733 	&	0.9316 	&	0.9300 	&	0.9293 	&	0.9288 	&	\B 0.9332 	&	0.9285 	&	0.9303 	\\ 
\(41159^-\) 	&	0.5211 	&	0.7765 	&	0.5849 	&	0.7237 	&	\ul{0.7617} 	&	0.7443 	&	0.7329 	&	\B 0.7973 	&	0.7118 	&	0.7585 	\\ 
\(41161^+\) 	&	0.6243 	&	0.9351 	&	0.7037 	&	\ul{0.9863} 	&	\ul{0.9868} 	&	\ul{0.9863} 	&	0.9855 	&	\B 0.9884 	&	\ul{0.9868} 	&	\ul{0.9868} 	\\ 
\(41163^-\) 	&	0.2001 	&	0.9171 	&	0.6670 	&	0.9384 	&	\ul{0.9473} 	&	0.9270 	&	0.9295 	&	\B 0.9485 	&	0.9401 	&	0.9406 	\\ 
\(41164^-\) 	&	0.1620 	&	0.6657 	&	0.6544 	&	0.6864 	&	\B 0.6951 	&	0.6892 	&	0.6896 	&	\ul{0.6924} 	&	\ul{0.6909} 	&	\ul{0.6935} 	\\ 
\(41165^-\) 	&	0.0989 	&	0.3104 	&	0.3271 	&	\ul{0.3897} 	&	0.3654 	&	0.3745 	&	\B 0.4055 	&	\ul{0.4055} 	&	0.3956 	&	\ul{0.3940} 	\\ 
\(41166^-\) 	&	0.1481 	&	0.6116 	&	0.3813 	&	0.6439 	&	0.6451 	&	0.6328 	&	0.6306 	&	\B 0.6508 	&	0.6321 	&	0.6349 	\\ 
\(41167^+\) 	&	0.0029 	&	\B 0.8720 	&	0.4201 	&	0.7447 	&	0.8553 	&	0.8399 	&	0.8603 	&	\ul{0.8543} 	&	0.7388 	&	0.8089 	\\ 
\(41168^-\) 	&	0.3593 	&	0.6588 	&	0.5277 	&	\ul{0.6887} 	&	\ul{0.6890} 	&	0.6850 	&	\ul{0.6880} 	&	\B 0.6913 	&	0.6848 	&	\ul{0.6886} 	\\ 
\(41169^-\) 	&	0.0225 	&	0.2917 	&	0.1725 	&	\ul{0.3242} 	&	\B 0.3330 	&	0.3248 	&	0.3202 	&	\ul{0.3320} 	&	0.3235 	&	0.3222 	\\ 
\\
Average & 0.3902 	&	0.8335 	&	0.6964 	&	0.8746 	&	0.8782 	&	0.8725 	&	0.8748 	&	\B0.8821 	&	0.8711 	&	0.8732 	\\

	\bottomrule
	
	\caption{
	Average accuracy of \ac{CASH} solvers on selected \name{OpenML} data sets. Data sets containing missing values are omitted. The best results per data set are highlighted in bold. Results not significantly worse than the best result---according to a Wilcoxon signed-rank test---are underlined. On data sets marked by \(^+\) and \(^-\), \ac{CASH} solvers performed better and worse, respectively, than \ac{AutoML} frameworks.
}
\label{tbl:results_evaluation_cash}
\end{longtable}

\end{footnotesize}

\begin{figure}
	\centering
	
	\begin{subfigure}[b]{0.24\textwidth}
		\includegraphics[width=\textwidth]{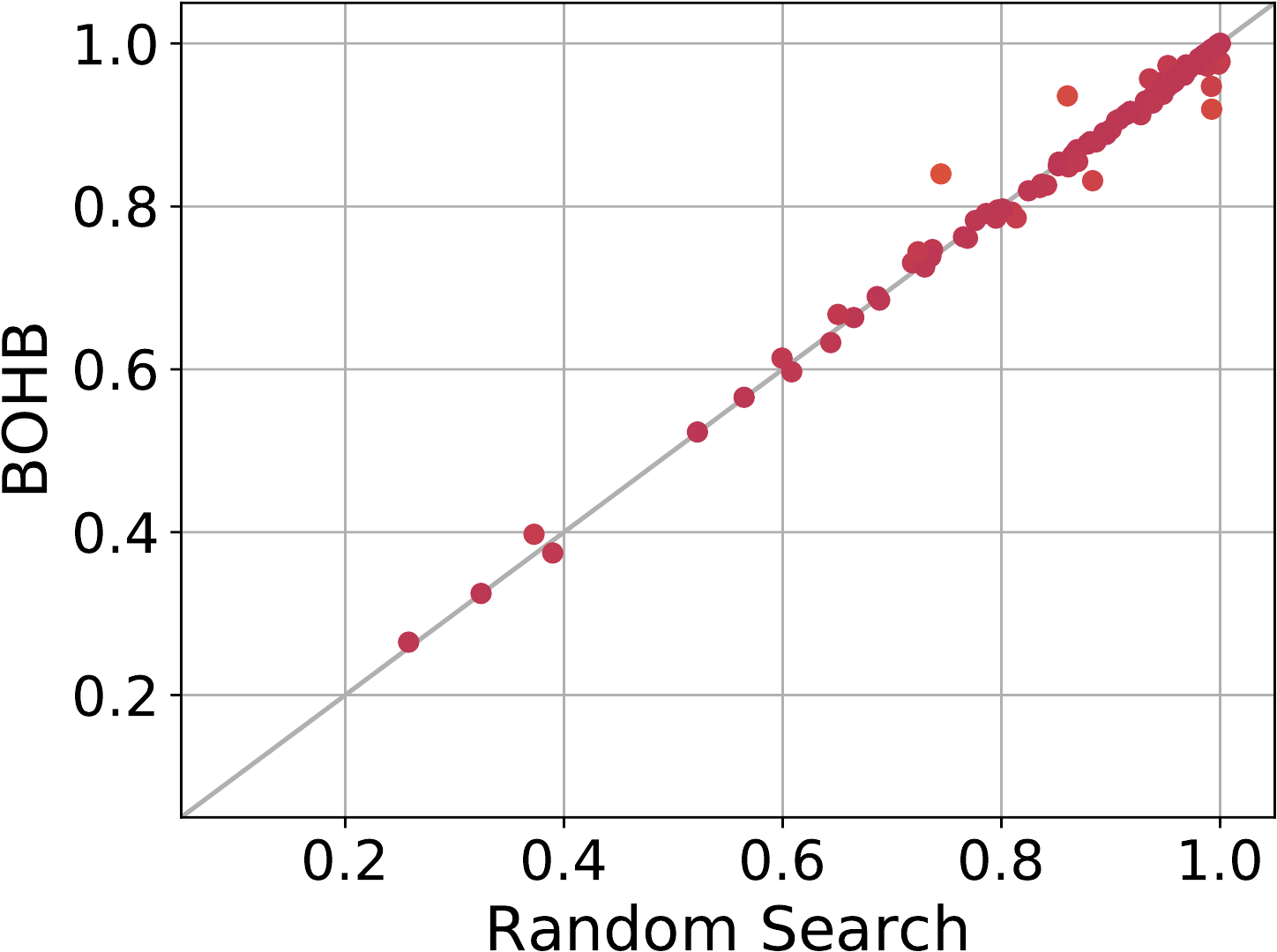}
	\end{subfigure}
	\begin{subfigure}[b]{0.24\textwidth}
		\includegraphics[width=\textwidth]{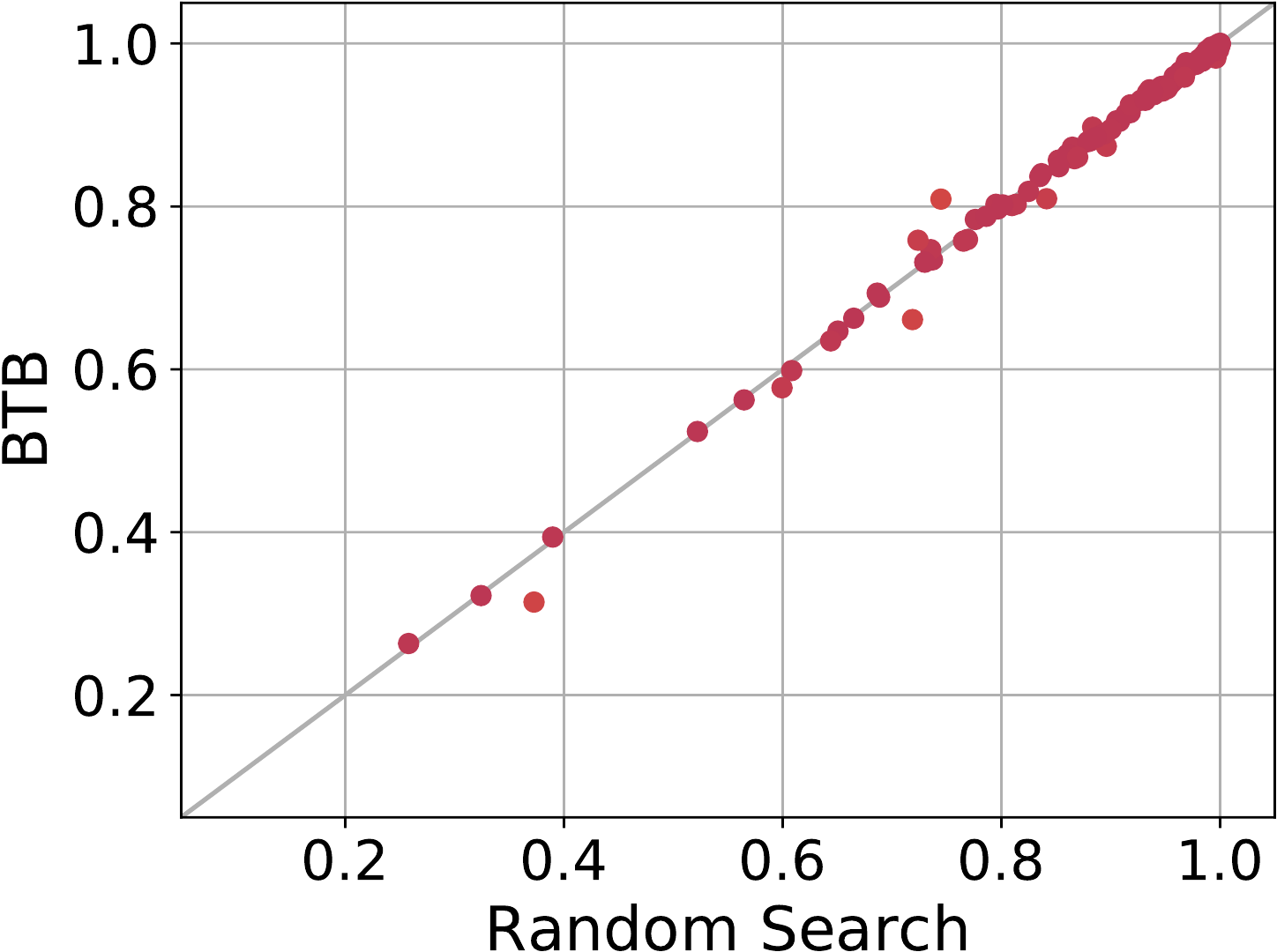}
	\end{subfigure}
	\begin{subfigure}[b]{0.24\textwidth}
		\includegraphics[width=\textwidth]{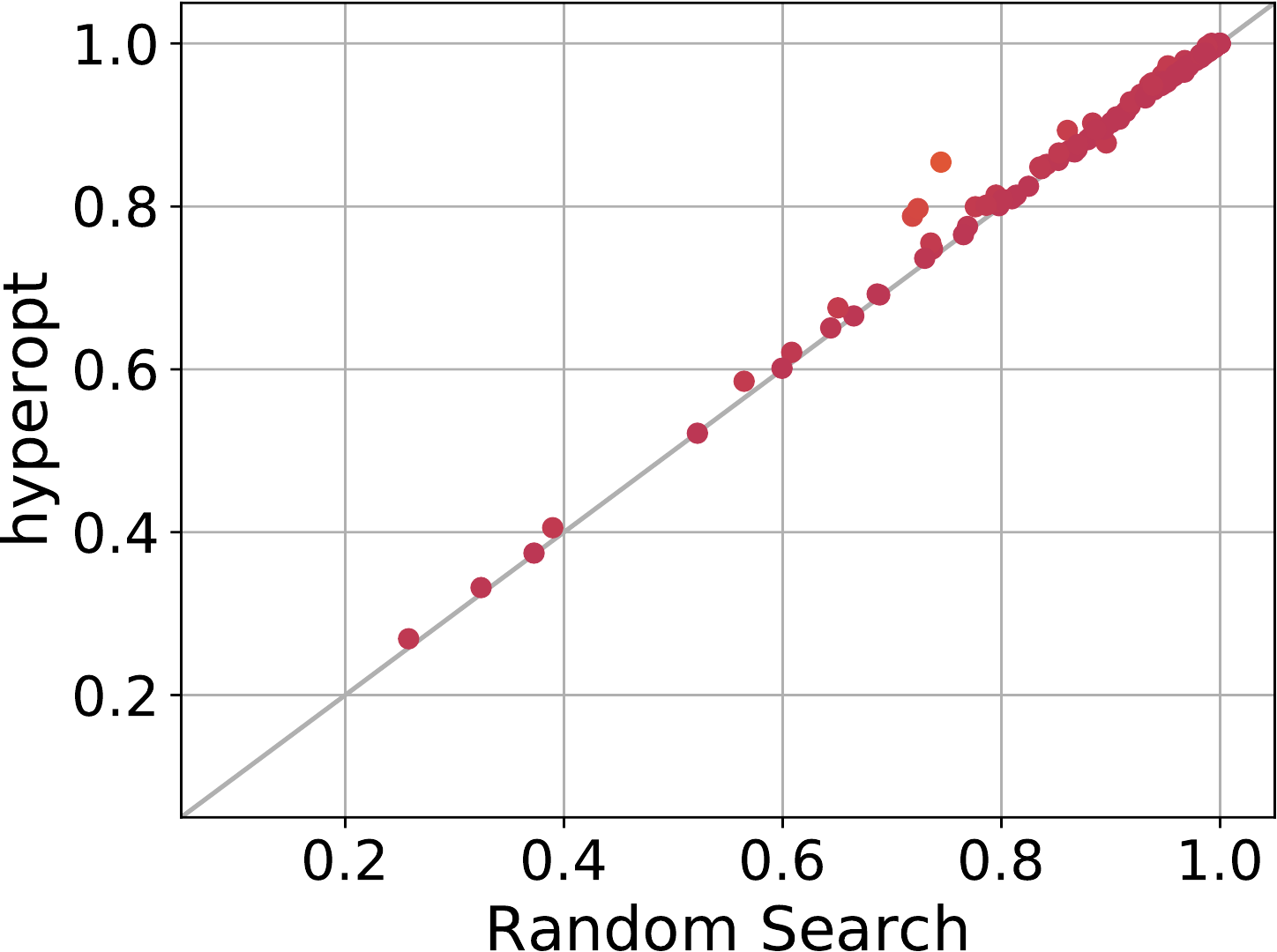}
	\end{subfigure}
	\begin{subfigure}[b]{0.24\textwidth}
		\includegraphics[width=\textwidth]{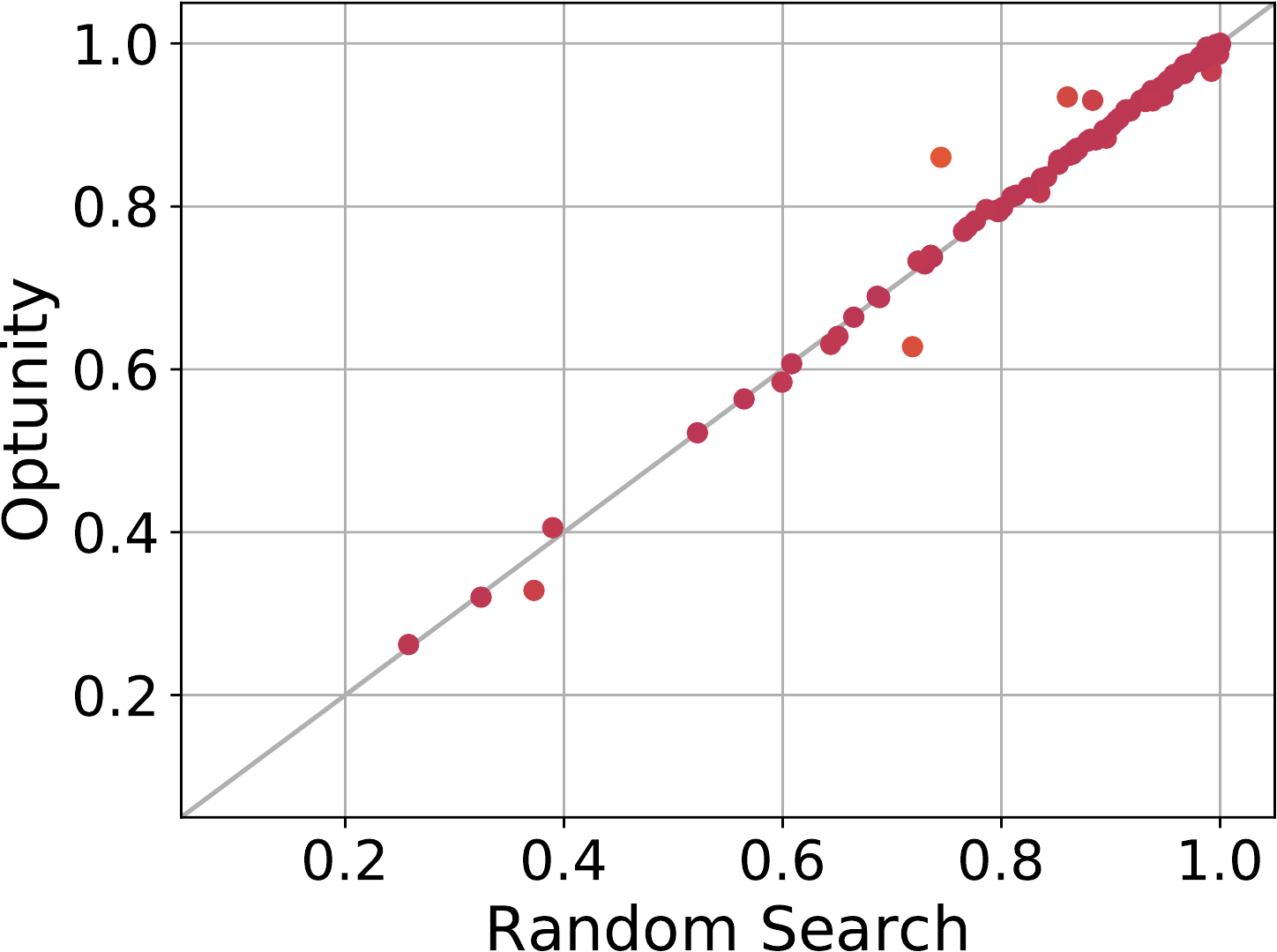}
	\end{subfigure}
	
	\begin{subfigure}[b]{0.24\textwidth}
		\includegraphics[width=\textwidth]{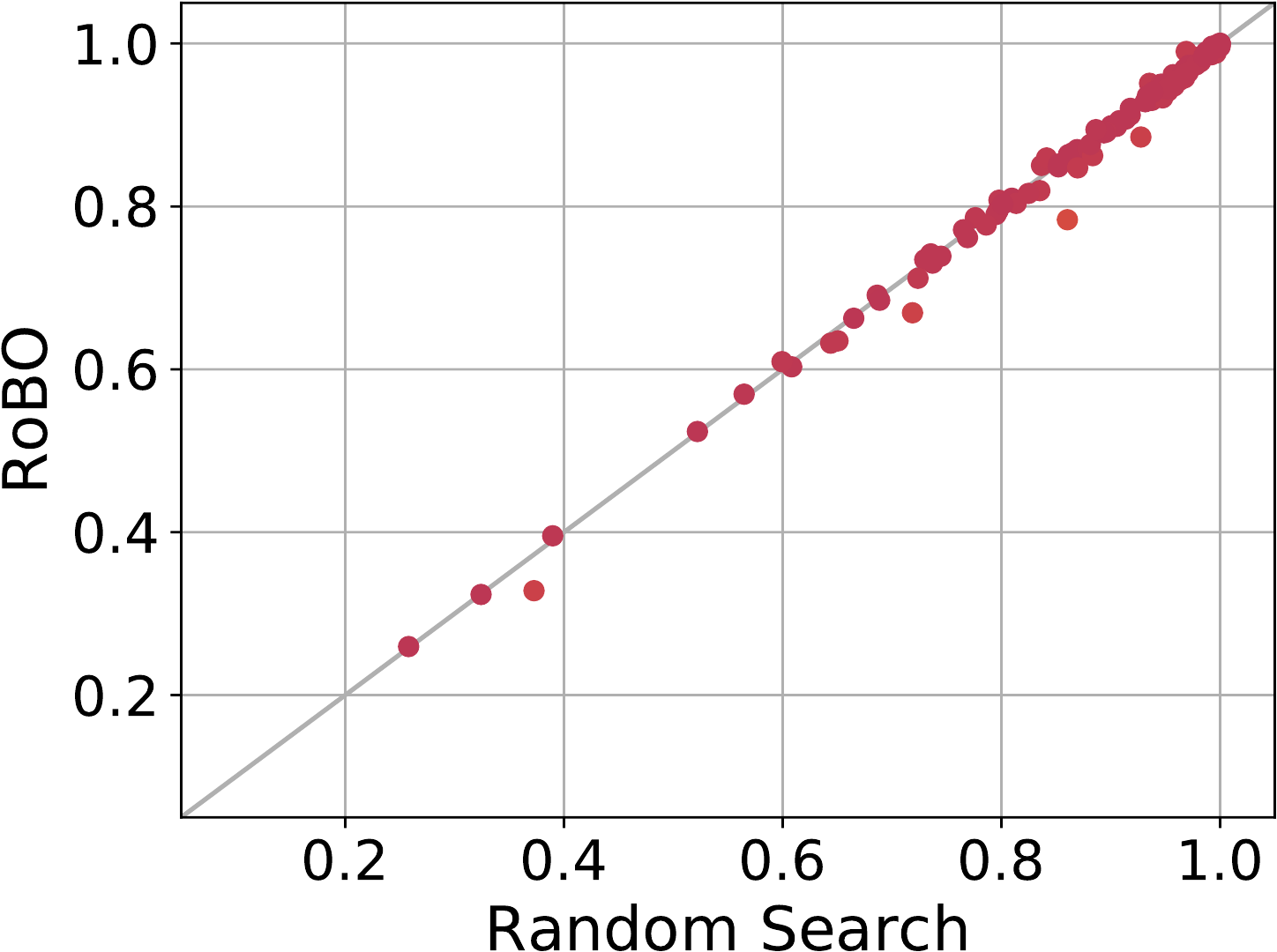}
	\end{subfigure}
	\begin{subfigure}[b]{0.24\textwidth}
		\includegraphics[width=\textwidth]{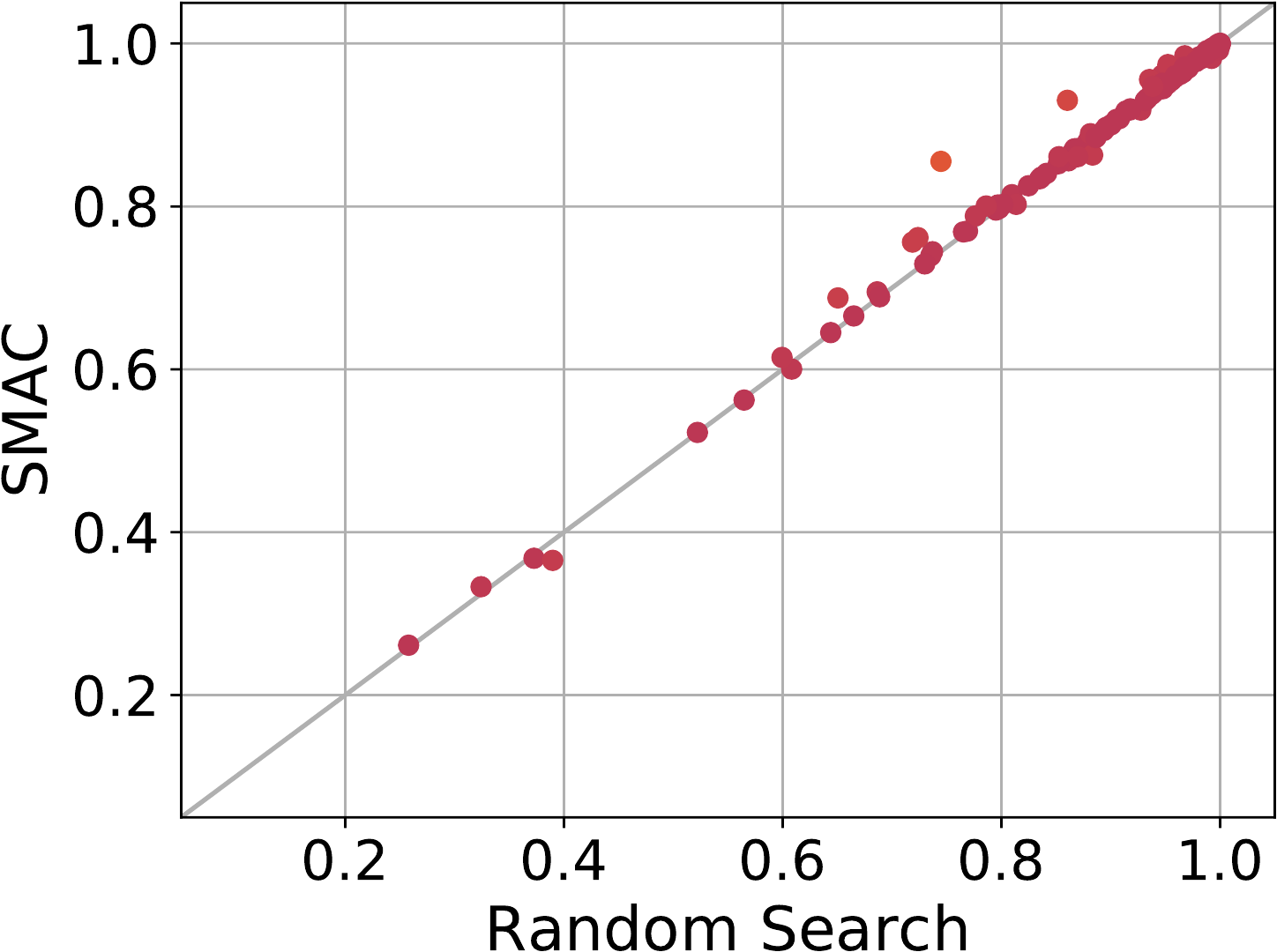}
	\end{subfigure}
	\begin{subfigure}[b]{0.24\textwidth}
		\includegraphics[width=\textwidth]{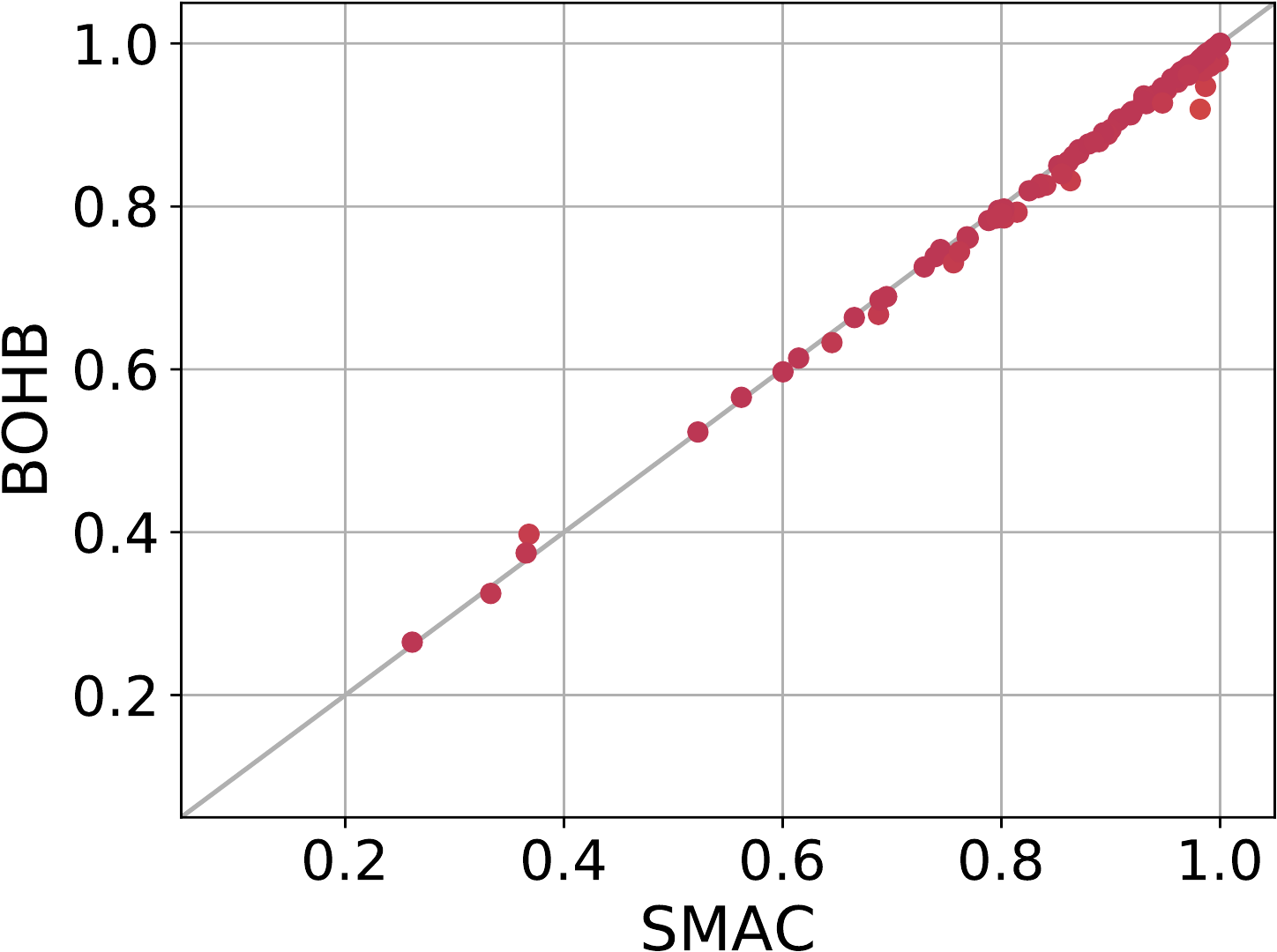}
	\end{subfigure}
	\begin{subfigure}[b]{0.24\textwidth}
		\includegraphics[width=\textwidth]{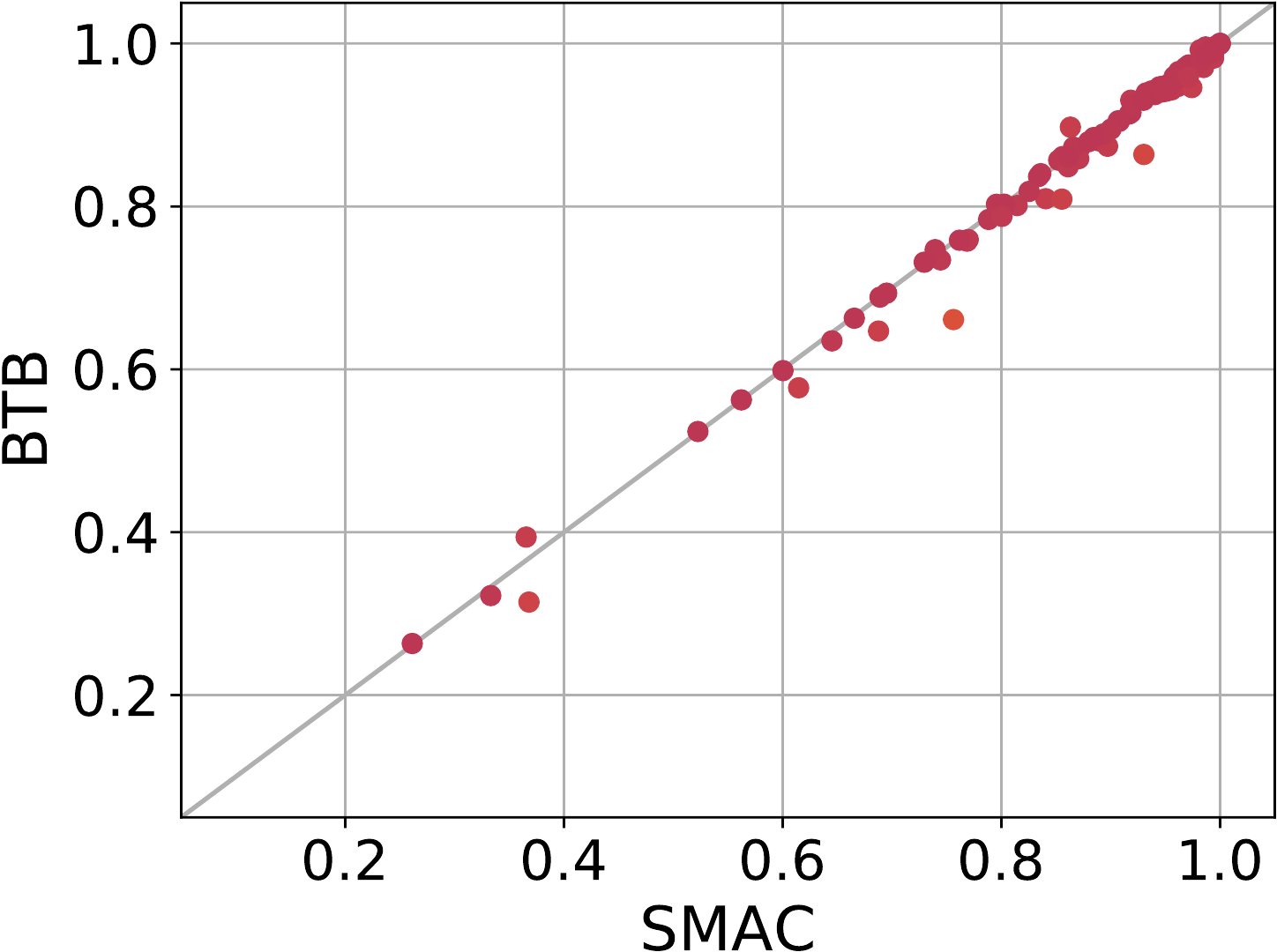}
	\end{subfigure}
	
	\begin{subfigure}[b]{0.24\textwidth}
		\includegraphics[width=\textwidth]{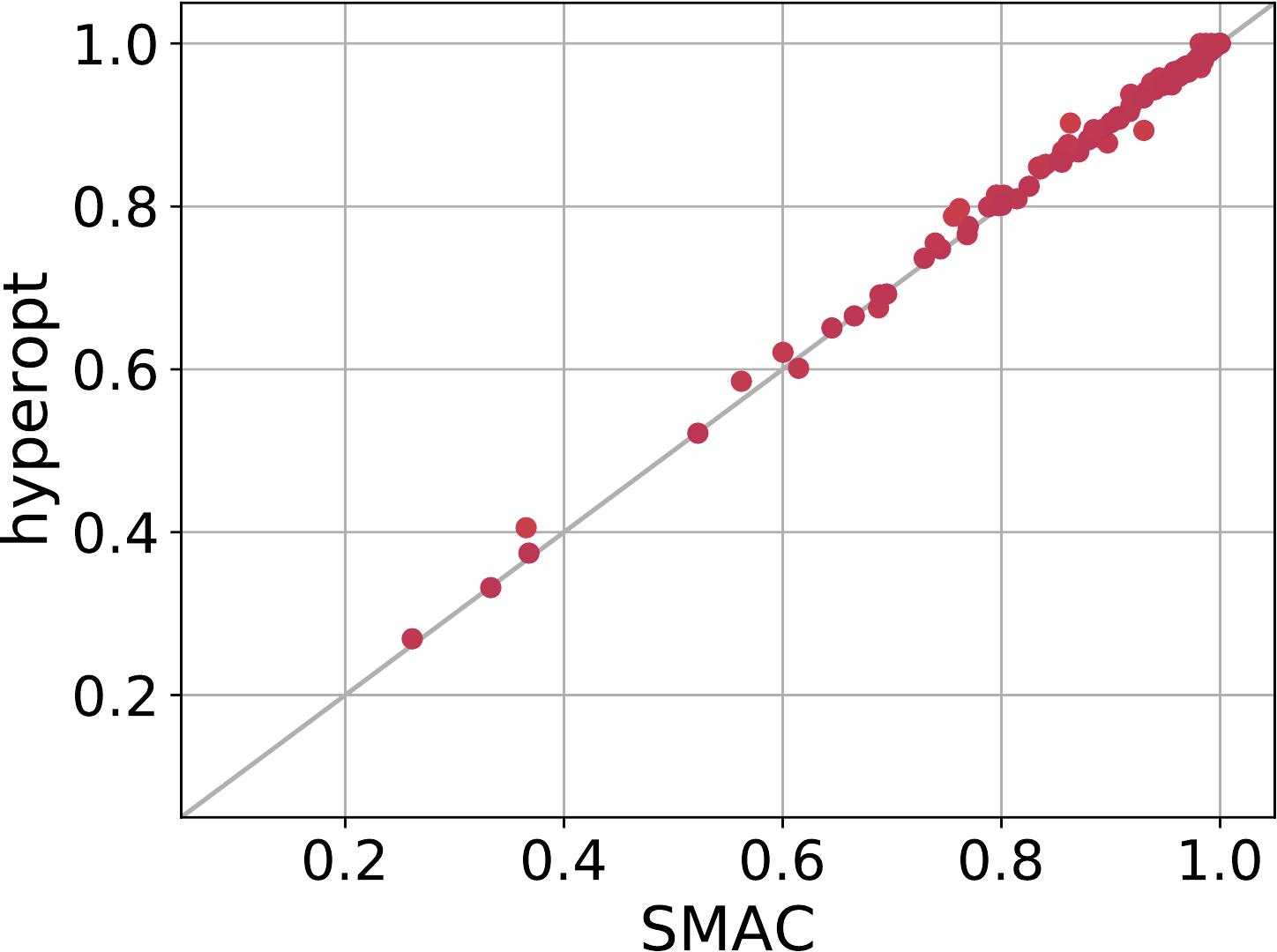}
	\end{subfigure}
	\begin{subfigure}[b]{0.24\textwidth}
		\includegraphics[width=\textwidth]{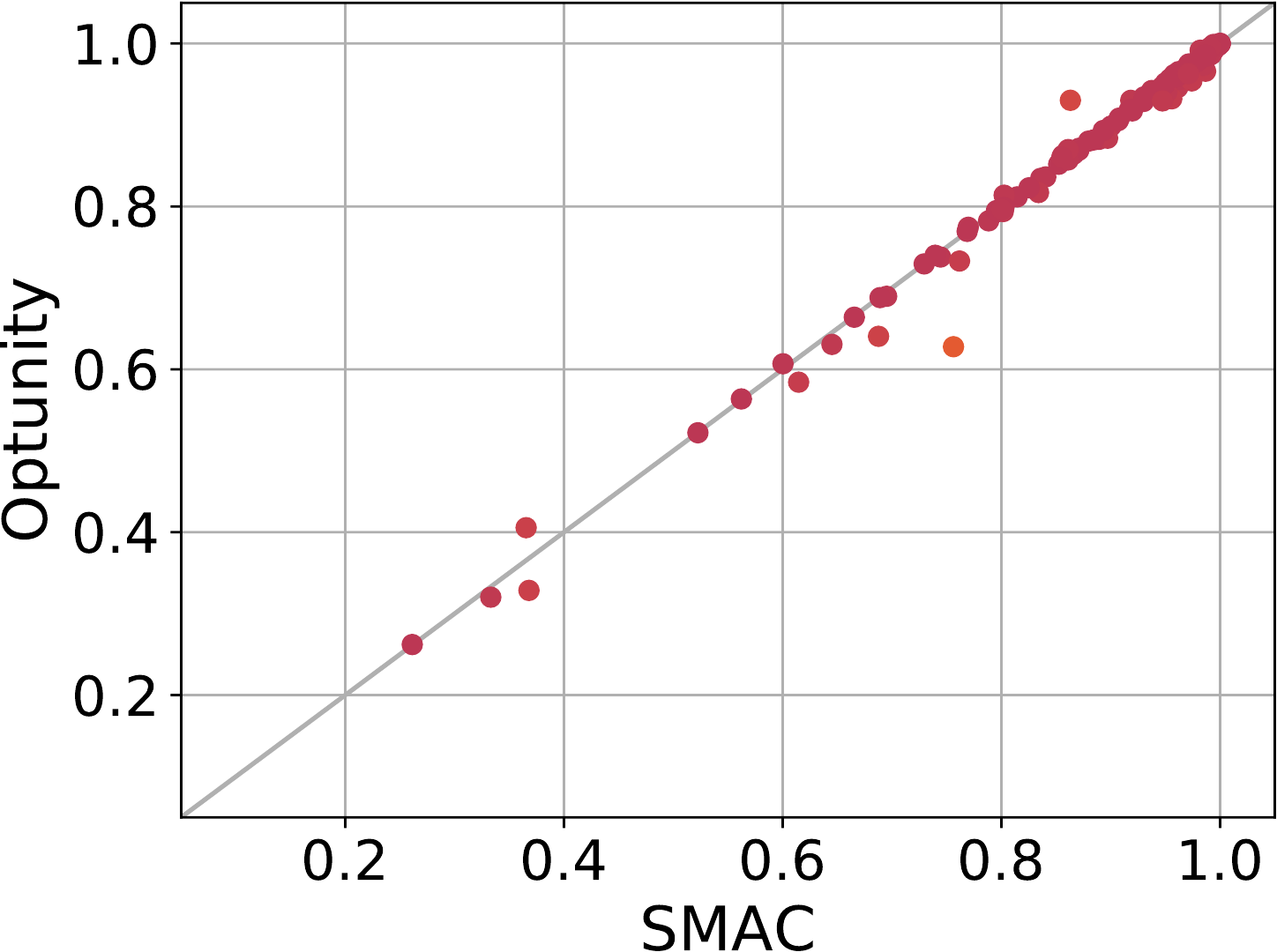}
	\end{subfigure}
	\begin{subfigure}[b]{0.24\textwidth}
		\includegraphics[width=\textwidth]{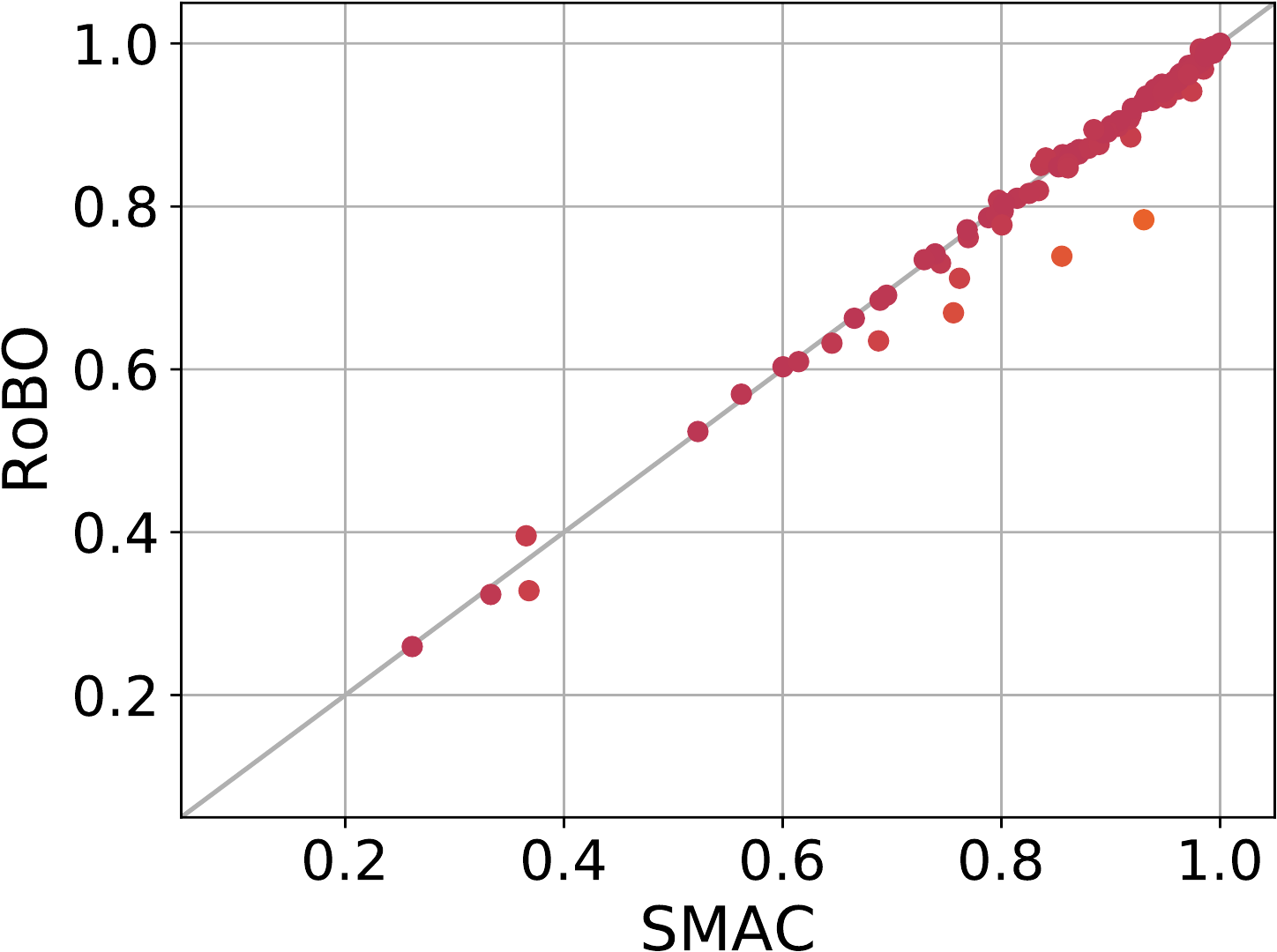}
	\end{subfigure}
	\begin{subfigure}[b]{0.24\textwidth}
		\includegraphics[width=\textwidth]{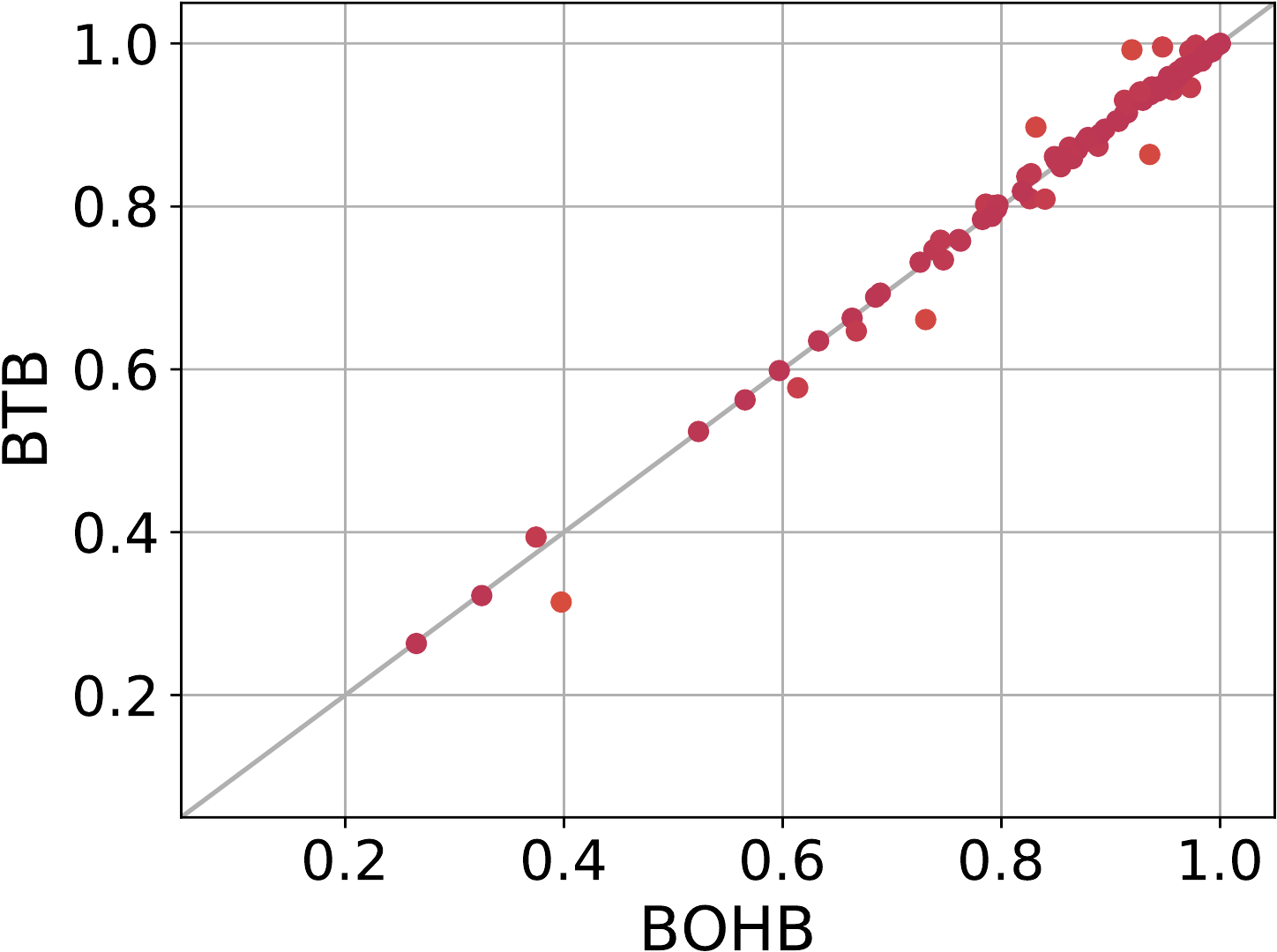}
	\end{subfigure}
	
	\begin{subfigure}[b]{0.24\textwidth}
		\includegraphics[width=\textwidth]{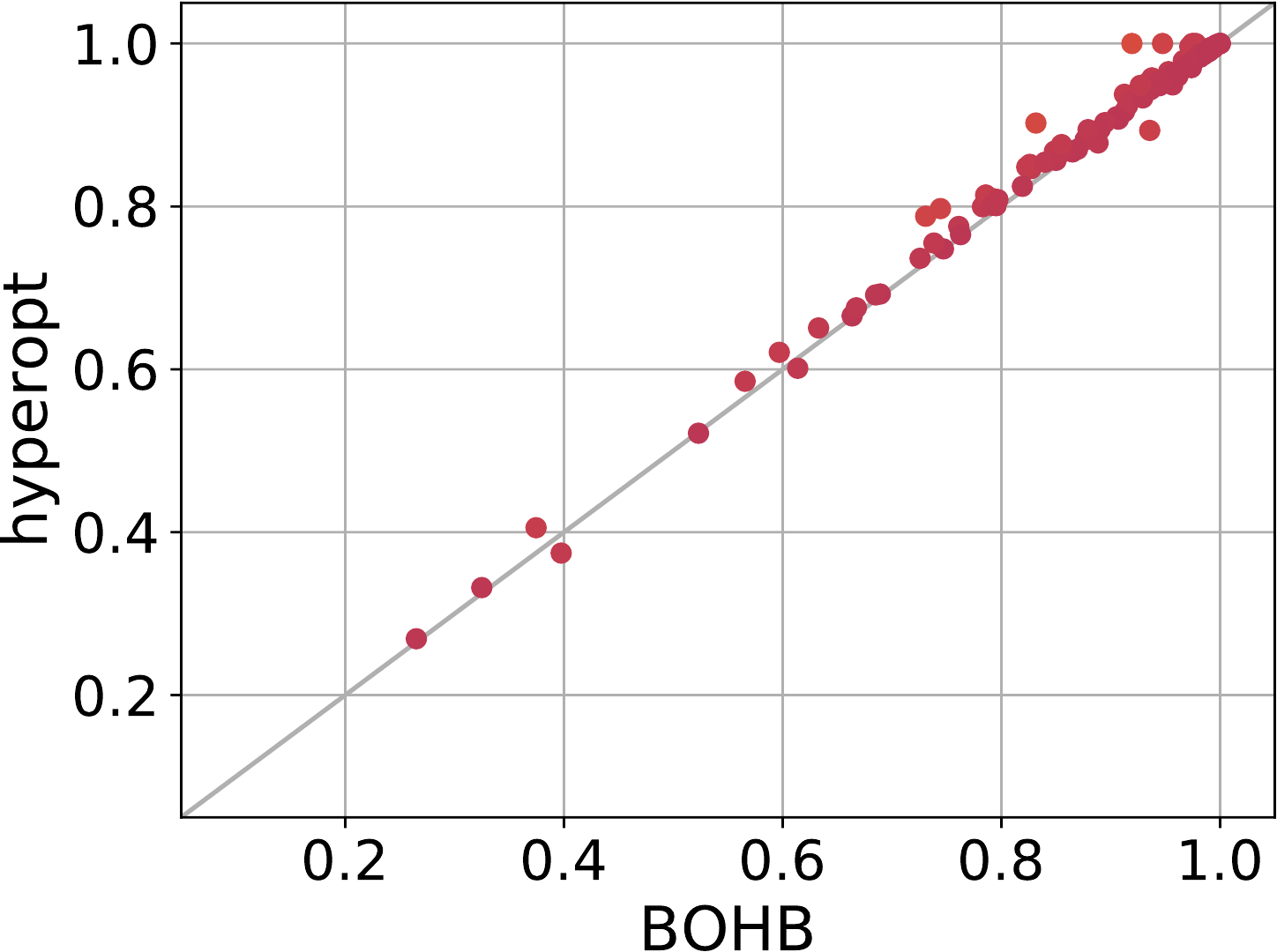}
	\end{subfigure}
	\begin{subfigure}[b]{0.24\textwidth}
		\includegraphics[width=\textwidth]{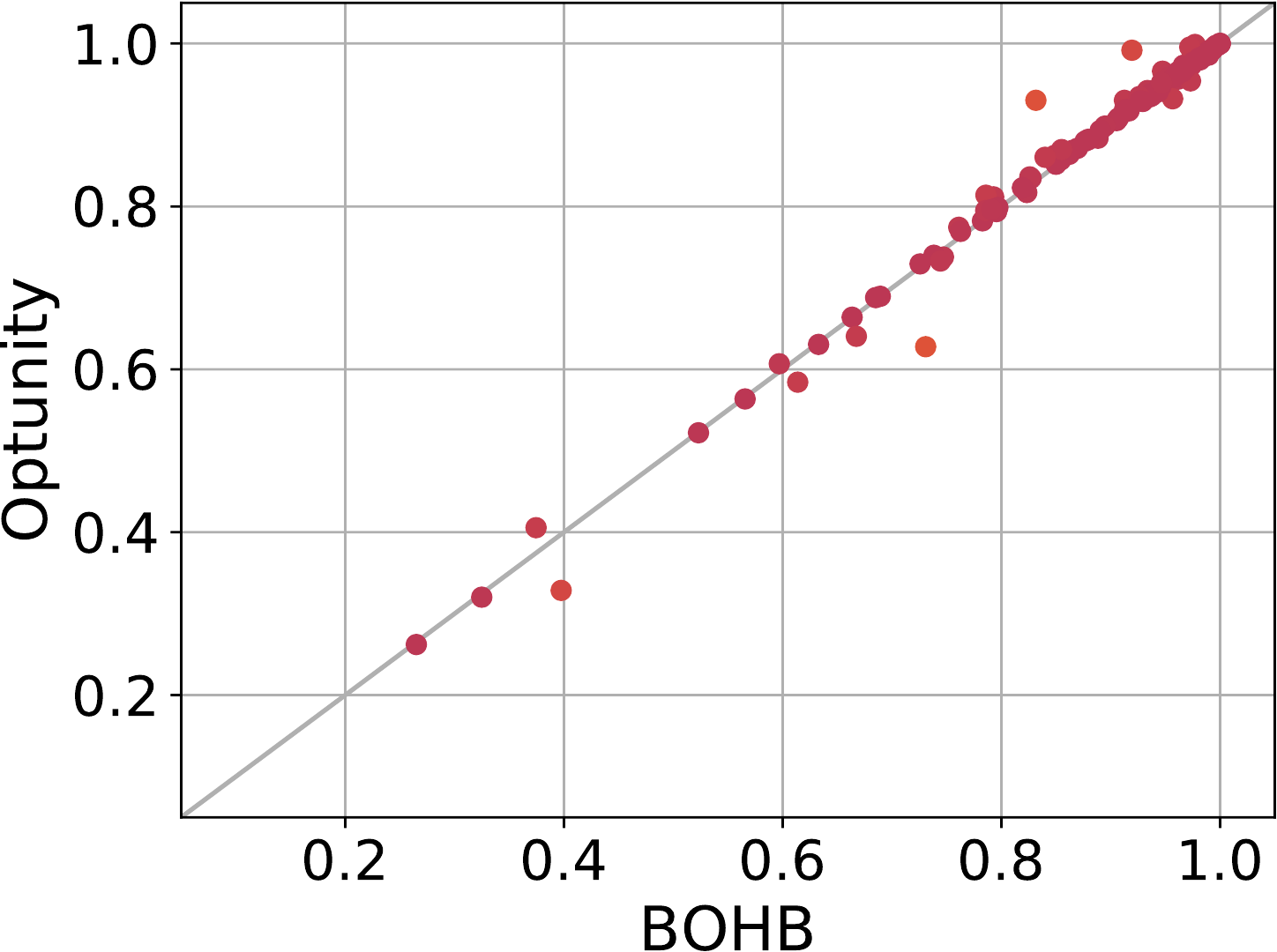}
	\end{subfigure}
	\begin{subfigure}[b]{0.24\textwidth}
		\includegraphics[width=\textwidth]{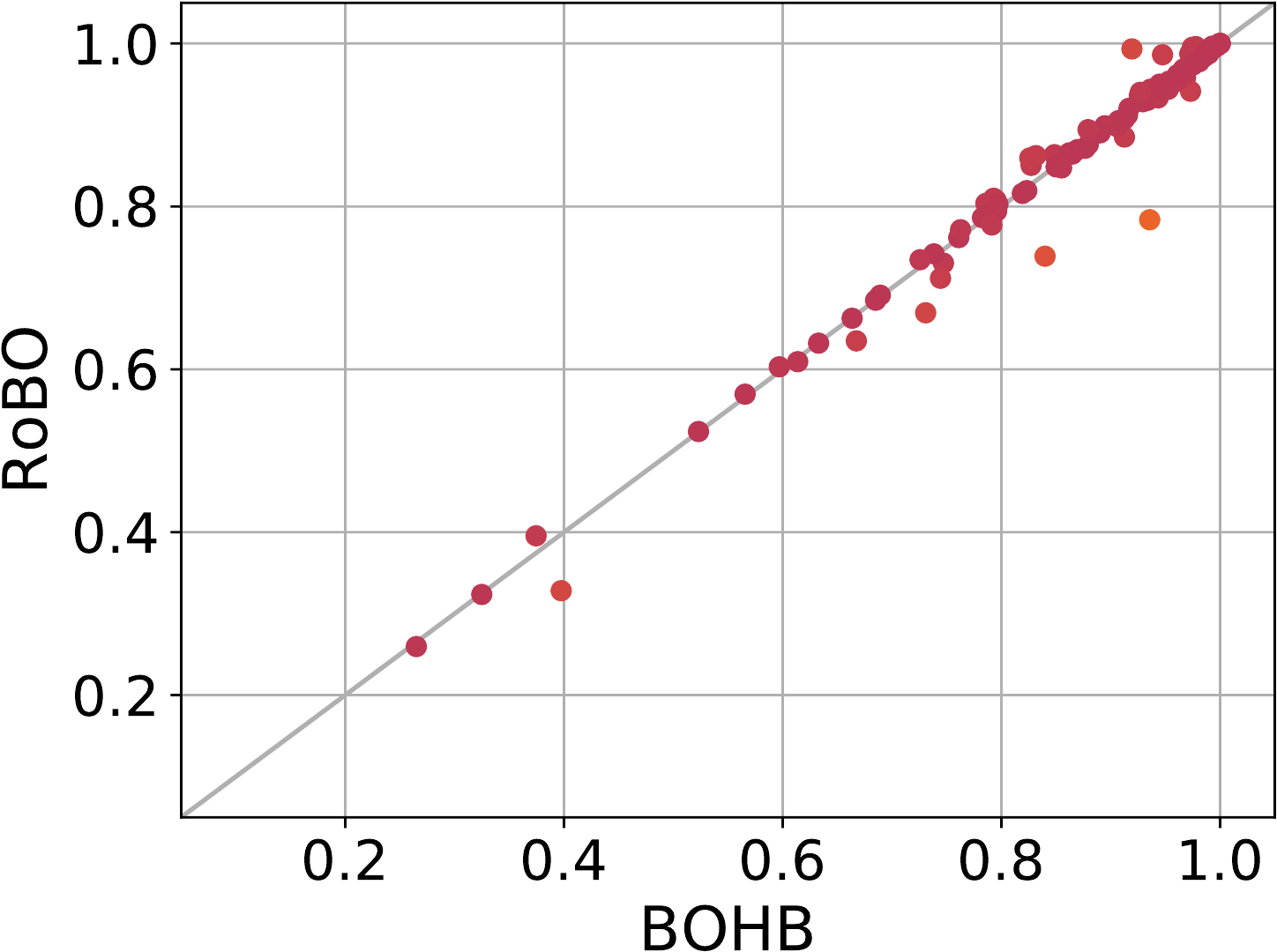}
	\end{subfigure}
	\begin{subfigure}[b]{0.24\textwidth}
		\includegraphics[width=\textwidth]{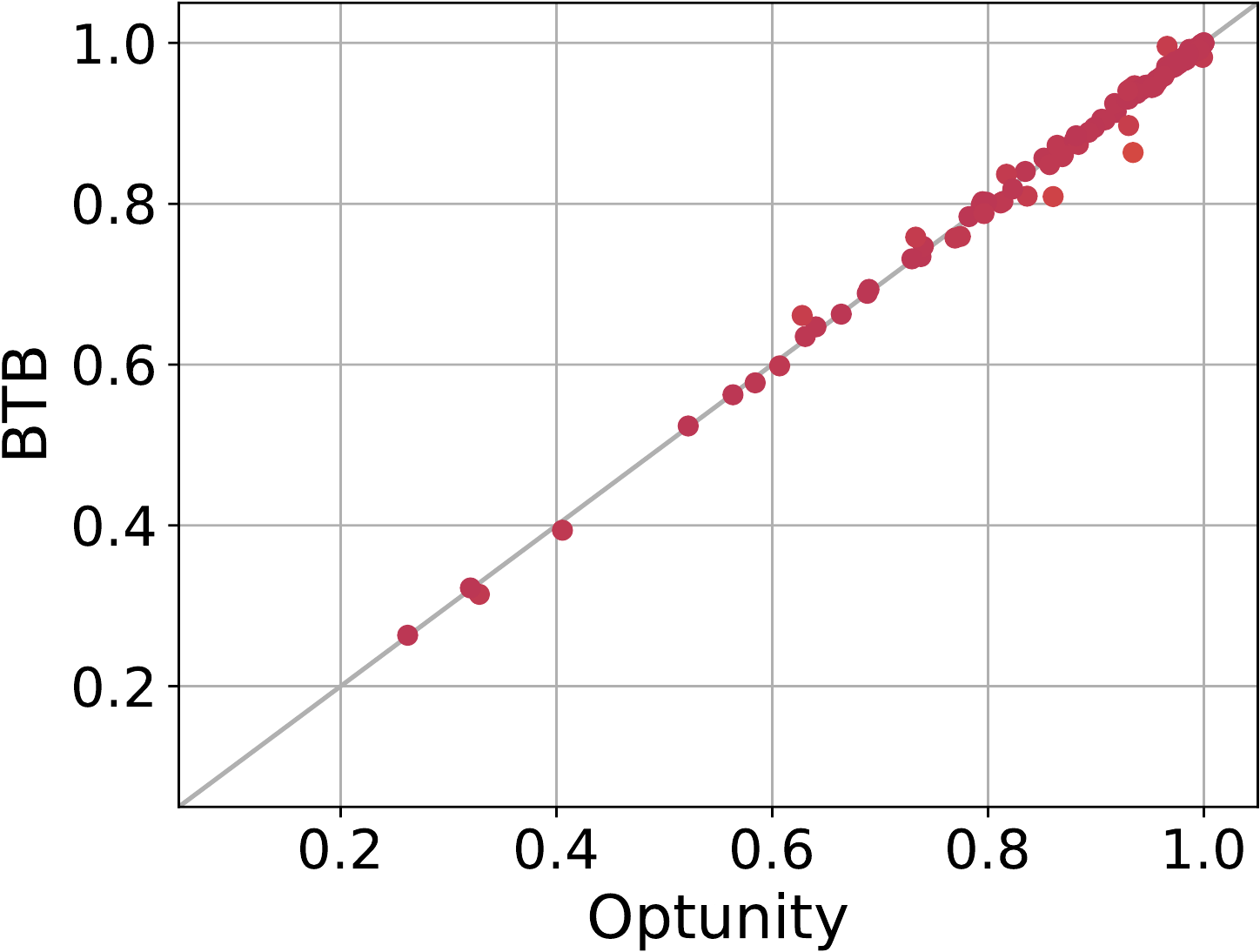}
	\end{subfigure}
	
	\begin{subfigure}[b]{0.24\textwidth}
		\includegraphics[width=\textwidth]{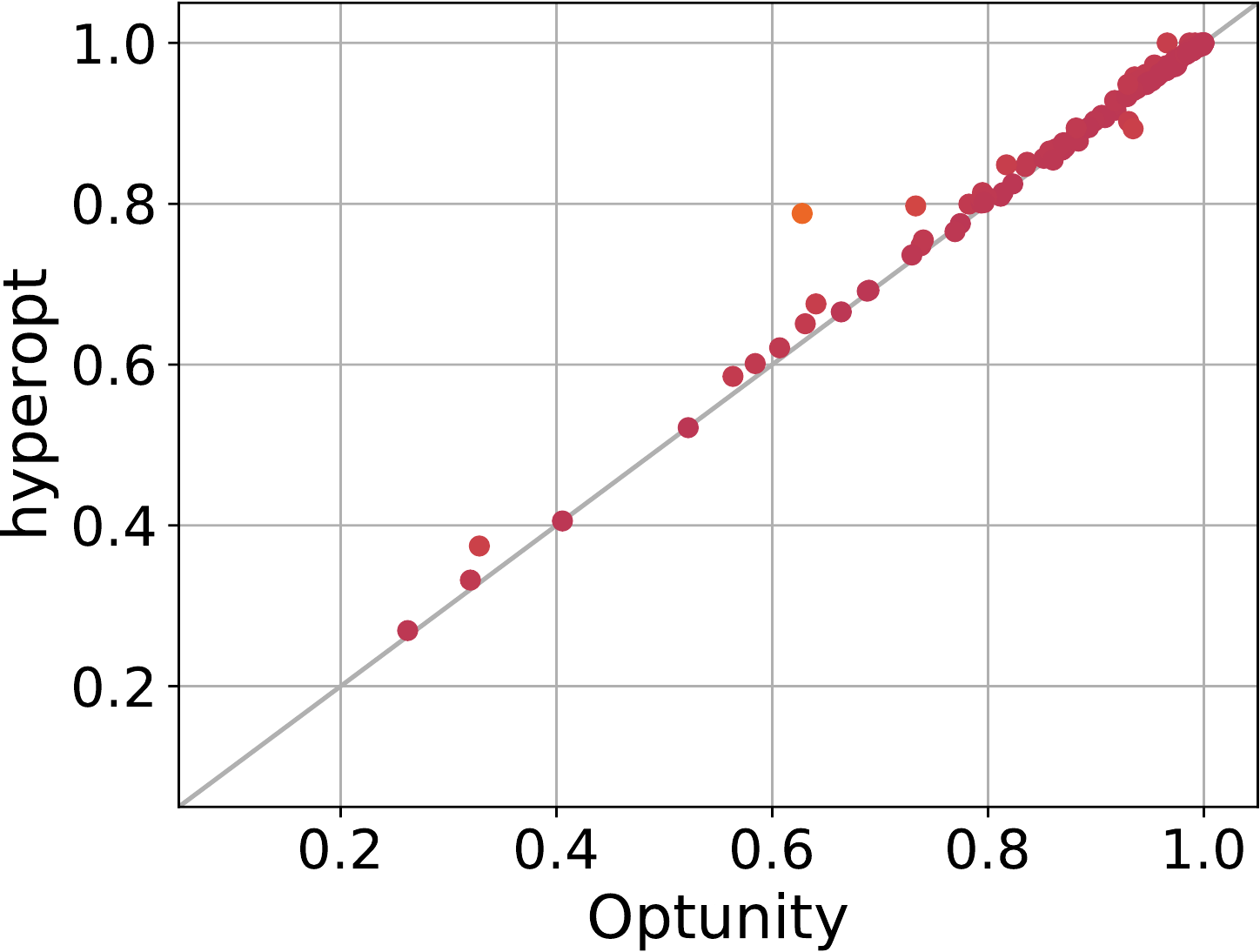}
	\end{subfigure}
	\begin{subfigure}[b]{0.24\textwidth}
		\includegraphics[width=\textwidth]{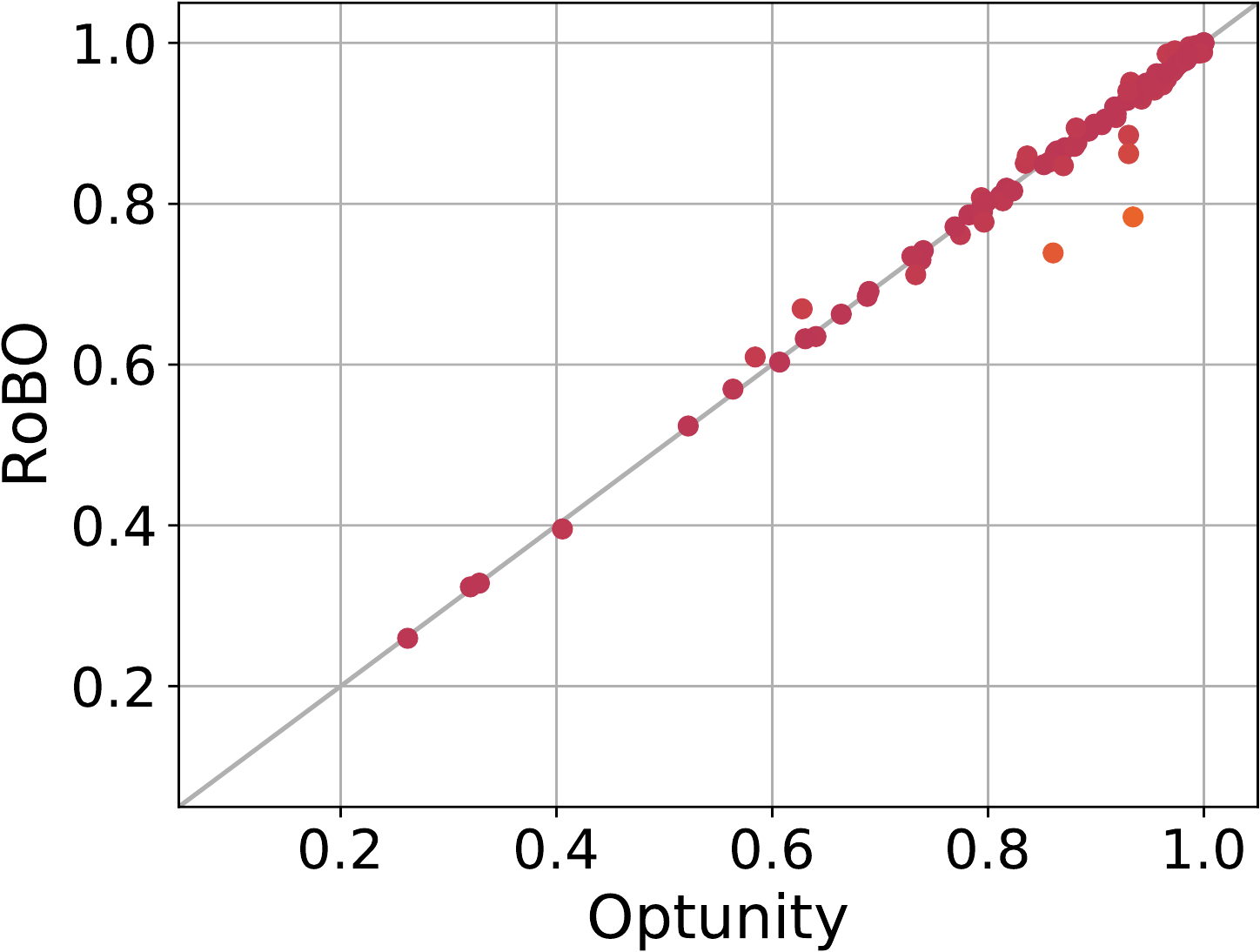}
	\end{subfigure}
	\begin{subfigure}[b]{0.24\textwidth}
		\includegraphics[width=\textwidth]{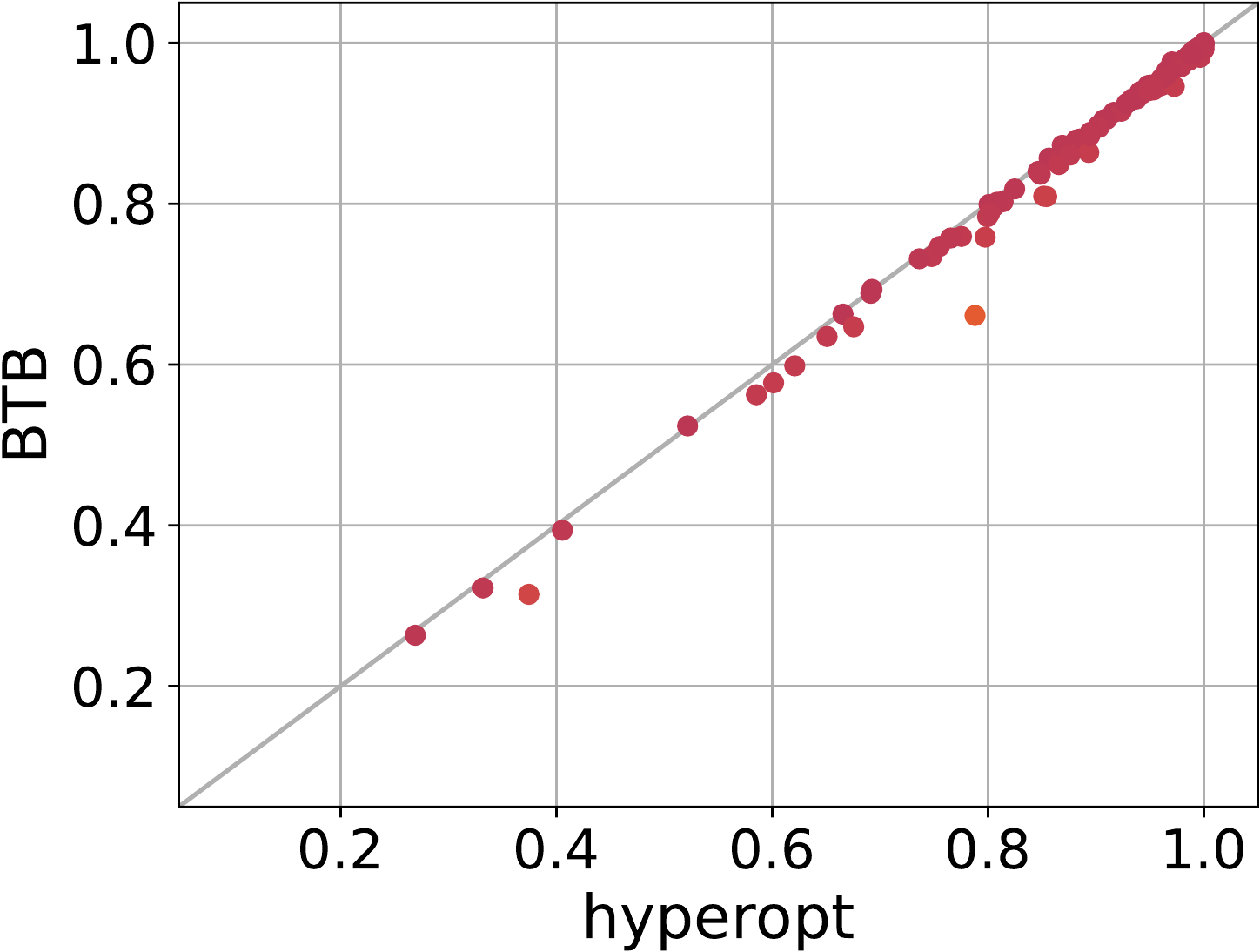}
	\end{subfigure}
	\begin{subfigure}[b]{0.24\textwidth}
		\includegraphics[width=\textwidth]{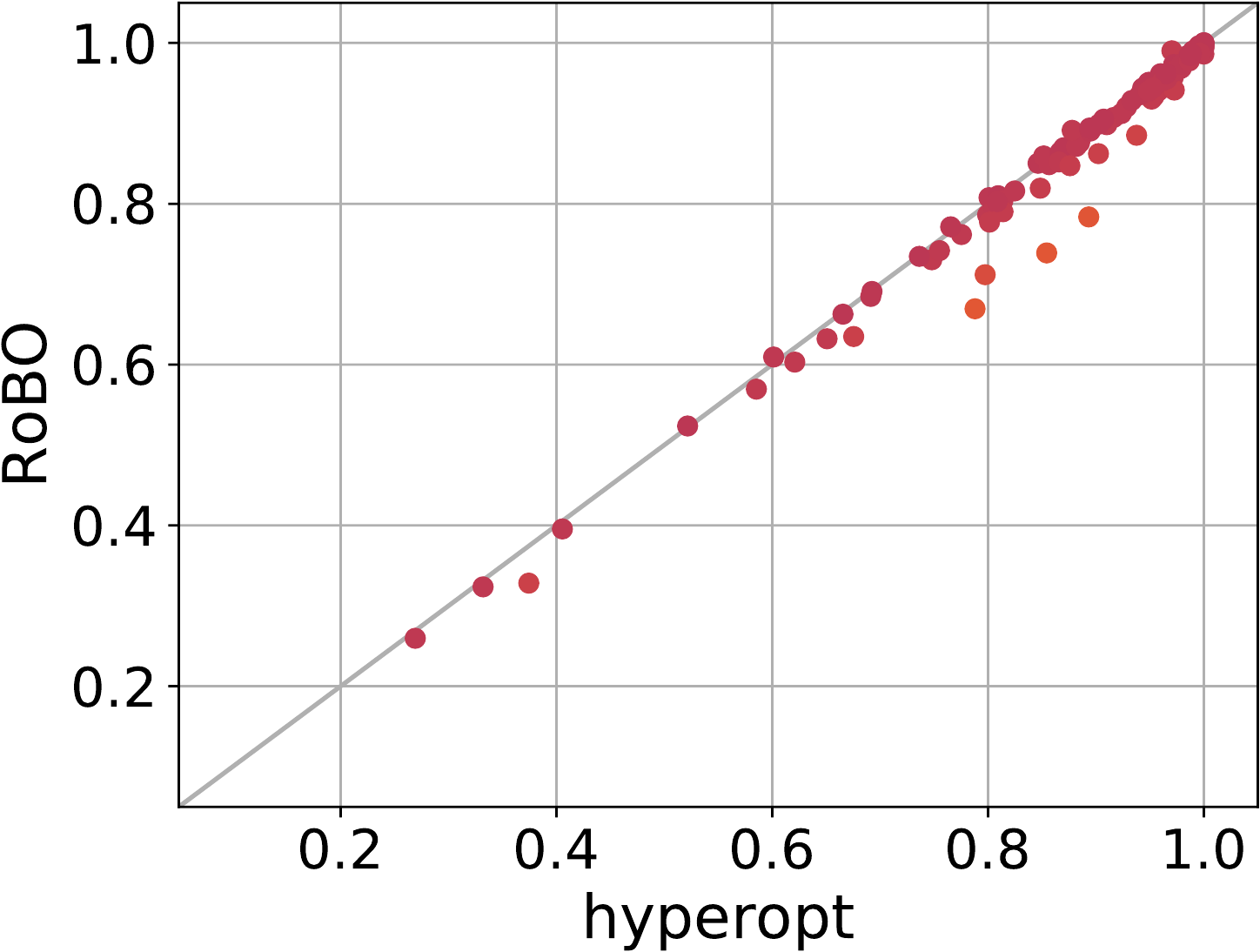}
	\end{subfigure}
	
	\begin{subfigure}[b]{0.24\textwidth}
		\includegraphics[width=\textwidth]{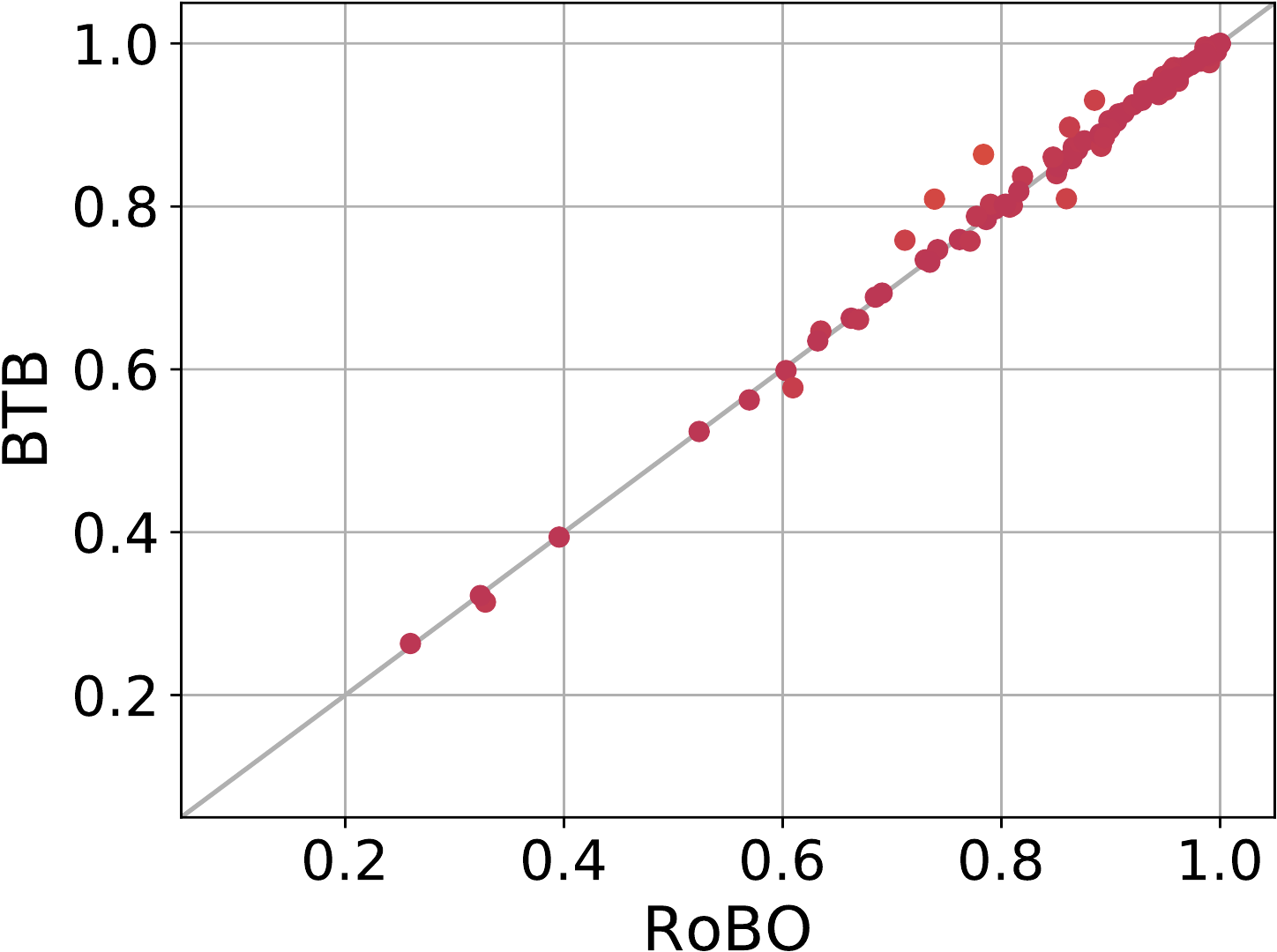}
	\end{subfigure}
	
	\caption{Pair-wise comparison of the mean precision of \ac{CASH} algorithms. The axes represent the accuracy score of the stated \ac{CASH} algorithm. Each point represents the averaged results for a single data set. Identical performances are plotted directly on the angle bisector. The comparison with grid search is omitted due to spacial constrictions.}
	\label{fig:
	}
\end{figure}

\begin{footnotesize}
\centering

\newrobustcmd{\B}{\fontseries{b}\selectfont}

\renewcommand{\arraystretch}{0.9}

\begin{longtable}{@{} l l l l l l l l l @{}}
	\toprule
	Data Set	& Dummy	& RF & Random	& auto-sklearn	& TPOT	& ATM	& hpsklearn	& H2O	\\
	\midrule

\(3^-\) 	&	0.50761 	&	0.98467 	&	0.99062 	&	0.98986 	&	\B 0.99431 	&	\ul{0.99326} 	&	\ul{0.99051} 	&	\ul{0.99426} 	\\
\(12^-\) 	&	0.10317 	&	0.94617 	&	\ul{0.97633} 	&	\ul{0.97767} 	&	0.97333 	&	\B 0.98178 	&	0.94758 	&	0.97433 	\\
\(15\) 	&	0.52857 	&	0.95714 	&	0.95873 	&	0.96875 	&	0.96571 	&	\B 0.98474 	&	0.96000 	&	0.96286 	\\
\(23^-\) 	&	0.35249 	&	0.50950 	&	0.53262 	&	0.54638 	&	0.55882 	&	\B 0.58100 	&	0.53047 	&	0.53733 	\\
\(24\) 	&	0.49922 	&	\B 1.00000 	&	\ul{0.99993} 	&	\B 1.00000 	&	\B 1.00000 	&	\B 1.00000 	&	\B 1.00000 	&	\ul{0.99848} 	\\
\(29\) 	&	0.51111 	&	0.84976 	&	0.85507 	&	\ul{0.87289} 	&	0.86377 	&	\B 0.89133 	&	0.85956 	&	0.86184 	\\
\(31^-\) 	&	0.56867 	&	0.72667 	&	0.72400 	&	0.73433 	&	0.74400 	&	\B 0.76578 	&	0.70121 	&	0.74867 	\\
\(38\) 	&	0.88207 	&	\ul{0.98454} 	&	\ul{0.98550} 	&	\ul{0.98288} 	&	\B 0.98746 	&	--		&	0.97438 	&	\ul{0.98419} 	\\
\(42\) 	&	0.08439 	&	0.91561 	&	0.91911 	&	0.91954 	&	0.92732 	&	\B 0.94504 	&	0.92585 	&	0.93122 	\\
\(54^-\) 	&	0.26417 	&	0.72165 	&	\ul{0.81969} 	&	\ul{0.82008} 	&	\ul{0.81811} 	&	\ul{0.81522} 	&	0.75787 	&	\B 0.82717 	\\
\(188\) 	&	0.21267 	&	\ul{0.61086} 	&	\ul{0.62670} 	&	\ul{0.63886} 	&	\ul{0.65566} 	&	\ul{0.64190} 	&	\ul{0.64072} 	&	\B 0.65570 	\\
\(451\) 	&	0.50533 	&	\ul{0.99933} 	&	0.99081 	&	0.99019 	&	0.99091 	&	\B 1.00000 	&	\ul{0.99404} 	&	\ul{0.97967} 	\\
\(469^+\) 	&	0.16583 	&	0.18625 	&	0.20382 	&	0.20365 	&	0.20833 	&	\B 0.27028 	&	0.19139 	&	0.19542 	\\
\(470\) 	&	0.56733 	&	0.65050 	&	0.64563 	&	0.65687 	&	0.66832 	&	\B 0.71221 	&	0.63762 	&	\ul{0.71089} 	\\
\(1053\) 	&	0.68766 	&	0.80505 	&	0.81126 	&	0.81344 	&	\ul{0.81810} 	&	\B 0.82100 	&	0.80998 	&	0.74819 	\\
\(1067^-\) 	&	0.74060 	&	0.84739 	&	0.85340 	&	0.85118 	&	\ul{0.86019} 	&	\B 0.86856 	&	0.84044 	&	0.80869 	\\
\(1111\) 	&	0.96487 	&	\ul{0.98235} 	&	\ul{0.98228} 	&	\B 0.98244 	&	\ul{0.98182} 	&	--		&	\ul{0.98189} 	&	0.96555 	\\
\(1112\) 	&	0.86358 	&	0.92542 	&	\ul{0.92586} 	&	\B 0.92725 	&	\ul{0.92624} 	&	--		&	0.92599 	&	0.78802 	\\
\(1114\) 	&	0.86357 	&	0.94048 	&	\ul{0.95030} 	&	\B 0.95094 	&	\ul{0.95085} 	&	--		&	\ul{0.95068} 	&	0.93415 	\\
\(1169^+\) 	&	0.50570 	&	0.61520 	&	0.59845 	&	0.66665 	&	\B 0.66895 	&	0.63671 	&	0.65080 	&	0.61266 	\\
\(1461^-\) 	&	0.79323 	&	0.89985 	&	0.90398 	&	0.90447 	&	\B 0.90705 	&	0.89957 	&	\ul{0.90451} 	&	0.90060 	\\
\(1464^+\) 	&	0.63200 	&	0.74889 	&	0.77778 	&	0.76667 	&	0.78711 	&	\B 0.81956 	&	0.78044 	&	0.73378 	\\
\(1468^-\) 	&	0.10741 	&	0.88765 	&	0.93117 	&	0.94167 	&	\ul{0.94784} 	&	\B 0.96049 	&	0.94012 	&	\ul{0.95216} 	\\
\(1475^-\) 	&	0.24553 	&	0.58998 	&	0.58601 	&	0.59695 	&	\ul{0.61291} 	&	0.60272 	&	0.58293 	&	\B 0.61656 	\\
\(1486^+\) 	&	0.59173 	&	0.96344 	&	0.96656 	&	0.96903 	&	\ul{0.97026} 	&	0.96055 	&	\ul{0.96891} 	&	\B 0.97146 	\\
\(1489^+\) 	&	0.58453 	&	0.88890 	&	0.89205 	&	\ul{0.89716} 	&	\B 0.90450 	&	\ul{0.89963} 	&	\ul{0.89273} 	&	\ul{0.89205} 	\\
\(1492^+\) 	&	0.00687 	&	0.51333 	&	\ul{0.62795} 	&	\B 0.65172 	&	\ul{0.61146} 	&	\ul{0.61097} 	&	\ul{0.54667} 	&	\ul{0.56435} 	\\
\(1590\) 	&	0.63379 	&	0.85021 	&	\ul{0.87013} 	&	\ul{0.86938} 	&	\B 0.87089 	&	0.85448 	&	\ul{0.86727} 	&	0.86656 	\\
\(1596^+\) 	&	0.37644 	&	0.93818 	&	0.89143 	&	\B 0.96395 	&	\ul{0.94542} 	&	0.66390 	&	0.95227 	&	0.92908 	\\
\(4134^-\) 	&	0.50462 	&	0.76314 	&	0.77762 	&	\ul{0.78890} 	&	\B 0.80249 	&	0.77087 	&	0.77798 	&	\ul{0.80044} 	\\
\(4135^+\) 	&	0.88895 	&	0.94491 	&	0.94444 	&	0.94761 	&	0.94891 	&	0.94606 	&	0.94750 	&	\B 0.95114 	\\
\(4534^-\) 	&	0.50612 	&	\ul{0.96847} 	&	0.96244 	&	0.96590 	&	\ul{0.96913} 	&	0.96464 	&	\ul{0.96964} 	&	\B 0.97160 	\\
\(4538^+\) 	&	0.23130 	&	0.59207 	&	0.65004 	&	0.67733 	&	0.67586 	&	0.66217 	&	0.67272 	&	\B 0.70165 	\\
\(4550\) 	&	0.12346 	&	0.99414 	&	\ul{0.99907} 	&	\B 1.00000 	&	\B 1.00000 	&	\B 1.00000 	&	\ul{0.99983} 	&	\B 1.00000 	\\
\(6332\) 	&	0.52407 	&	0.73951 	&	0.76173 	&	0.79012 	&	\ul{0.81009} 	&	\B 0.81701 	&	0.76667 	&	\ul{0.78333} 	\\
\(6332\) 	&	0.49877 	&	0.76481 	&	0.77058 	&	0.77353 	&	\B 0.81173 	&	\ul{0.79155} 	&	0.75823 	&	\ul{0.80000} 	\\
\(23380\) 	&	0.18677 	&	0.95000 	&	0.99841 	&	0.98265 	&	\B 1.00000 	&	--		&	0.97131 	&	\B 1.00000 	\\
\(23381\) 	&	0.50333 	&	0.55867 	&	0.55556 	&	0.56667 	&	0.56867 	&	\B 0.66978 	&	0.56844 	&	0.58400 	\\
\(23512\) 	&	0.50065 	&	0.67445 	&	0.71930 	&	\B 0.72296 	&	\ul{0.72031} 	&	0.67135 	&	0.70743 	&	0.71281 	\\
\(23517^-\) 	&	0.49962 	&	0.50259 	&	\ul{0.51939} 	&	\ul{0.51926} 	&	\B 0.52082 	&	0.51941 	&	\ul{0.52033} 	&	0.50635 	\\
\(40536\) 	&	0.72550 	&	0.85195 	&	\ul{0.86225} 	&	\ul{0.86291} 	&	\ul{0.86392} 	&	\ul{0.86128} 	&	\B 0.86661 	&	0.84968 	\\
\(40668^+\) 	&	0.50439 	&	0.78341 	&	0.79628 	&	0.82109 	&	0.84123 	&	0.77698 	&	0.82886 	&	\B 0.86500 	\\
\(40670^+\) 	&	0.39100 	&	0.91412 	&	0.95889 	&	0.95962 	&	0.95931 	&	0.95282 	&	0.96109 	&	\B 0.96904 	\\
\(40685^+\) 	&	0.64405 	&	0.99962 	&	0.99968 	&	\ul{0.99978} 	&	\ul{0.99974} 	&	0.99955 	&	0.99253 	&	\B 0.99987 	\\
\(40701^-\) 	&	0.76320 	&	0.94313 	&	0.95313 	&	\ul{0.95620} 	&	\B 0.96000 	&	0.95007 	&	\ul{0.94533} 	&	0.95370 	\\
\(40923^+\) 	&	0.02127 	&	0.78048 	&	0.02169 	&	0.74009 	&	--		&	\B 0.89470 	&	0.86438 	&	0.58220 	\\
\(40927^-\) 	&	0.10096 	&	0.35102 	&	--		&	--		&	0.29429 	&	\ul{0.32001} 	&	\ul{0.32093} 	&	\B 0.36389 	\\
\(40966\) 	&	0.12407 	&	0.94228 	&	0.99506 	&	0.99043 	&	0.99506 	&	\B 1.00000 	&	0.96380 	&	0.99551 	\\
\(40975^-\) 	&	0.53218 	&	0.95318 	&	0.97958 	&	0.97264 	&	\B 0.99422 	&	0.96763 	&	0.98786 	&	\ul{0.99191} 	\\
\(40978^-\) 	&	0.75346 	&	\ul{0.97368} 	&	0.97114 	&	\B 0.97774 	&	\ul{0.97398} 	&	0.96900 	&	0.97358 	&	--		\\
\(40979^-\) 	&	0.09983 	&	0.95217 	&	0.97367 	&	\ul{0.97783} 	&	0.96883 	&	\ul{0.97750} 	&	\B 0.98121 	&	0.97600 	\\
\(40981^-\) 	&	0.49324 	&	0.85604 	&	0.85556 	&	\ul{0.87053} 	&	0.86184 	&	\B 0.89050 	&	\ul{0.86913} 	&	\ul{0.87633} 	\\
\(40982^-\) 	&	0.21681 	&	0.74425 	&	0.76364 	&	\ul{0.78268} 	&	\B 0.79091 	&	0.76415 	&	0.75955 	&	\ul{0.78062} 	\\
\(40983^-\) 	&	0.89683 	&	0.97886 	&	\ul{0.98581} 	&	\ul{0.98612} 	&	\ul{0.98540} 	&	\B 0.98657 	&	0.95289 	&	\ul{0.98574} 	\\
\(40984^+\) 	&	0.14473 	&	0.93001 	&	\ul{0.93333} 	&	0.93088 	&	\ul{0.94055} 	&	0.92564 	&	0.90664 	&	\B 0.94185 	\\
\(40994^-\) 	&	0.83704 	&	0.91914 	&	0.92407 	&	0.94074 	&	0.94547 	&	\B 0.96975 	&	0.92593 	&	0.93642 	\\
\(40996^+\) 	&	0.09844 	&	0.85777 	&	0.84450 	&	\B 0.87844 	&	0.78089 	&	0.82114 	&	0.85060 	&	\ul{0.87341} 	\\
\(41027^+\) 	&	0.42598 	&	0.78945 	&	0.85378 	&	0.86775 	&	\ul{0.88735} 	&	0.87540 	&	\ul{0.88691} 	&	\B 0.90047 	\\
\(41138\) 	&	0.96474 	&	0.99268 	&	0.99137 	&	0.99287 	&	\ul{0.99339} 	&	0.97097 	&	\ul{0.99360} 	&	\B 0.99369 	\\
\(41142^+\) 	&	0.50234 	&	0.67977 	&	0.73081 	&	\B 0.74754 	&	0.72645 	&	0.72169 	&	0.71630 	&	0.72811 	\\
\(41143^-\) 	&	0.50748 	&	0.78170 	&	0.80603 	&	\ul{0.82009} 	&	\B 0.82366 	&	0.79911 	&	0.80078 	&	\ul{0.80906} 	\\
\(41146^+\) 	&	0.49532 	&	0.93062 	&	0.94753 	&	0.93921 	&	\B 0.95533 	&	0.93476 	&	0.94675 	&	0.92510 	\\
\(41147\) 	&	0.49923 	&	0.62564 	&	0.66709 	&	0.68314 	&	0.66110 	&	\B 0.80064 	&	0.66694 	&	0.64798 	\\
\(41150^+\) 	&	0.59589 	&	0.92356 	&	0.92891 	&	0.94334 	&	0.93850 	&	0.90234 	&	0.87477 	&	\B 0.94604 	\\
\(41159^+\) 	&	0.51942 	&	0.77610 	&	--		&	0.64227 	&	0.72548 	&	0.66063 	&	\ul{0.74347} 	&	\B 0.81928 	\\
\(41161^-\) 	&	0.62482 	&	0.93468 	&	0.75042 	&	0.74757 	&	\B 0.98495 	&	0.90729 	&	0.82518 	&	0.95625 	\\
\(41163^+\) 	&	0.19703 	&	0.92263 	&	0.94793 	&	\B 0.98357 	&	0.96254 	&	0.95391 	&	0.97243 	&	0.96988 	\\
\(41164^+\) 	&	0.16375 	&	0.66570 	&	0.67395 	&	0.70255 	&	0.68336 	&	0.67357 	&	0.69104 	&	\B 0.71752 	\\
\(41165^+\) 	&	0.09480 	&	0.30877 	&	0.39922 	&	\B 0.44843 	&	--		&	0.35252 	&	0.34203 	&	--		\\
\(41166^+\) 	&	0.14885 	&	0.61045 	&	0.63762 	&	0.66933 	&	0.65075 	&	\B 0.67940 	&	0.65451 	&	\ul{0.67841} 	\\
\(41167^-\) 	&	0.00286 	&	\B 0.87164 	&	--		&	--		&	--		&	0.38666 	&	0.77971 	&	--		\\
\(41168^+\) 	&	0.36200 	&	0.65848 	&	0.69273 	&	\B 0.71814 	&	0.69642 	&	0.63788 	&	0.68494 	&	\ul{0.71786} 	\\
\(41169^+\) 	&	0.02272 	&	0.29082 	&	\ul{0.29566} 	&	0.30692 	&	\B 0.33576 	&	\ul{0.32108} 	&	0.28741 	&	--		\\
\\
Average & 0.44921 	&	0.79980 	&	0.80853 	&	\ul{0.82606} 	&	\B 0.83040 	&	\ul{0.80292} 	&	0.81075 	&	\ul{0.82910} 	\\
\\

	\bottomrule

\caption{
	Average accuracy of \ac{AutoML} frameworks on selected \name{OpenML} data sets. Entries marked by -- consistently failed to generate an \ac{ML} pipeline. The best results per data set are highlighted in bold. Results not significantly worse than the best result---according to a Wilcoxon signed-rank test---are underlined. On data sets marked by \(^+\) and \(^-\), \ac{AutoML} frameworks performed better and worse, respectively, than \ac{CASH} solvers.
}
\label{tbl:results_evaluation_frameworks}

\end{longtable}
\end{footnotesize}

\begin{figure}
	\centering
	
	\begin{subfigure}[b]{0.32\textwidth}
		\includegraphics[width=\textwidth]{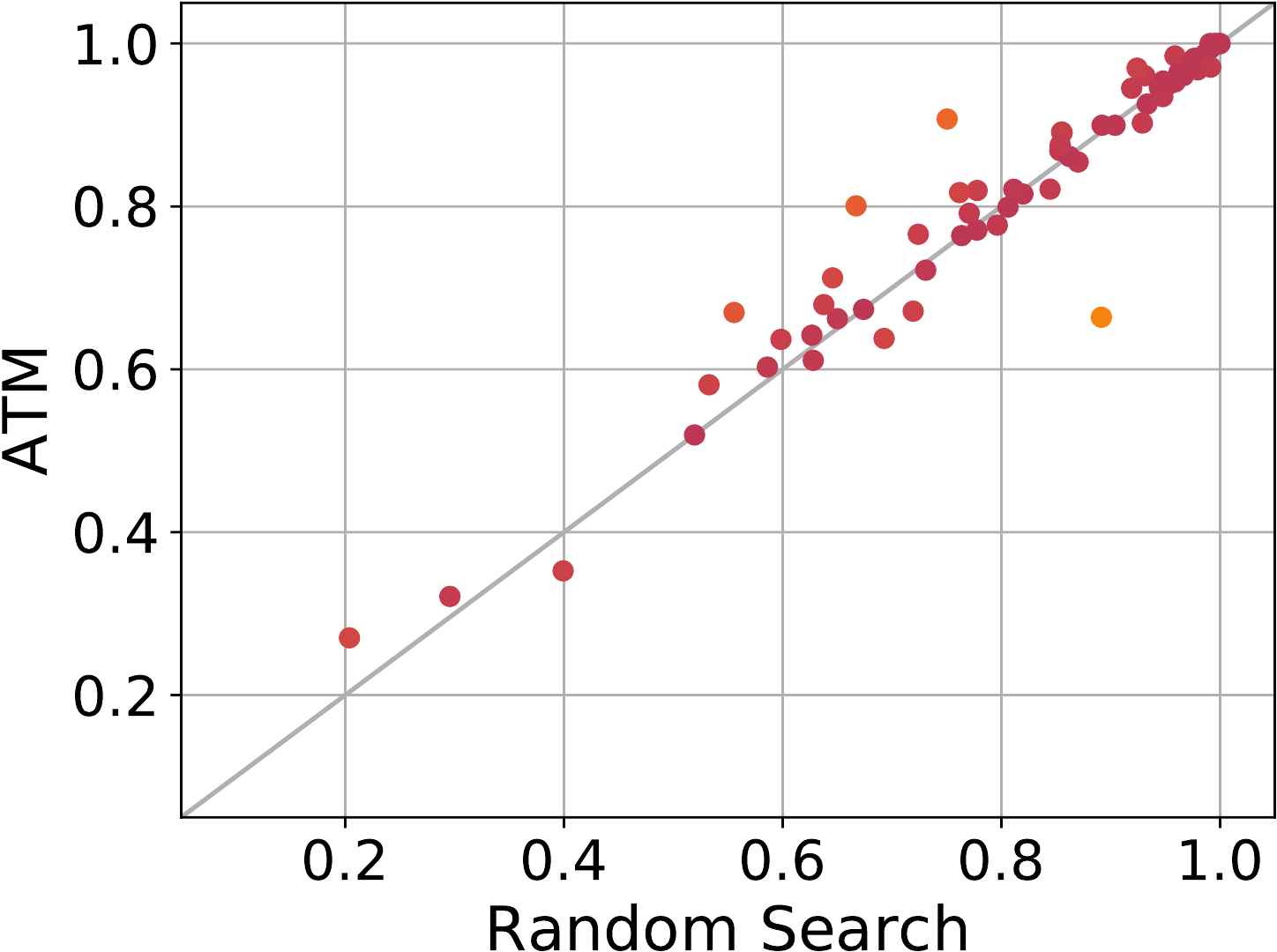}
	\end{subfigure}
	\begin{subfigure}[b]{0.32\textwidth}
		\includegraphics[width=\textwidth]{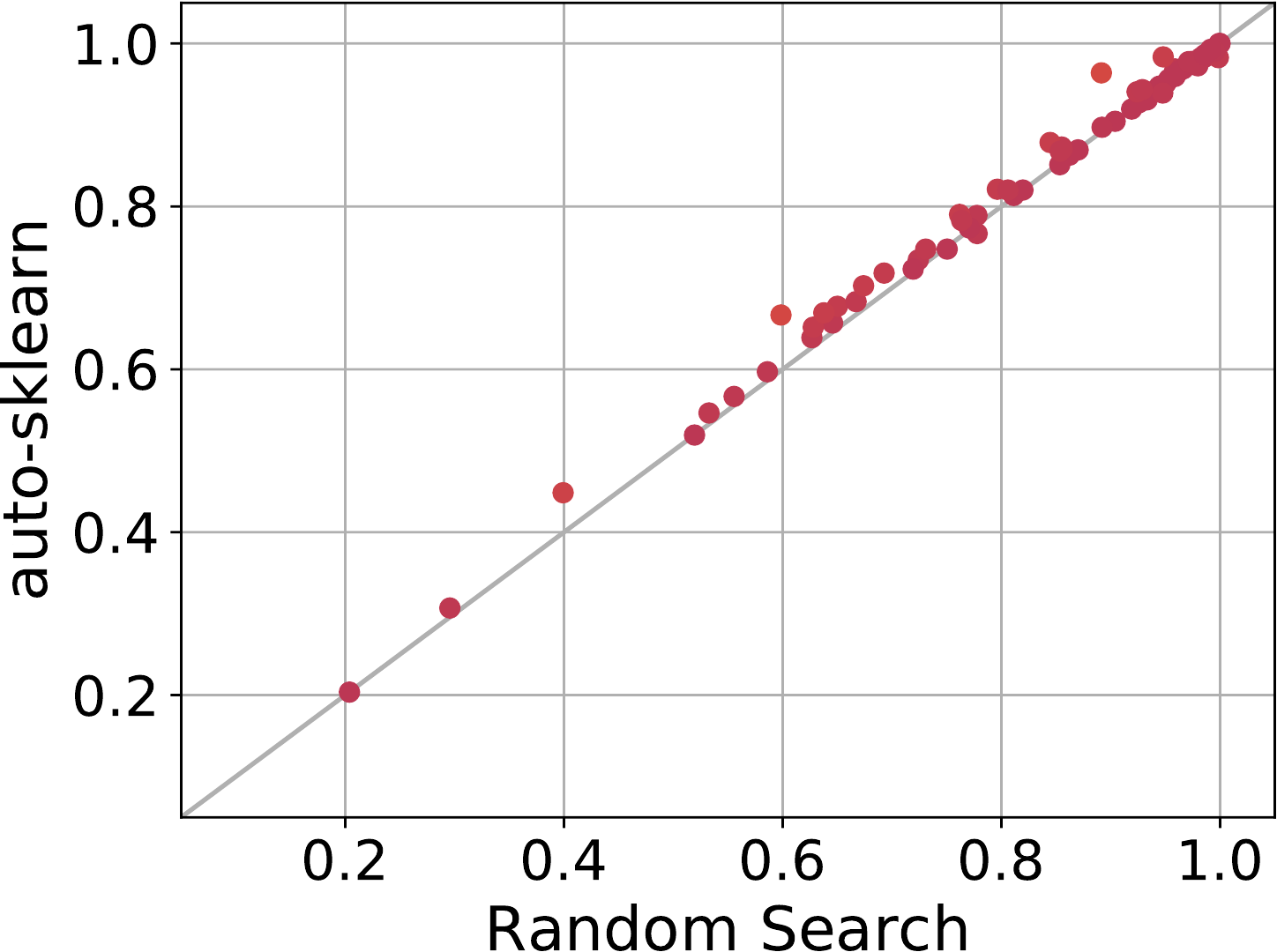}
	\end{subfigure}
	\begin{subfigure}[b]{0.32\textwidth}
		\includegraphics[width=\textwidth]{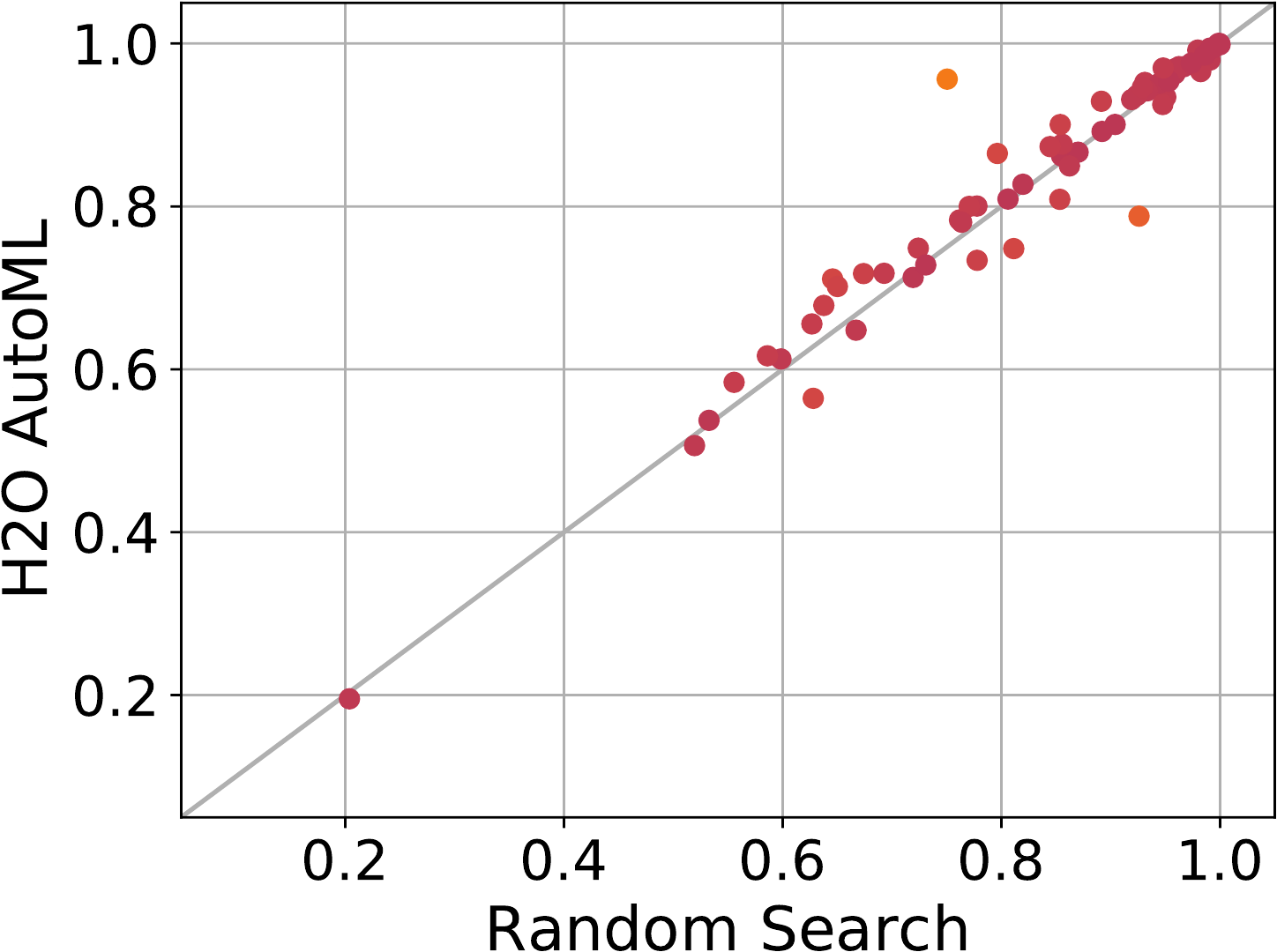}
	\end{subfigure}
	
	\begin{subfigure}[b]{0.32\textwidth}
		\includegraphics[width=\textwidth]{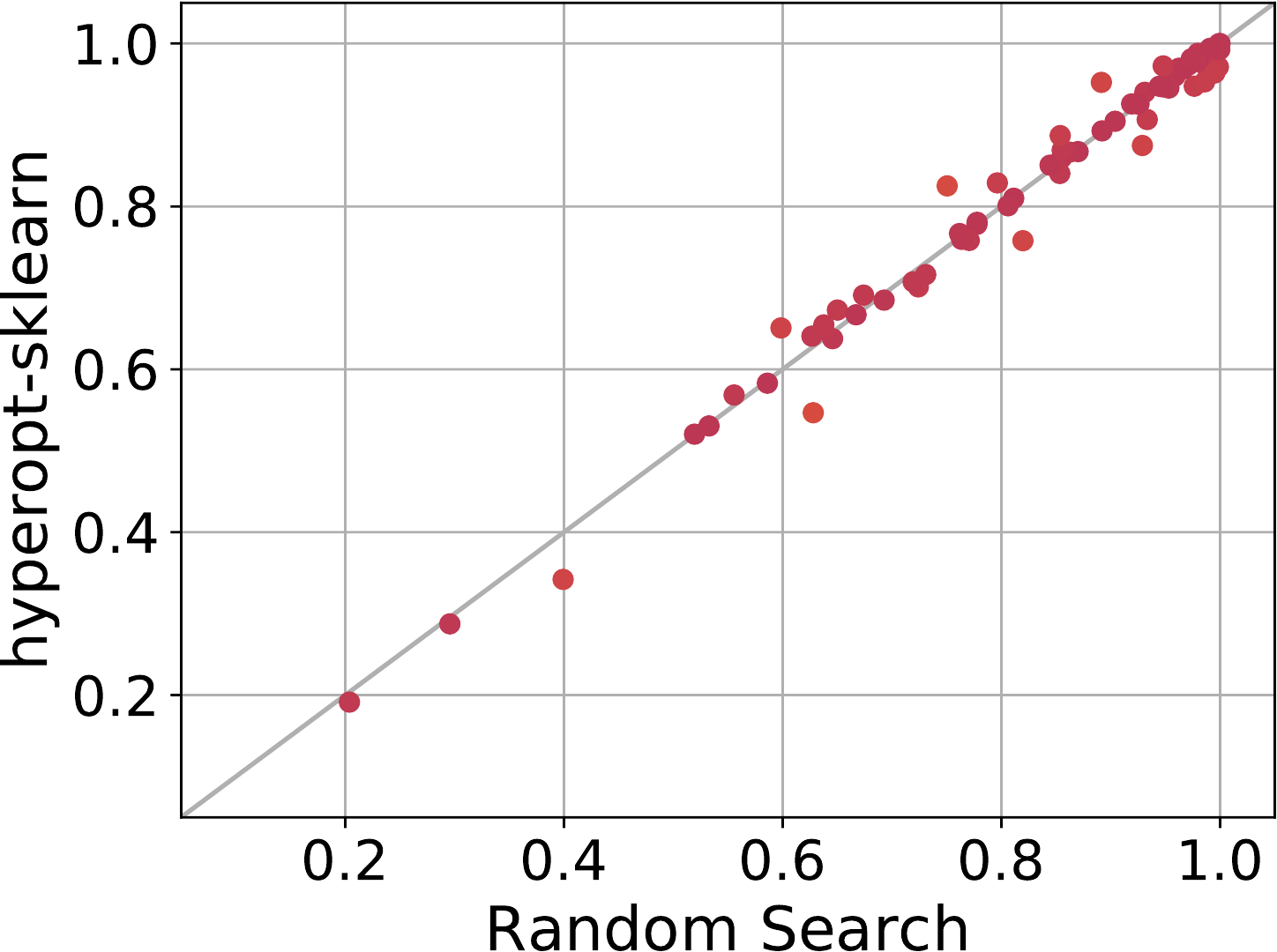}
	\end{subfigure}
	\begin{subfigure}[b]{0.32\textwidth}
		\includegraphics[width=\textwidth]{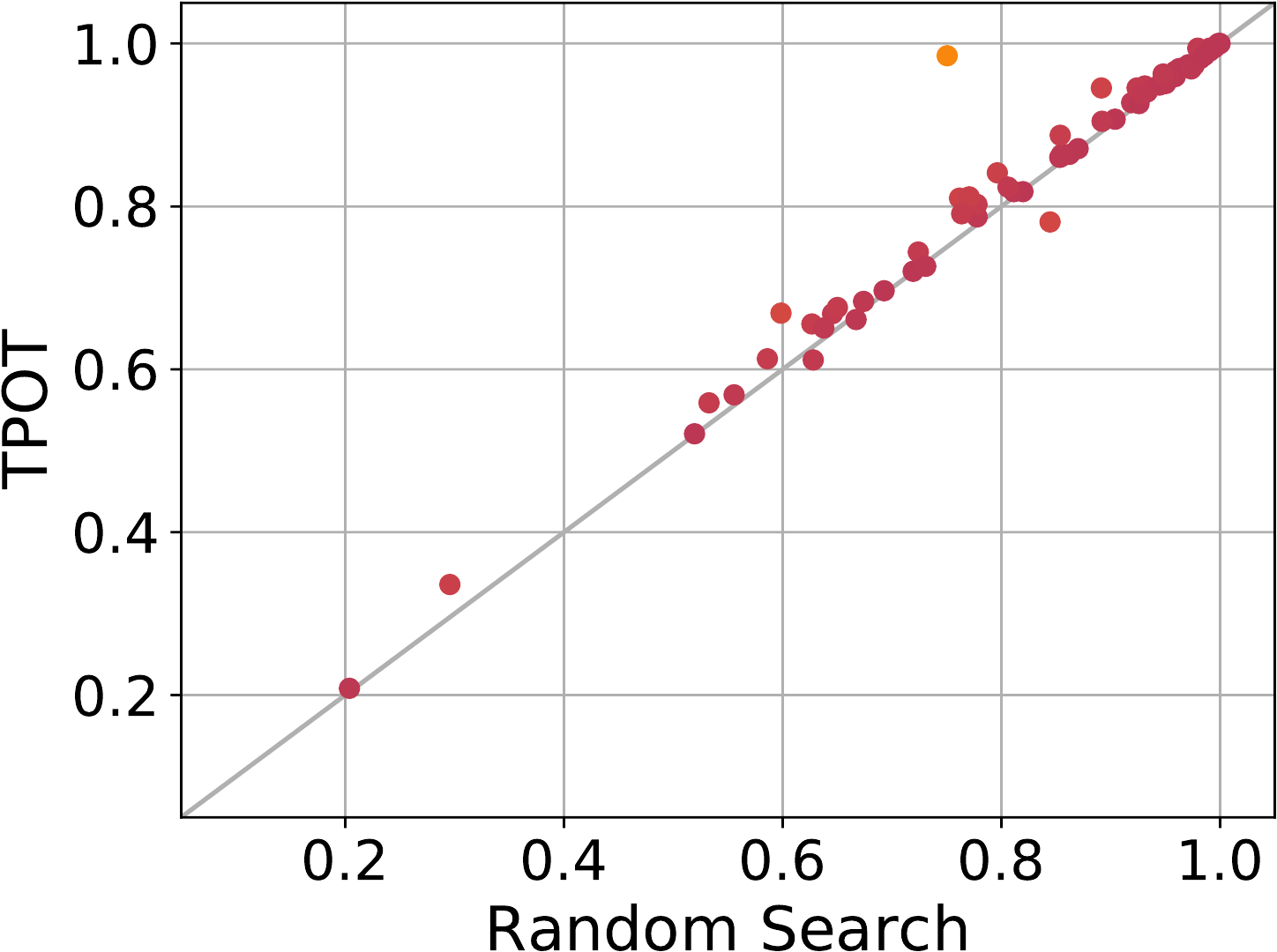}
	\end{subfigure}
	\begin{subfigure}[b]{0.32\textwidth}
		\includegraphics[width=\textwidth]{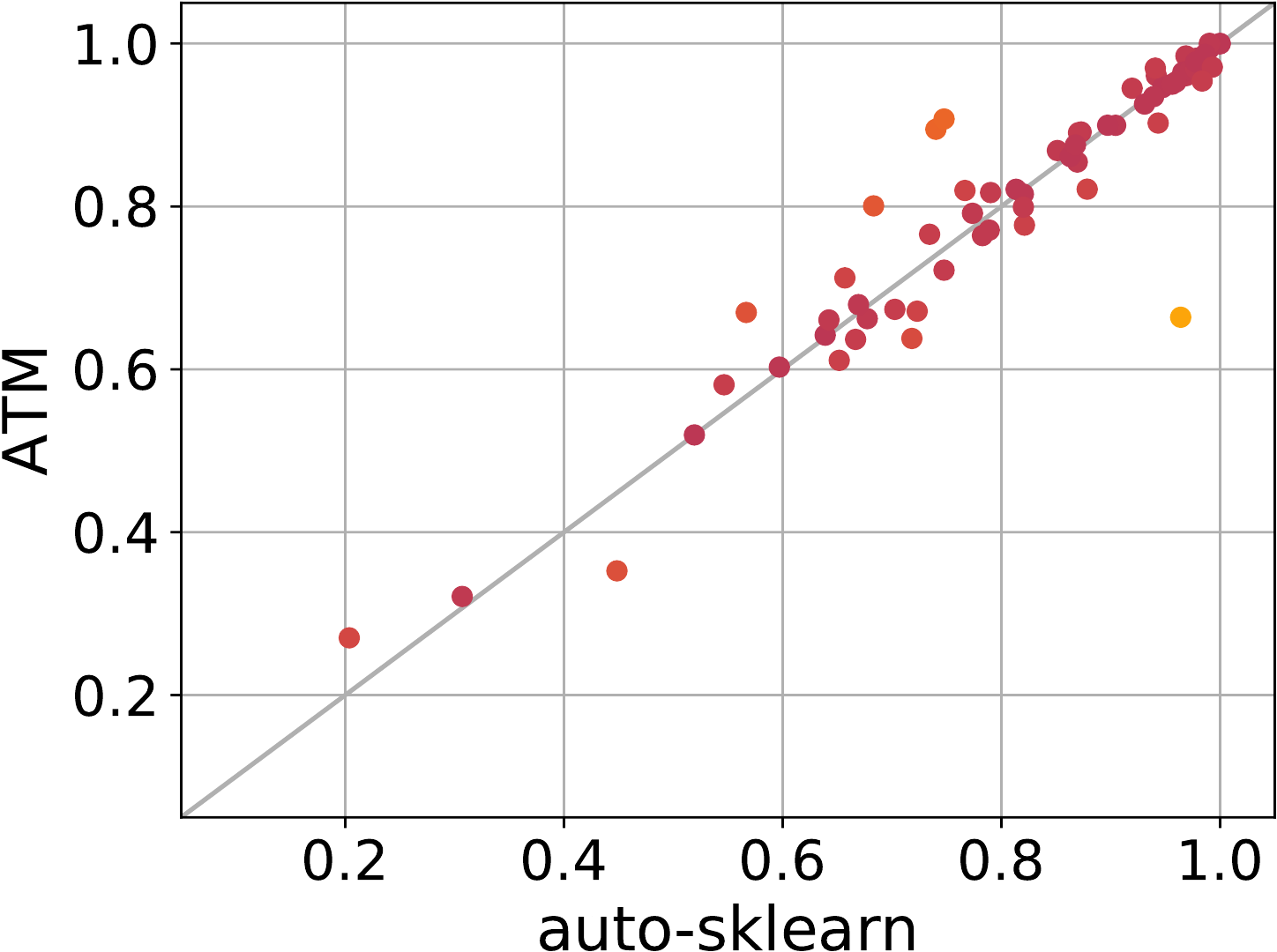}
	\end{subfigure}
	
	\begin{subfigure}[b]{0.32\textwidth}
		\includegraphics[width=\textwidth]{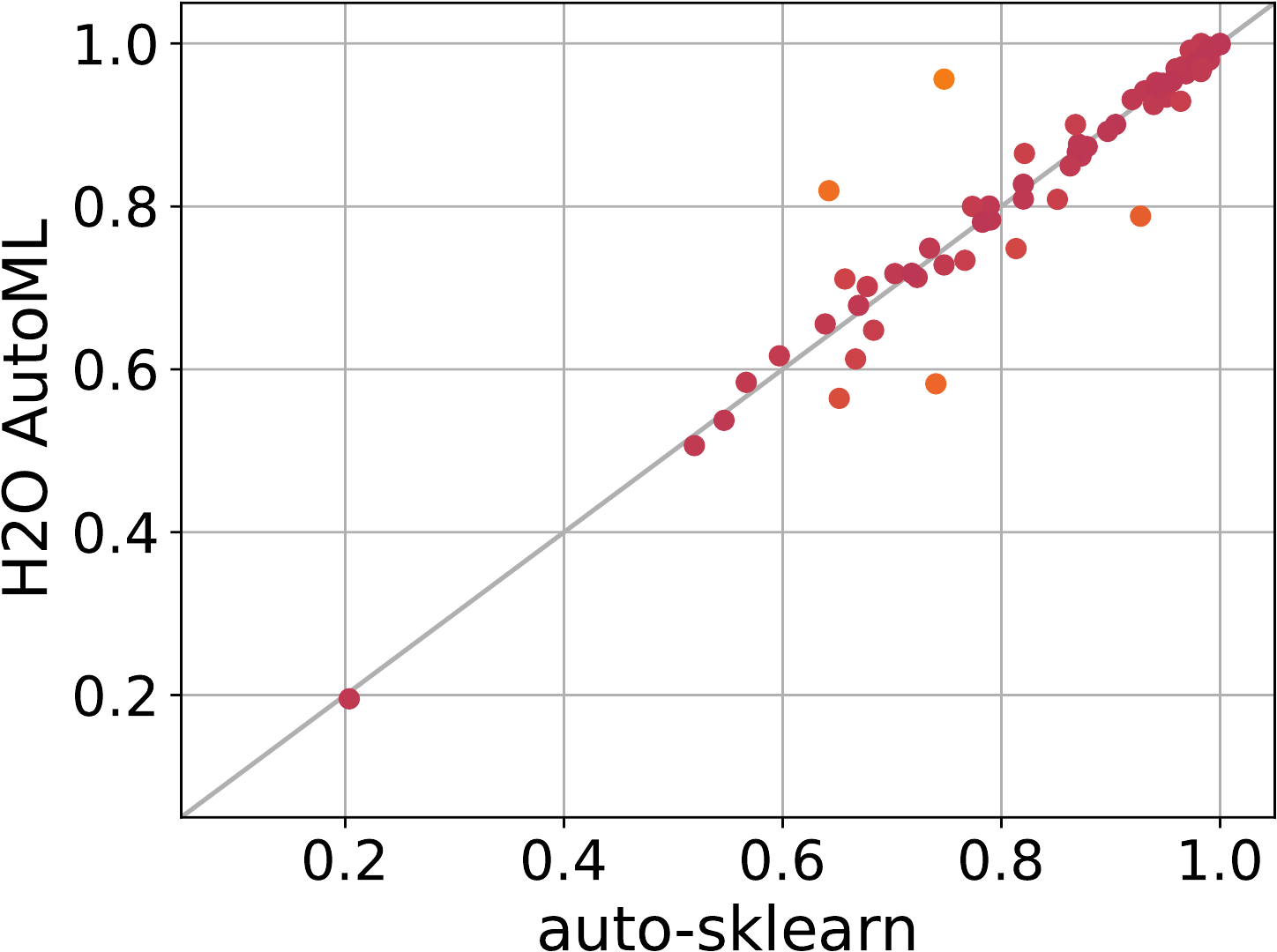}
	\end{subfigure}
	\begin{subfigure}[b]{0.32\textwidth}
		\includegraphics[width=\textwidth]{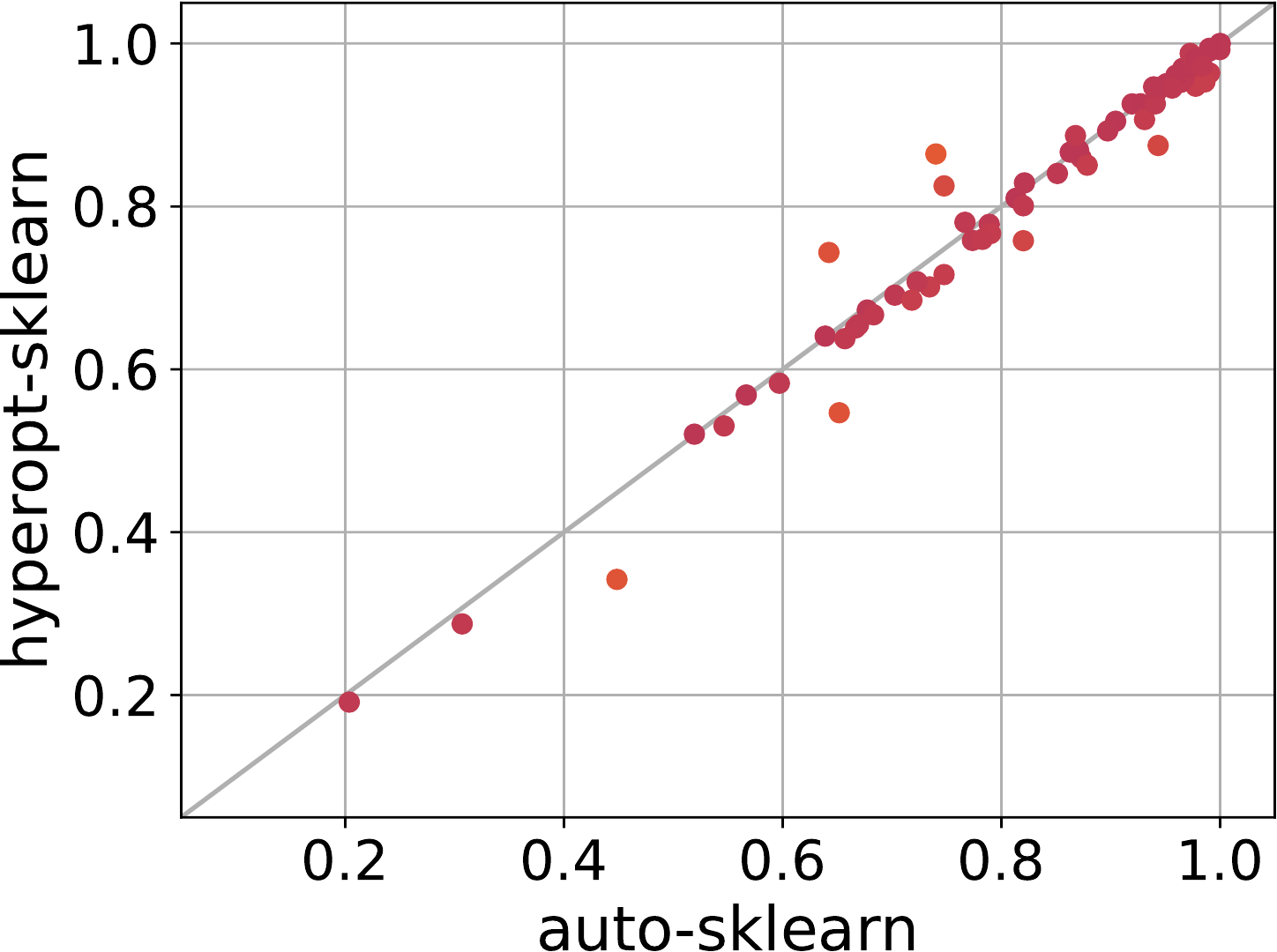}
	\end{subfigure}
	\begin{subfigure}[b]{0.32\textwidth}
		\includegraphics[width=\textwidth]{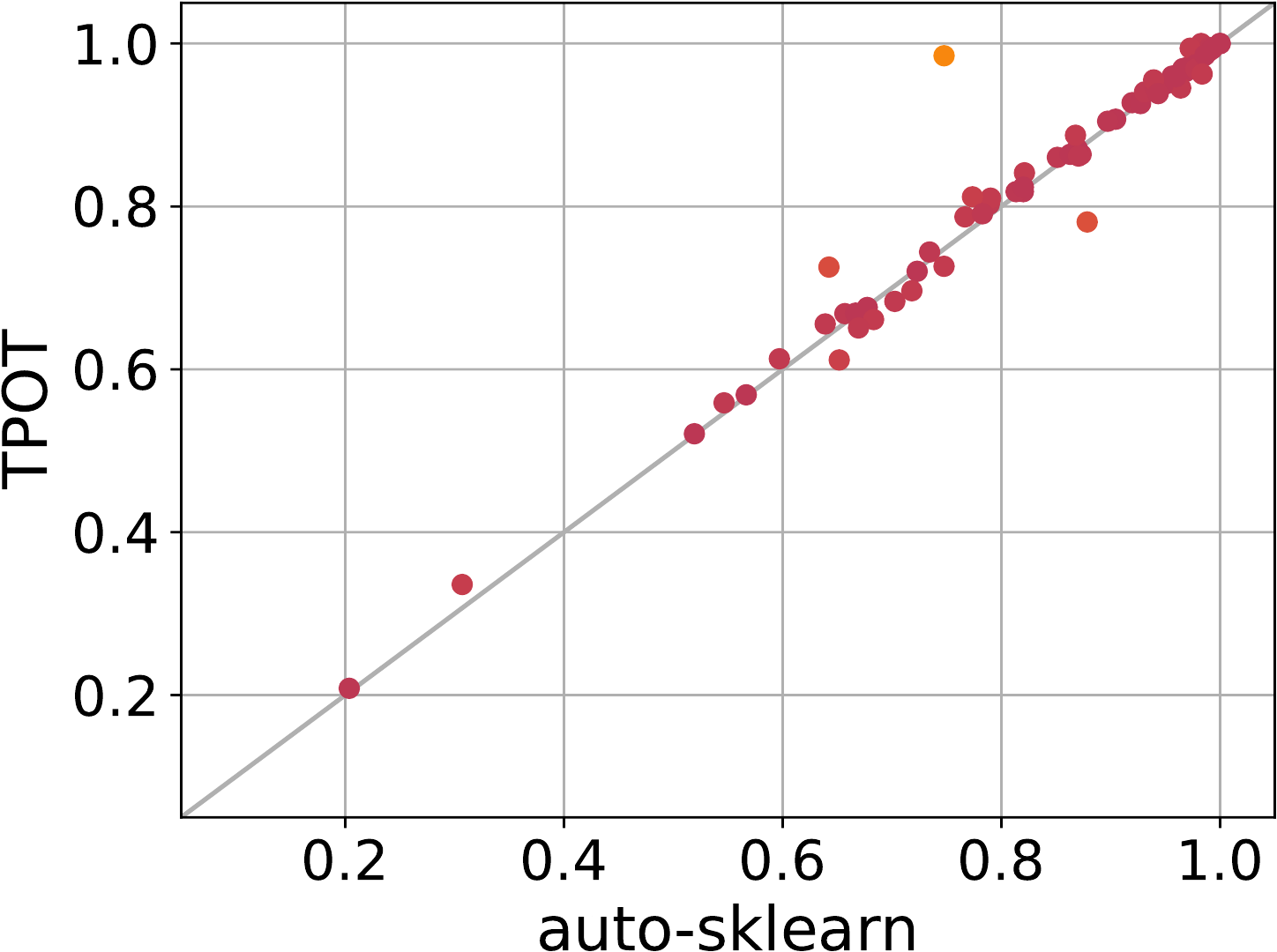}
	\end{subfigure}
	
	\begin{subfigure}[b]{0.32\textwidth}
		\includegraphics[width=\textwidth]{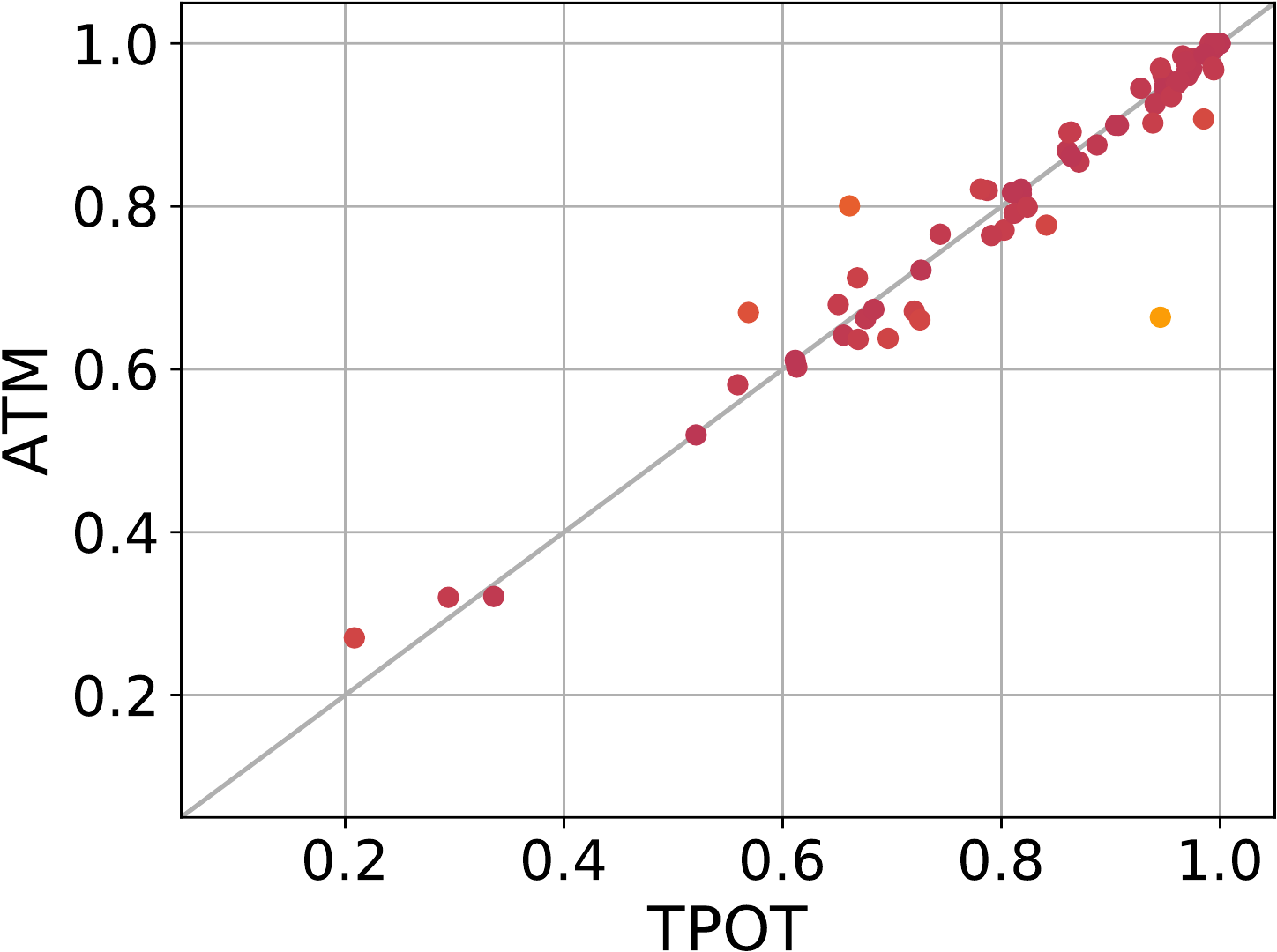}
	\end{subfigure}
	\begin{subfigure}[b]{0.32\textwidth}
		\includegraphics[width=\textwidth]{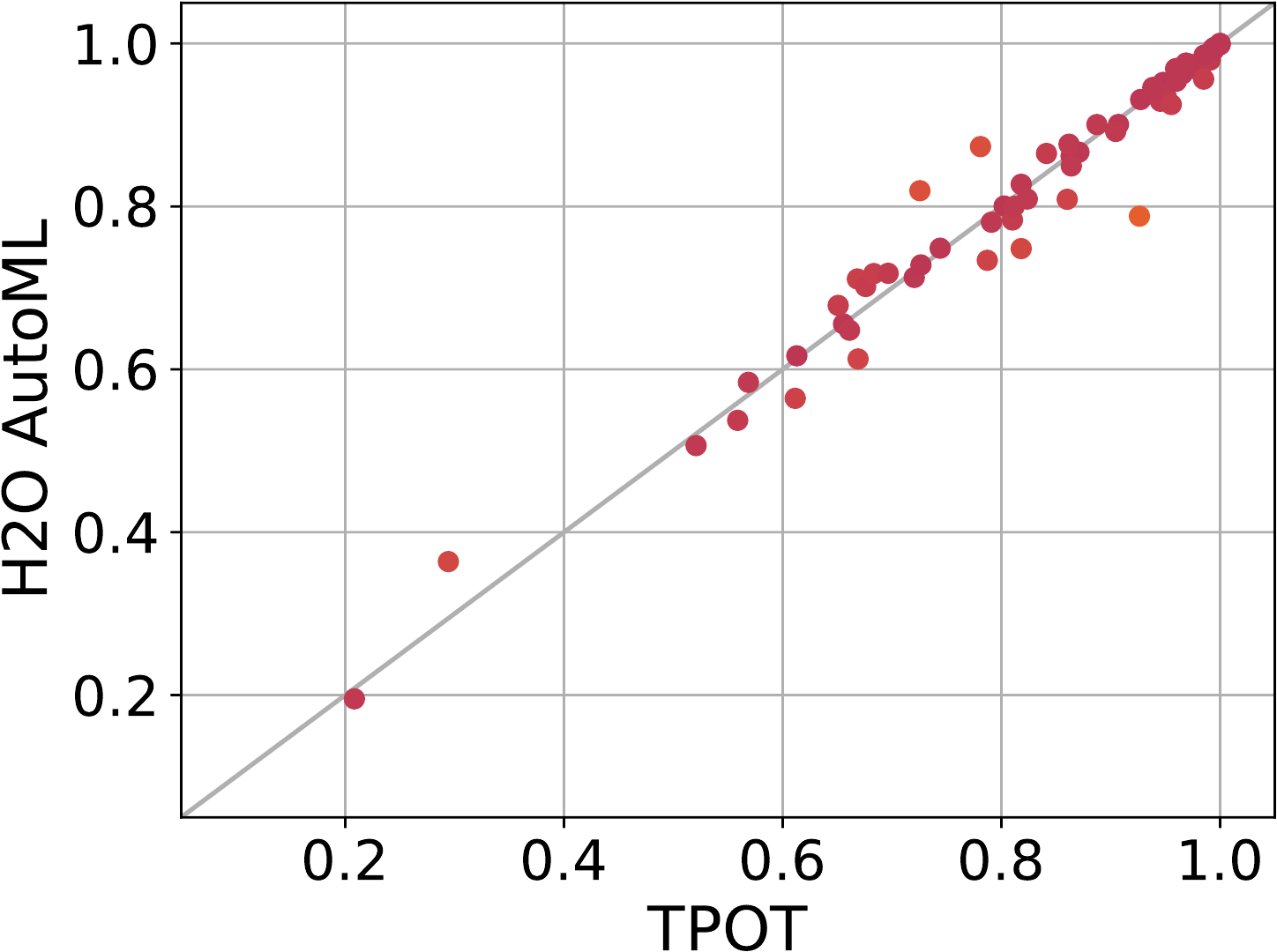}
	\end{subfigure}
	\begin{subfigure}[b]{0.32\textwidth}
		\includegraphics[width=\textwidth]{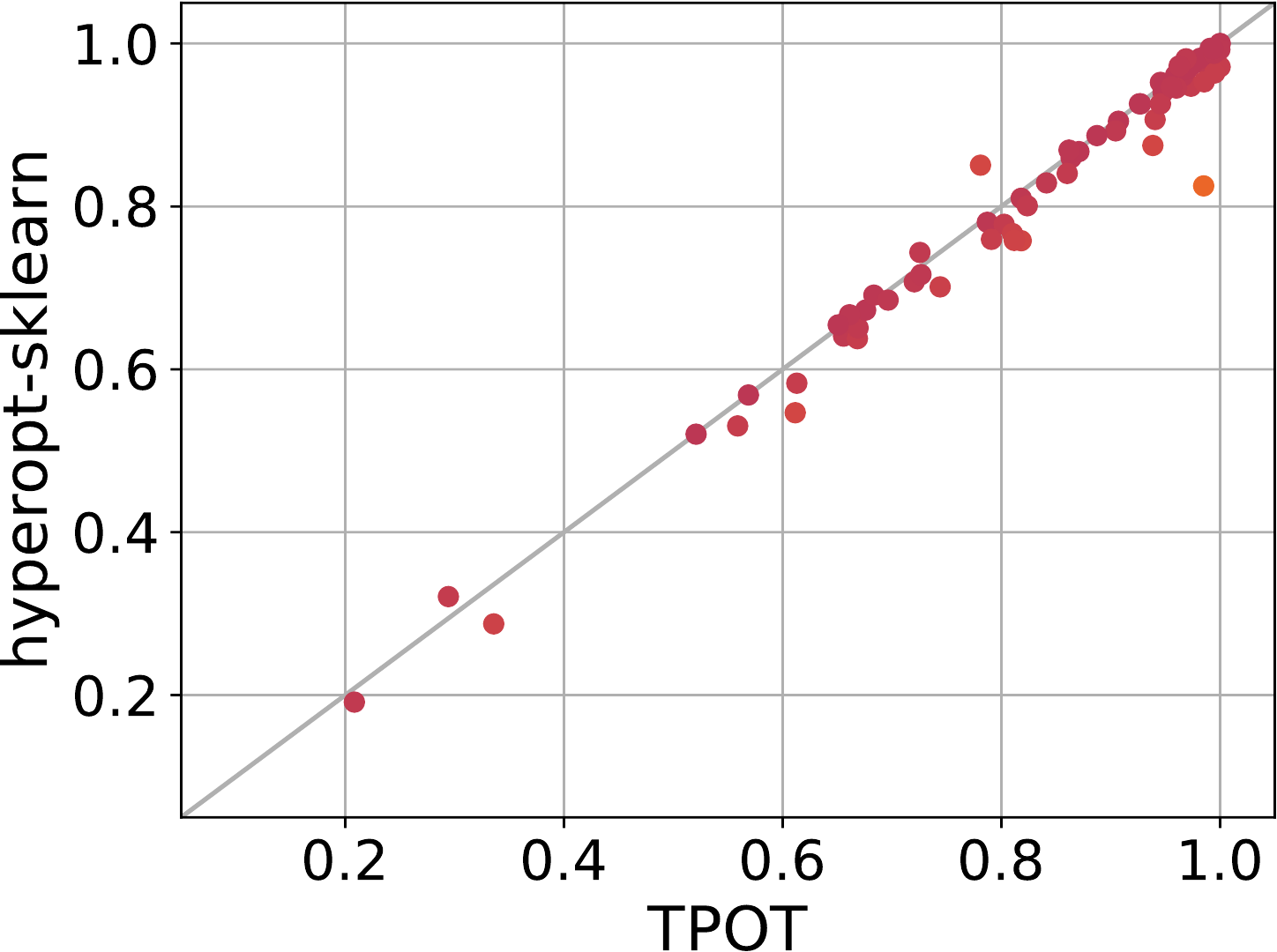}
	\end{subfigure}
	
	\begin{subfigure}[b]{0.32\textwidth}
		\includegraphics[width=\textwidth]{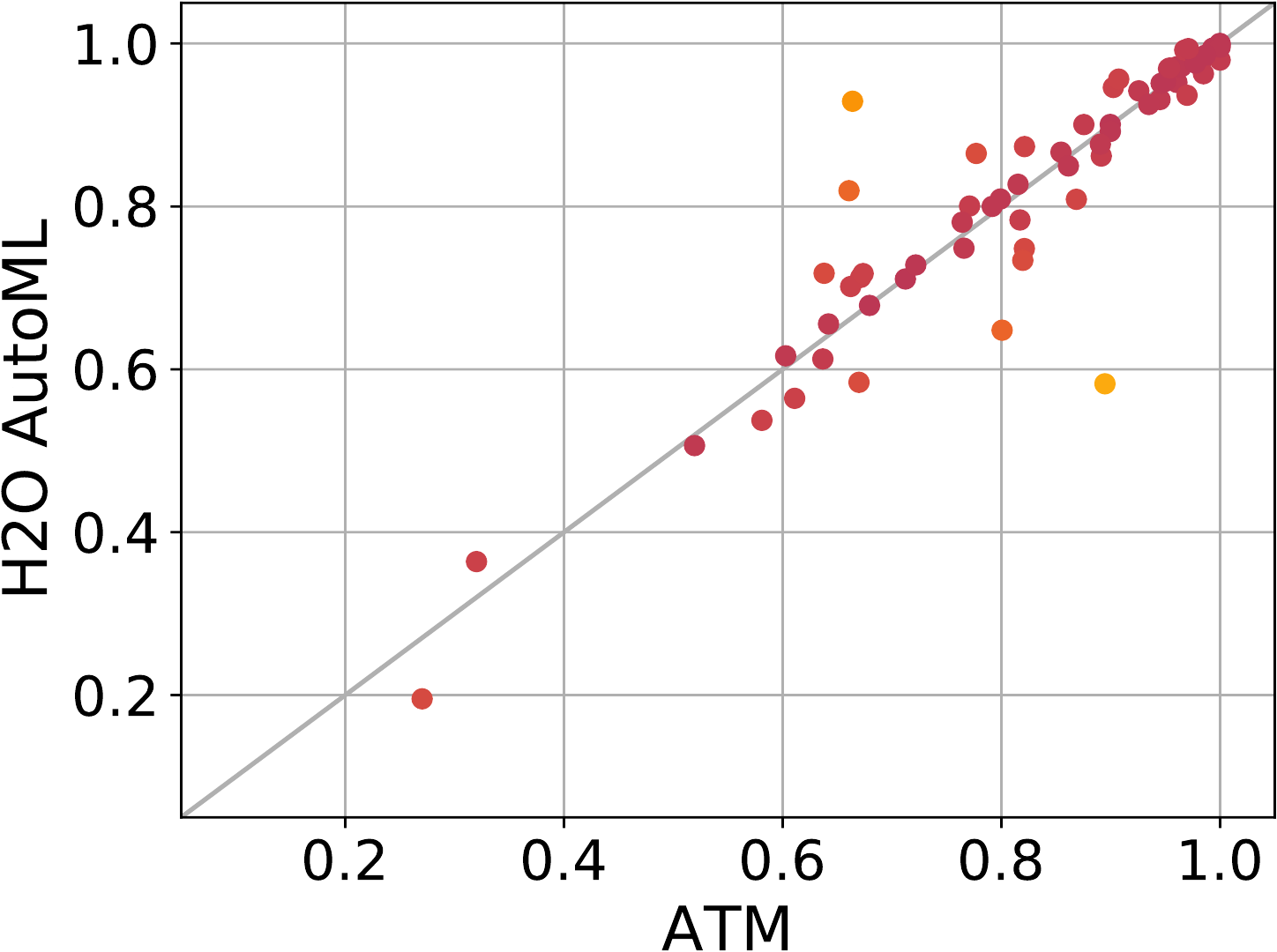}
	\end{subfigure}
	\begin{subfigure}[b]{0.32\textwidth}
		\includegraphics[width=\textwidth]{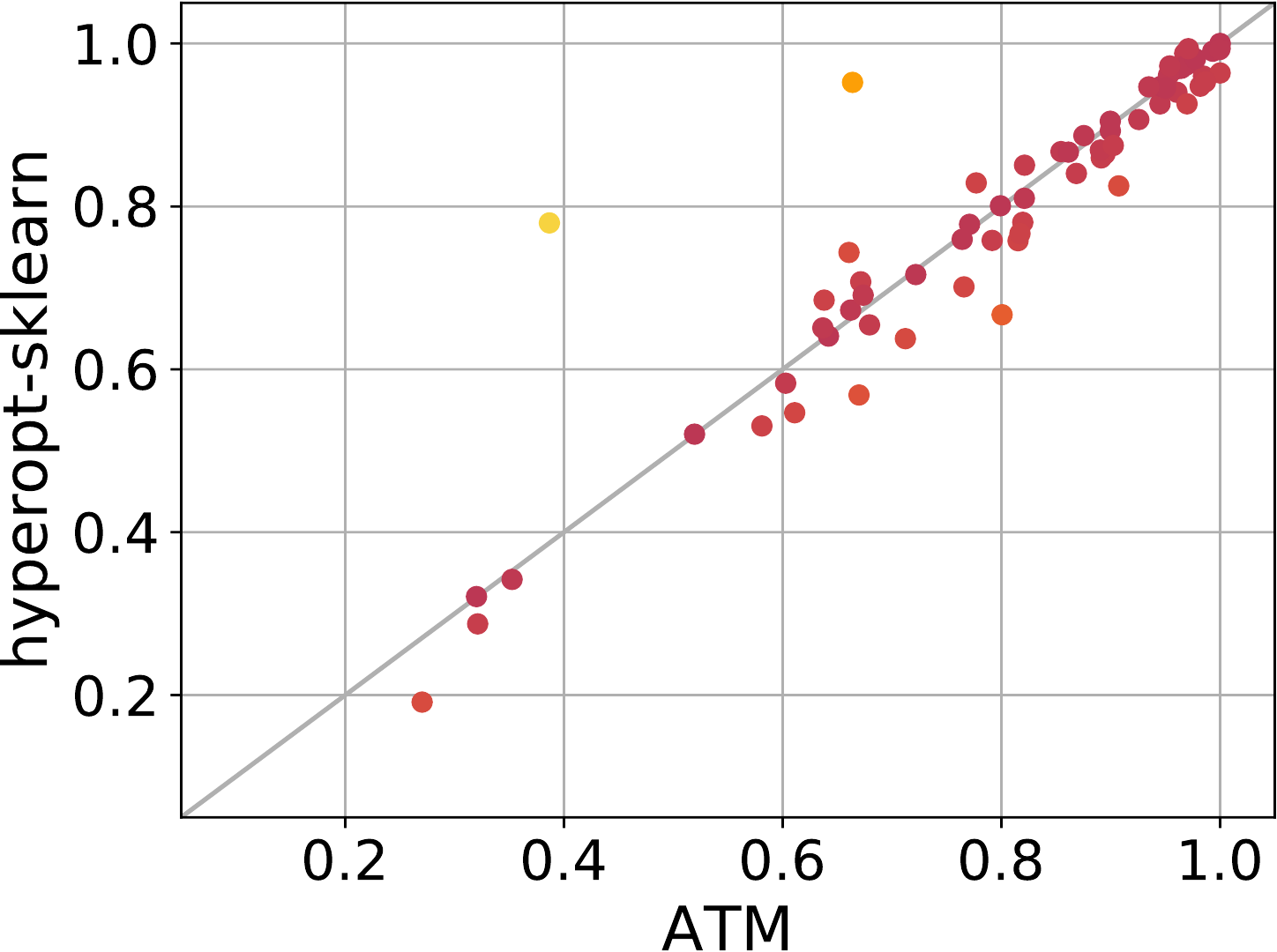}
	\end{subfigure}
	\begin{subfigure}[b]{0.32\textwidth}
		\includegraphics[width=\textwidth]{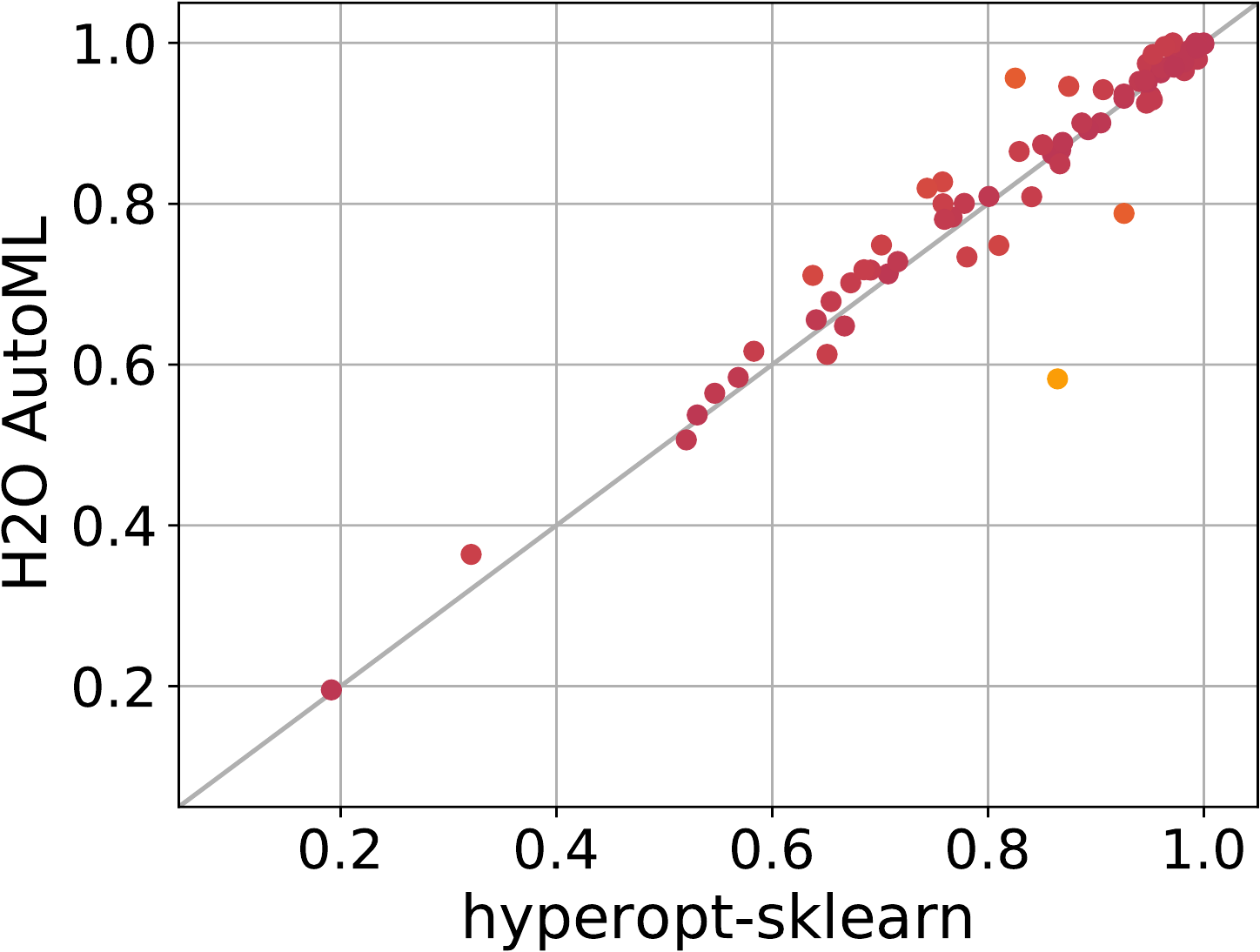}
	\end{subfigure}
	
	\caption{Pair-wise comparison of normalized performances of \ac{AutoML} frameworks. The axes represent the accuracy score of the stated \ac{AutoML} framework. Each point represents the averaged results for a single data set. Identical performances are plotted directly on the angle bisector.}
	\label{fig:pair_wise_automl_results}
\end{figure}

\FloatBarrier

\bibliography{library}
\bibliographystyle{theapaurl}

\end{document}